\documentclass{article}
\usepackage{arxiv}
\usepackage{natbib}
\usepackage{amsmath}
\usepackage{amssymb}
\usepackage{mathtools}
\usepackage{amsthm}
\usepackage{subfigure}
\usepackage{caption}
\usepackage[utf8]{inputenc} 
\usepackage[T1]{fontenc}    
\usepackage{hyperref}       
\usepackage{url}            
\usepackage{booktabs}       
\usepackage{amsfonts}       
\usepackage{nicefrac}       
\usepackage{microtype}      
\usepackage{xcolor}         
\usepackage{longtable} 
\usepackage{booktabs}  
\usepackage{fourier}
\usepackage{graphicx}
\usepackage{multirow}
\usepackage{microtype}
\usepackage{booktabs} 
\newcommand{\res}[2]{#1\,{\scriptsize$\pm$\,#2}}

\title{A Monosemantic Attribution Framework for Stable Interpretability in Clinical Neuroscience Transformer-Based Language Models}

\author{
 Michail Mamalakis \\
   Department of Computer Science and Technology \\
  Cancer Research UK Cambridge Institute\\
  University of Cambridge\\
  United Kingdom \\
  \texttt{mm2703@cam.ac.uk} \\  
   \And
 Tiago Azevedo \\
  Department of Computer Science and Technology\\
  University of Cambridge\\
  United Kingdom \\
  \And
 Cristian Cosentino \\
DIMES\\
University of Calabria\\ 
  Italy \\
    \And
Chiara D'Ercoli\\
Department of Computer\\
Automatic and Management Engineering (DIAG)\\
Sapienza Università di Roma\\
Italy\\
    \And
 Subati Abulikemu \\
    Department of Psychiatry\\
  University of Cambridge\\
  United Kingdom \\
    \And
Zhongtian Sun\\
School of Computing\\
University of Kent\\ 
United Kingdom\\
      \And
Richard Bethlehem\\
    Department of Psychology\\
  University of Cambridge\\
  United Kingdom \\
    \And
 Pietro Lio \\
  Department of Computer Science and Technology\\
  University of Cambridge\\
  United Kingdom \\
}

\begin{document}
\maketitle
\begin{abstract}
Interpretability remains a key challenge for deploying language models (LMs) in clinical settings such as Alzheimer’s disease progression diagnosis, where early and trustworthy predictions are essential. Existing attribution methods exhibit high inter-method variability and unstable explanations due to the polysemantic nature of Transformer-Based LM and LLM representations, while mechanistic interpretability approaches lack direct alignment with model inputs and outputs and do not provide explicit importance scores. We introduce a unified interpretability framework that integrates attributional and mechanistic perspectives through monosemantic feature extraction. By constructing a monosemantic embedding space at the level of an transformer-based LM layer and optimizing the framework to explicitly reduce inter-method variability, our approach produces stable input-level importance scores and highlights salient features via a decompressed representation of the layer of interest, advancing the safe and trustworthy application of LMs in cognitive health and neurodegenerative disease.
\end{abstract}

\keywords{Mechanistic Interpretability \and explainable AI \and Atributional Interpretability \and Alzheimer \and monosemantic \and polysemantic}

\section{Introduction}
\label{intro}
Explainable Artificial Intelligence (XAI) plays a crucial role in building trust in machine learning systems, especially in sensitive and high-stakes areas such as finance, climate, and healthcare \cite{doshi2017towards,manif2021trust}. In medical settings, interpretability is essential for clinical integration and regulatory approval, especially in complex diseases such as Alzheimer's Disease (AD) progression \cite{tahami}, where early and accurate detection can substantially alter treatment results \cite{jack2018nia}.

Although machine learning has shown promise in Alzheimer’s progression (AD) diagnostics using multimodal clinical data \cite{lee2019predicting}, the application of large language models (LLMs) such as GPT-4, LLaMA-2 and Transformer-Based Language models (LMs) like BERT family \cite{devlin2018bert} in structured clinical settings remains limited \cite{brown2020language, touvron2023llama}. A key obstacle is the \emph{polysemanticity} of internal neural representations—individual neurons or features often encode multiple, semantically unrelated concepts \cite{olah2020zoom, sae,toy}. This entanglement undermines the interpretability of standard attribution techniques such as gradients, perturbations, and integrated paths, which typically assume one-to-one correspondence between features and meanings. To this end, existing attribution methods assign importance scores to input features (e.g., words or tokens), yet they fall short in addressing the polysemantic nature of internal representations. This limitation often leads to ambiguous or misleading explanations—particularly problematic in clinical applications, where interpretability is critical \cite{xai, mx1, mx2, mine}. Moreover, the inter-method variability of attributional techniques is another limitation of dataset-driven explanations, as it reduces trust in the interpretability of AI models \cite{mamalakis}. 

In contrast, mechanistic interpretability aims to uncover the internal structure of neural computation by identifying semantically coherent components within the model. Sparse Autoencoders (SAE) have played a pivotal role in advancing our understanding of such representations in both language and vision domains~\cite{ref15}. SAEs aim to solve the superposition problem in neural feature representations by mapping the model’s activations into a more monosemantic latent space, where individual features are better aligned with specific concepts in the network \cite{sae}. However, these mechanistic tools typically lack attributional resolution, limiting their utility for explaining how specific inputs contribute to model predictions in real-world, decision-critical scenarios \cite{toy}. 

This reveals a critical gap in the current landscape: attributional techniques offer only surface-level explanations using polysemantic features, while mechanistic methods provide structural insights based on monosemantic feature alignment but lack attributional grounding. To date, no unified framework successfully integrates both paradigms—particularly in the domain of LLMs and Transformer-Based LM clinical inference.

\begin{figure*}
\centering 
\includegraphics[width=0.9\textwidth]{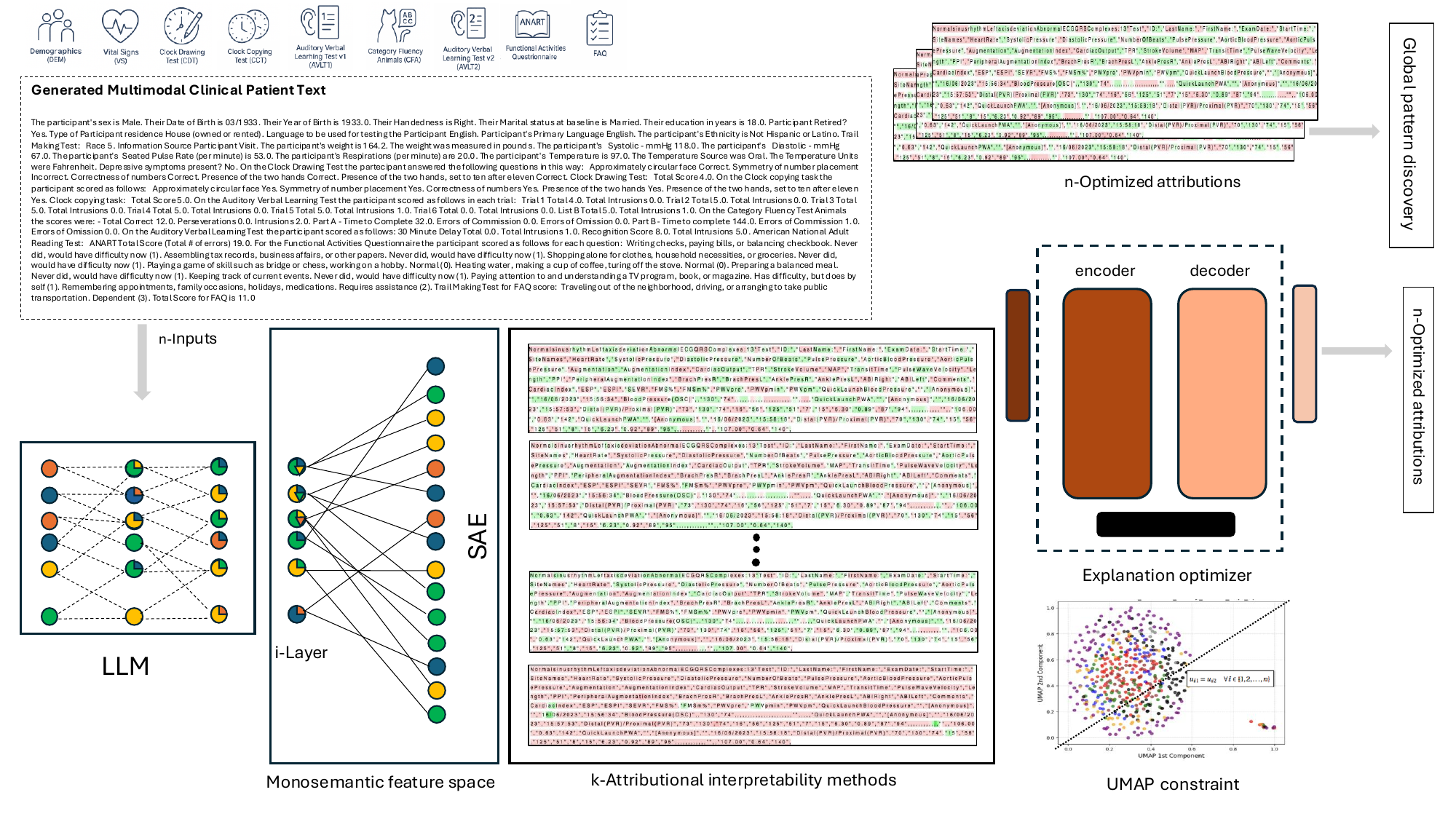}
\caption{Proposed interpretability framework for LM in Alzheimer’s diagnosis. The model integrates k-attributional methods with a SAE to generate a monosemantic feature space. An explanation optimizer refines attribution outputs, enhancing clarity and reducing variability. Global explanation quality is visualized and assessed using UMAP and a linear meta-rule, supporting both individual prediction interpretability and cohort-level pattern discovery.}
\label{abst1}
\end{figure*}

To address these challenges and produce stable, clinically meaningful explanations from large language models, we propose a unified monosemantic attribution framework that integrates both attributional and mechanistic interpretability approaches (Figure~\ref{abst1}). Our approach first employs sparse autoencoders (SAEs) to transform polysemantic LLM activations into a more monosemantic latent space, where individual features are encouraged to correspond to disentangled semantic factors. This bottleneck substantially reduces representational complexity and enables classical attribution methods to assign more precise and semantically coherent importance scores. For the attribution component, we apply six established techniques—Feature Ablation, Layer Activations, Layer Conductance, Layer Gradient SHAP~\cite{lundberg2017unified}, Layer Integrated Gradients~\cite{int}, and Layer Gradient$\times$Activation—both in the native activation space (polysemantic features) and in the SAE-induced latent space (more monosemantic features). 

To address the inter-method variability problem~\cite{mamalakis}, we combine the resulting attribution vectors and introduce the \emph{Transformer Explanation Optimizer (TEO)}, a learning-based optimization mechanism that selects explanations with maximal alignment to model behavior and dataset-level consistency~\cite{mamalakis}. We experiment with encoder–decoder architectures based on 1D transformers (TEO) and diffusion networks (DEO). For meta-level assessment and visualization, we embed these optimized attributions into a 2D manifold using UMAP~\cite{umap} and principal component analysis (PCA)~\cite{pca}, and impose a global coherence constraint by evaluating their linear structure along the primary UMAP components. This geometry-aware constraint acts as an additional regularizer, imposing a global linear structure along the dominant statistical directions of the explanation space (the first two principal components from PCA). By doing so, the principal global feature patterns exhibit similar behavior across the majority of the cohort, facilitating the extraction of robust, population-level patterns.

\section{Hypothesis and Evaluation Protocol}
The central hypothesis of this work is that \textit{as the embedding layer approaches a more monosemantic representation, attribution scores become more stable, less complex, and more diagnostically informative compared to those derived from polysemantic features}. This hypothesis is evaluated through a series of experiments in which different LMs are trained and evaluated on ADNI~\cite{adni} (Independent and Identically Distributed; \textbf{IID}) and subsequently tested on BrainLAT~\cite{brainlat} (Out-of-Distribution; \textbf{OOD}) to assess robustness under demographic and protocol shifts. We study both \textbf{binary classification} (Control vs.\ Alzheimer’s disease (AD)) and \textbf{three-class classification} settings (Control, Early Mild Cognitive Impairment (EMCI), and Late Mild Cognitive Impairment (LMCI) in ADNI; Control, Frontotemporal Dementia (FTD), and AD in BrainLAT). 

The data are partitioned into 80\% training and 20\% validation within the training portion (which itself constitutes 80\% of the full dataset), with the remaining 20\% held out for testing. The last-layer embeddings of the LM achieving the strongest baseline performance are subsequently used to evaluate the proposed interpretability framework. Importantly, both the SAE bottleneck architectures (TopK, JumpReLU, and Gated-SAEs) and TEO (with and without the SAE bottleneck) are trained exclusively on \textbf{IID} explanation cohorts using only these training and validation splits and are then evaluated on the \textbf{OOD} dataset \textbf{without any additional training or adjustments}, enabling a rigorous assessment of out-of-distribution robustness under strict generalization conditions.

\section{A Monosemantic Attribution Framework}
\label{meth}
\subsection{Attributional Theory and Methods}
Attribution explainability methods follow the framework of additive feature attribution, where the explanation model \( g(f, \mathbf{x}) \) is represented as a linear function of simplified input features:

\begin{equation}
    g(f, \mathbf{x}) = \phi_0 + \sum_{i=1}^M \phi_i x_i
    \label{eq:additive}
\end{equation}
Here, \( f \) is the predictive model, \( \phi_i \in \mathbb{R} \) is the attribution (importance) assigned to feature \( x_i \), and \( M \) is the number of simplified input features (here 512). 

For this study, we employed six well-established attributional interpretability methods applied to Transformer-Based language models (LMs), denoted as \( K = 6 \): \textit{Feature Ablation}, \textit{Layer Activations} (which capture the embedding activation space of a specific layer of interest within the LM), \textit{Layer Conduction}, \textit{Layer Gradient-SHAP} \cite{shap}, \textit{Layer Integrated Gradients} \cite{int}, and \textit{Layer Gradient $\times$ Activation} (For analytical mathematic formulation see Appendix A.1.). 
To align these layer-wise interpretability methods with the additive feature attribution framework, we reinterpret the internal activations (i.e., latent units) of a network layer \( L \) as simplified input features. The objective is to estimate an attribution score \( \phi_i \) for each unit, where \( \phi_i \in \mathbb{R} \) quantifies the contribution of the corresponding neuron to the model’s prediction.

All attribution methods were applied to the final (22nd) layer of the \textsc{Modern-BERT} LM—the model variant that achieved the highest classification accuracy in our evaluations (see Supplementary section ~1.2).
These formulations allow us to ground various neural attribution techniques within a unified additive explanation model, facilitating their comparison and hybridization under shared theoretical assumptions. 

\subsection{Attributional Explanation Optimizer Framework}
Let \( \mathcal{A} = \{A_1, A_2, \dots, A_K\} \) denote the set of \( K = 6 \) attribution methods applied to the final layer \( L \) of the model \( f \). Each method \( A_k \) generates an attribution vector \( \boldsymbol{\phi}^{(k)} = [\phi^{(k)}_1, \phi^{(k)}_2, \dots, \phi^{(k)}_M] \), where \( M \) is the number of latent features (neurons) in layer \( L \). The goal is to derive a unified attribution vector \( \bar{\boldsymbol{\phi}} \) that captures the consensus explanation across methods.
Each attribution vector \( \boldsymbol{\phi}^{(k)} \) is evaluated using the Relative Input Stability (RIS), Relative Output Stability (ROS) \cite{ros}, and Sparseness \cite{spar} metrics (Appendix A.2.).
The weighted average attribution vector \( \bar{\boldsymbol{\phi}} \) serves as the target explanation for the optimization process and it is calculated as:
\begin{equation}
\bar{\boldsymbol{\phi}} = \sum_{k=1}^K w_k \cdot \boldsymbol{\phi}^{(k)}
\end{equation}
An encoder–decoder model is trained to generate a reconstructed explanation \( \hat{\boldsymbol{\phi}} \) from the original input \( \boldsymbol{x} \). Two architectures are considered the Diffusion UNet1D \cite{unet} (\textit{Diffusion Explanation Optimizer}, DEO) and the x-transformer autoencoder \cite{att,self} (\textit{Transformer Explanation Optimizer}, TEO). For the analytical mathematical formulation, see Appendix A.2.4.
As previously highlighted, the reconstruction of the optimal explanation and the associated cost function adhere to the same principles and architectural design outlined in \cite{mamalakis}. The cost function consists of three key components: sparseness, as defined in \cite{spar}; ROS and RIS scores \cite{ros}; and similarity. The integration of these components ensures a robust and interpretable evaluation. 
The total cost function for training the reconstruction model is:
\begin{align}
\mathcal{L}_{\text{total}}(\boldsymbol{\phi}^{(k)}, \hat{\boldsymbol{\phi}}) = & 
\lambda_1 \cdot \frac{1}{M_{\text{RIS}}(f, \hat{\boldsymbol{\phi}})} + 
\lambda_2 \cdot \frac{1}{M_{\text{ROS}}(f, \hat{\boldsymbol{\phi}})} \notag \\
& + \lambda_3 \cdot M_{\text{sparse}}(f, \hat{\boldsymbol{\phi}}) + 
\lambda_4 \cdot \mathcal{L}_{\text{similarity}}(\hat{\boldsymbol{\phi}}, \bar{\boldsymbol{\phi}})
\end{align}
where:\( \lambda_1, \lambda_2, \lambda_3, \lambda_4 \) are hyperparameters controlling the influence of each loss term. This formulation enables a principled and quantitative integration of multiple attribution methods, optimizing toward a robust and interpretable explanation.

\subsection{UMAP Linear Constraint}
To obtain a unified and comparable low-dimensional representation of attribution scores
across all tokenizer features, we apply a feature-wise UMAP projection to the normalized
attribution tensor of shape $\mathbb{R}^{M \times T}$, where $M$ is the number of test
samples and $T$ denotes the dimensionality of the tokenizer embedding space. For each
feature $j$, the attribution vector $\mathbf{x}^{(j)} \in \mathbb{R}^{M}$ is min--max
normalized and embedded using a one-dimensional UMAP transformation into a two-dimensional
space $\mathbf{y}^{(j)} \in \mathbb{R}^{M \times 2}$. The resulting coordinates are
normalized to $[0,1]$ so that all feature-wise embeddings share a common bounded range.
This nonlinear projection preserves the local neighborhood structure of each attribution
distribution while mapping all $T$ tokenizer features into an aligned and directly
comparable representation space across attribution methods (Appendix A.4).

\textbf{Linear Constraint on UMAP Embeddings:}
To encourage structural consistency in the UMAP \cite{umap} embeddings, we introduce a linear
constraint requiring equality between the first and second embedding components for each
point, expressed as $u_{i1} = u_{i2}$. This can be written equivalently as
$u_{i1} - u_{i2} = 0$ for all $i$. We incorporate this constraint into the overall
optimization objective via a penalty term,
\[
\lambda_5 \sum_{i=1}^{n} (u_{i1} - u_{i2})^2,
\]
where $\lambda_5$ controls the strength of the constraint. This formulation enforces
consistency of the reconstructed embeddings while preserving flexibility in the learned
attribution representation.

\subsection{The SAE Approach and Architectures}
The mathematical formulation situates SAE architectures within the theoretical framework of superposition and semantic disentanglement (for an analytical mathematical formulation, see Appendix A.5). By expressing hidden states as sparse linear combinations of interpretable features, SAEs bridge the gap between low-level activations and human-understandable concepts. 
Let $\mathbf{x} \in \mathbb{R}^d$ denote a layer's neuron activation vector in a pretrained model. A Sparse Autoencoder learns a sparse feature representation $\mathbf{a} \in \mathbb{R}^F$ such that:
\begin{equation}
\hat{\mathbf{x}} = W \mathbf{a} + \mathbf{b}, \label{eq:reconstruction2}
\end{equation}
where $W \in \mathbb{R}^{d \times F}$ is the decoder (dictionary) matrix and $\mathbf{b} \in \mathbb{R}^d$ is a learned bias term. Each column $W_{\cdot,i}$ represents the direction of feature $i$ in neuron space, and $a_i$ is its activation. This linear mapping enables complex activations to be expressed as combinations of more interpretable features.
If $F > d$, then the feature space is overcomplete, and $W$ cannot be full-rank. This leads to superposition, where multiple features overlap in the same subspace, and individual neurons encode multiple unrelated concepts \cite{elhage2022superposition}.
If $W$ is invertible and aligned to a basis, each neuron corresponds to a single feature. The representation is monosemantic and disentangled \cite{olah2020zoom}.
When $W$ has overlapping columns, neurons can respond to multiple features, yielding polysemantic behavior. That is, for some $j$, $x_j = \sum_i W_{j,i} a_i$ involves multiple nonzero terms \cite{bills2023language}.  Variants of SAEs like TopK, JumpReLU, and Gated-SAEs offer increasingly precise control over the mapping between low-level activations and human-understandable concepts, enabling fine-grained analysis and intervention (analytical mathematical formulation, see Appendix A.6.).

\subsection{Attribution from Sparse Feature Space to Input Tokens}
Let \( \mathbf{x}_{\text{input}} \in \mathbb{R}^{d_{\text{input}}} \) 
denote the input embedding vector (e.g., LM token embeddings), 
\( \mathbf{x} = f(\mathbf{x}_{\text{input}}) \in \mathbb{R}^d \) 
the hidden layer activation of the LM, 
\( \mathbf{a} = \text{Encoder}(\mathbf{x}) \in \mathbb{R}^F \) 
the SAE sparse feature vector, and 
\( \hat{\mathbf{x}} = W \mathbf{a} + \mathbf{b} \) 
the reconstructed activation from the SAE decoder. Now suppose we have a sparse attribution vector \( \psi_i \) over features \( \mathbf{a} \), i.e., \( \psi \in \mathbb{R}^F \), where each \( \psi_i \) reflects the importance of SAE feature \( a_i \). We aim to assign importance \( \Phi_k \) to each input token dimension \( x_{\text{input},k} \).

We propagate the feature attributions backward through the encoder to the input. Using the chain rule:

\begin{equation}
\Phi_k = \sum_{i=1}^F \psi_i \cdot \frac{\partial a_i}{\partial x_{\text{input},k}} = \sum_{i=1}^F \psi_i \cdot \frac{\partial a_i}{\partial \mathbf{x}} \cdot \frac{\partial \mathbf{x}}{\partial x_{\text{input},k}}
\label{eq:backprop-attribution}
\end{equation}
where \( \frac{\partial a_i}{\partial \mathbf{x}} \) is the encoder Jacobian (SAE layer), and \( \frac{\partial \mathbf{x}}{\partial x_{\text{input},k}} \) is the LM gradient from input token to hidden layer.
This gives us a scalar attribution \( \Phi_k \in \mathbb{R} \) for each token/input embedding dimension \( k \).
This represents how much each input token contributes to the sparse SAE features identified as important. In doing so, we assess the contribution of input features based on the monosemantic behavior of the trained network’s internal mechanisms. Building on our study, we apply the six previously discussed attribution methods at two levels: from the SAE feature space to the encoder layer, and from the encoder layer to the input embedding space. This dual-level attribution analysis enables us to investigate how interpretable sparse features relate to model internals and ultimately shape the input-level representations. Attribution methods (e.g., Integrated Gradients, SHAP) can directly estimate:
\begin{equation}
\boldsymbol{\phi}^{\text{input}} = \text{AttributionMethod}(f, \mathbf{x}_{\text{input}}, \boldsymbol{\phi}^{\text{SAE}})
\end{equation}
where \( \boldsymbol{\phi}^{\text{SAE}} \) denotes the monosemantic feature space of the SAE network.
Thus, the dual-level approach allows us to connect semantically meaningful sparse features to the raw input representation space.

\subsection{LM Networks and Hyperparameter Tuning}
We evaluate encoder-based LMs (BERT, RoBERTa, DistilBERT, ALBERT, BioBERT, ModernBERT) on \textsc{ADNI} (IID) and \textsc{BrainLat} (OOD) under a unified protocol spanning full fine-tuning, zero-shot, few-shot with temperature control, and parameter-efficient LoRA. \textit{ModernBERT} outperformed all other networks on \textsc{ADNI}: in the binary task it achieved the highest F1 (75.89\%), AUC-PR (86.41\%), ROC-AUC (83.95\%), and Accuracy (72.37\%), and it remained strongest in the three-class setting (F1 68.80\%, AUC-PR 78.48\%, ROC-AUC 78.67\%, Accuracy 65.05\%). For OOD model adaptation, \textit{ModernBERT} zero-shot yielded ~55\% Accuracy, few-shot/LoRA provided modest gains (62\%), while full fine-tuning peaked at $\sim$84\% Accuracy but lies outside our scope (Supplementary 1.2). Accordingly, we use \textit{ModernBERT} fine-tuned on \textsc{ADNI} for IID and zero-shot on \textsc{BrainLat} for OOD, and all interpretability analyses are conducted on the 22\textsuperscript{nd} layer of \textit{ModernBERT}. We conducted extensive hyperparameter tuning for all components. The explanation optimizer performed best at a learning rate of 2e\textsuperscript{-4}, with the optimal weight configuration $(\lambda_{1}, \lambda_{2}, \lambda_{3}, \lambda_{4}) = (0.1, 0.3, 0.1, 0.5)$. UMAP constraints were most effective at the 4$\times$ batch size level (4$\times$64) (Supplementary 1.3; Figures 3). Among the four SAE variants, TopK produced the strongest overall performance (Supplementary 1.3; Figures 2, 4, 5), using a 32× feature depth. All models were trained using the AdamW optimizer with early stopping and standard evaluation metrics (Supplementary 1.3).
\subsection{Dataset and Code Availability}
The dataset used in this study originates from the ADNI cohort~\cite{adni} and is represented as text generated from multiple modalities, serving as the in-distribution (IID) dataset. We further split the generated text into nine subgroups based on input modality, as detailed in Supplementary Material~1.1.5, for pattern analysis and biomarker research purposes. For the out-of-distribution (OOD) cohort, we used text generated from multimodal sources (MRI and clinical files) in the Latin American Brain Health Institute (BrainLat) dataset, a multi-site initiative providing neuroimaging, cognitive, and clinical data across several Latin American countries~\cite{brainlat}. Additional demographic and preprocessing details are provided in Supplementary Sections~1.1.1–1.1.5.
The code is implemented in Python using PyTorch and runs on an NVIDIA cluster in one A100-SXM-80GB GPU. It leverages \texttt{SAE\_LENs}~\cite{sael} for SAE training, \texttt{quantus}~\cite{quan} for evaluation, and \texttt{captum} for attribution analysis. 

\section{Results of Experiments}
\subsection{Ablation Analysis of the Monosemantic Attribution Framework}
In this study, sparseness is defined such that higher values correspond to more selective and concentrated attributions across input features—that is, greater sparseness. However, sparseness alone is insufficient to assess explanation quality, as it does not account for robustness or stability. Therefore, the most effective explanation method is one that simultaneously achieves high sparseness and low RIS and ROS values.

Across IID, OOD datasets and for both binary and three-class classification tasks, Table~\ref{sample-table} shows a consistent stability–sparsity frontier governed by the proposed optimizers. In the binary IID case, SAE substantially improves stability for feature-learning explainers, most notably: Layer Conductance and especially TEO, with large drops in RIS/ROS for both Alzheimer’s and Control, whereas Activation–SAE increases RIS/ROS relative to its no-SAE variant and is therefore less robust. In the binary OOD case, this pattern persists and strengthens: TEO–SAE achieves the lowest RIS/ROS overall (strong cross-dataset stability), while TEO–UMAP recovers higher sparsity ($>0.40$) at a modest stability cost versus TEO–SAE, offering a tunable sparsity–stability trade-off. In the three-class IID setting, Feature Ablation is the sparsity leader across Control/LMCI/MCI ($\sim0.52$–$0.53$) with moderate, steady yet still high RIS/ROS values; Layer Conductance–SAE markedly reduces RIS/ROS for LMCI/MCI; and TEO–SAE again delivers the most stable attributions across all classes, albeit with reduced sparsity compared with no-SAE variants. The same rank ordering holds OOD. Across all blocks, gradient-formulaic methods (Grad-SHAP, Guided Backprop, Integrated Gradients) show near-invariant RIS/ROS ($\sim5.6/\sim16.93$) regardless of SAE, class, or domain, indicating that SAE chiefly benefits learned-attribution methods. Further analyses are provided in Supplementary §1.4, Tables 2–5.

\begin{table*}
\caption{
Unified attribution performance across tasks and evaluation settings.
Values report mean$\pm$std per class.
Binary tasks use (A/C); three-class tasks use (L/M/C) and (A/F/C).
Bold indicates the best performance within each setting.
}
\label{sample-table}
\centering
\scriptsize
\setlength{\tabcolsep}{3pt}
\renewcommand{\arraystretch}{1.10}
\resizebox{1.0\textwidth}{!}{
\begin{tabular}{llccc}
\toprule
\textbf{Task \& Setting} & \textbf{Method (variant)} & \textbf{Sparseness} & \textbf{RIS} & \textbf{ROS} \\
\midrule
\multicolumn{5}{l}{\textbf{Binary — IID (ADNI)} \quad Columns: (A/C)}\\
\midrule
& Gradient Activation
& \res{0.3277}{0.0384}/\res{0.2500}{0.0230}
& \res{5.6149}{0.0193}/\res{5.6170}{0.0221}
& \res{16.9303}{0.0034}/\res{16.9347}{0.0} \\

& Gradient Activation-SAE
& \res{0.2035}{0.0117}/\res{0.1668}{0.0072}
& \res{5.6252}{0.0213}/\res{5.6173}{0.0221}
& \res{16.9343}{6.8e-5}/\res{16.9347}{4.0e-5} \\

\addlinespace[1pt]
& DEO
& \res{0.3383}{0.0033}/\res{0.3377}{0.0017}
& \res{9.2839}{0.0800}/\res{9.3131}{0.1427}
& \res{20.6342}{0.0866}/\res{20.6159}{0.2026} \\

& DEO-SAE
& \res{0.3374}{0.0029}/\res{0.3140}{0.0010}
& \res{9.2790}{0.0646}/\res{9.1750}{0.1088}
& \res{20.6150}{0.0880}/\res{20.5150}{0.1299} \\

\addlinespace[1pt]
& TEO
& \textbf{\res{0.4220}{0.0003}/\res{0.4199}{0.0005}}
& \res{5.0520}{0.0192}/\res{5.0688}{0.0184}
& \res{16.3529}{0.0056}/\res{16.3777}{0.0011} \\

& TEO-SAE
& \res{0.2672}{0.0010}/\res{0.2682}{0.0007}
& \textbf{\res{1.6227}{0.1708}/\res{0.9964}{0.2639}}
& \textbf{\res{12.9250}{0.1703}/\res{12.2983}{0.2613}} \\

& TEO-UMAP (SAE)
& \res{0.3989}{0.0004}/\res{0.4057}{0.0003}
& \res{5.4394}{0.0332}/\res{5.4709}{0.1746}
& \res{16.3037}{0.0033}/\res{16.2102}{0.0079} \\

\midrule
\multicolumn{5}{l}{\textbf{Binary — OOD (BrainLAT)} \quad Columns: (A/C)}\\
\midrule
& Gradient Activation-SAE
& \res{0.1140}{0.0177}/\res{0.0630}{0.0069}
& \res{6.0328}{0.0277}/\res{6.0339}{0.0398}
& \res{16.9347}{3.6e-6}/\res{16.9348}{3.7e-5} \\

\addlinespace[1pt]
& TEO-SAE
& \res{0.2691}{0.0016}/\res{0.2725}{0.0004}
& \textbf{\res{0.6835}{0.6676}/\res{0.4734}{0.2801}}
& \textbf{\res{11.5236}{0.6591}/\res{11.2130}{0.5150}} \\

& TEO-UMAP (SAE)
& \textbf{\res{0.3989}{0.0005}/\res{0.4043}{0.0029}}
& \res{5.4394}{0.0383}/\res{5.4282}{0.1944}
& \res{16.3037}{0.0039}/\res{16.1577}{0.1054} \\

\midrule
\multicolumn{5}{l}{\textbf{Three-class — IID (ADNI)} \quad Columns: (L/M/C)}\\
\midrule
& Gradient Activation
& \res{0.3839}{0.0177}/\res{0.2917}{0.0200}/\res{0.2697}{0.0061}
& \res{5.6272}{0.0212}/\res{5.6269}{0.0193}/\res{5.6290}{0.0225}
& \res{16.9339}{0.0008}/\res{16.9340}{0.0006}/\res{16.9347}{0} \\

& Gradient Activation-SAE
& \textbf{\res{0.4310}{0.1156}/\res{0.2579}{0.1095}/\res{0.2296}{0.0036}}
& \res{5.6297}{0.9478}/\res{5.6172}{1.1684}/\res{5.6210}{0.0194}
& \res{16.9347}{2.8625}/\res{16.9347}{3.5241}/\res{16.9348}{1.22e-5} \\

\addlinespace[1pt]
& TEO
& \res{0.4131}{0.0003}/\res{0.3909}{0.0047}/\res{0.3918}{0.0008}
& \res{5.0938}{0.0188}/\res{4.8283}{0.0377}/\res{4.8080}{0.0184}
& \res{16.4043}{0.0024}/\res{16.1354}{0.0324}/\res{16.1172}{0.0090} \\

& TEO-SAE
& \res{0.2860}{0.0374}/\res{0.2838}{0.0523}/\res{0.2682}{0.0649}
& \textbf{\res{2.2642}{0.4877}/\res{2.1617}{0.4547}/\res{1.5468}{0.1171}}
& \textbf{\res{13.5646}{2.2745}/\res{13.4676}{2.7641}/\res{12.8570}{0.1179}} \\

& TEO-UMAP (SAE)
& \res{0.4161}{0.0870}/\res{0.4172}{0.2372}/\res{0.3973}{0.0749}
& \res{5.1017}{0.1697}/\res{5.1116}{0.1072}/\res{5.1086}{0.2083}
& \res{16.4031}{3.8616}/\res{16.4088}{0.4924}/\res{16.4123}{6.8439} \\

\midrule
\multicolumn{5}{l}{\textbf{Three-class — OOD (BrainLAT)} \quad Columns: (A/F/C)}\\
\midrule
& Gradient Activation-SAE
& \res{0.1836}{0.0016}/\res{0.4303}{0.0022}/\res{0.1772}{0.0112}
& \res{6.0269}{0.0302}/\res{6.1450}{0.1210}/\res{6.1234}{0.1912}
& \res{16.9348}{2.6e-6}/\res{16.9345}{2.8e-5}/\res{16.9346}{1.5e-5} \\

\addlinespace[1pt]
& TEO-SAE
& \res{0.3716}{0.0009}/\res{0.4224}{0.0002}/\res{0.4162}{0.0029}
& \textbf{\res{4.9396}{0.0148}/\res{5.5421}{0.0611}/\res{5.7520}{0.3645}}
& \textbf{\res{15.8121}{0.0099}/\res{16.2773}{0.0010}/\res{16.3792}{0.0034}} \\

& TEO-UMAP (SAE)
& \textbf{\res{0.4239}{5.0e-5}/\res{0.4246}{2.4e-4}/\res{0.4238}{5.6e-4}}
& \res{5.4583}{0.0297}/\res{5.5576}{0.0954}/\res{5.5525}{0.1862}
& \res{16.3726}{0.0017}/\res{16.3572}{0.0035}/\res{16.3661}{0.0073} \\

\bottomrule
\end{tabular}
}
\end{table*}

\begin{figure}
\centering 
\includegraphics[width=0.9\textwidth]{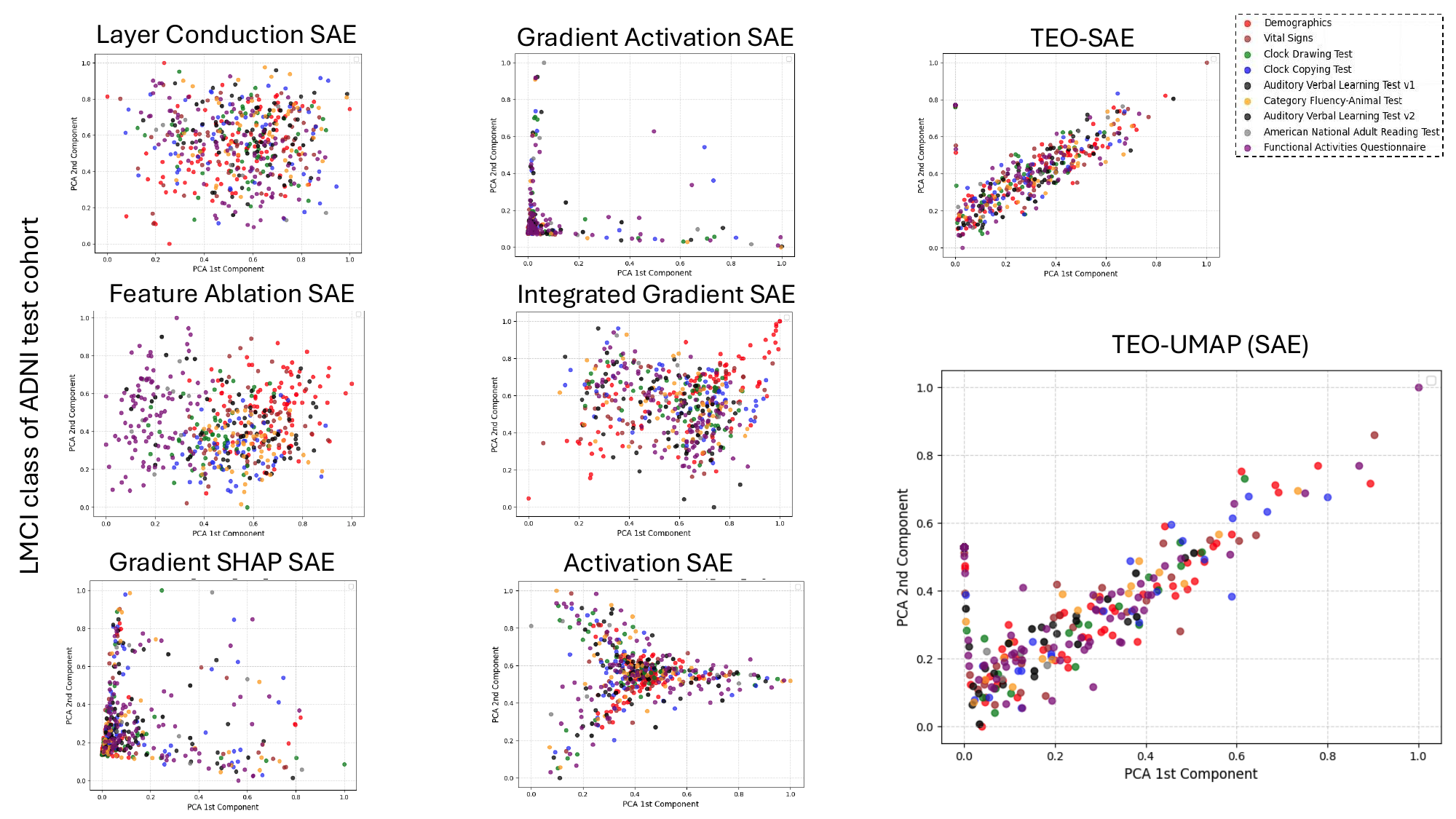}
\caption{PCA of token-level attribution representations (LMCI, ADNI test set).
The first two principal components are shown, computed from the top eight PCA components of the attribution matrix.}
\label{ar12sss}
\end{figure}

To validate that the proposed UMAP linear constraint effectively linearizes the attribution space and yields robust scores within tokenized features, we performed an additional PCA analysis. We extracted the top eight principal components and visualized the first two, which capture the majority of variance in the cohort distribution. Under the UMAP constraint, these components reveal an approximately linear structure along the dominant statistical directions (Figure \ref{ar12sss}), inducing a coherent global organization of the feature space and promoting consistent population-level attribution patterns across subjects (Appendix B.3).

\subsection{Individual- and Cohort-Level Quality of Explanations}
\begin{figure*}
\centering 
\includegraphics[width=0.9\textwidth]{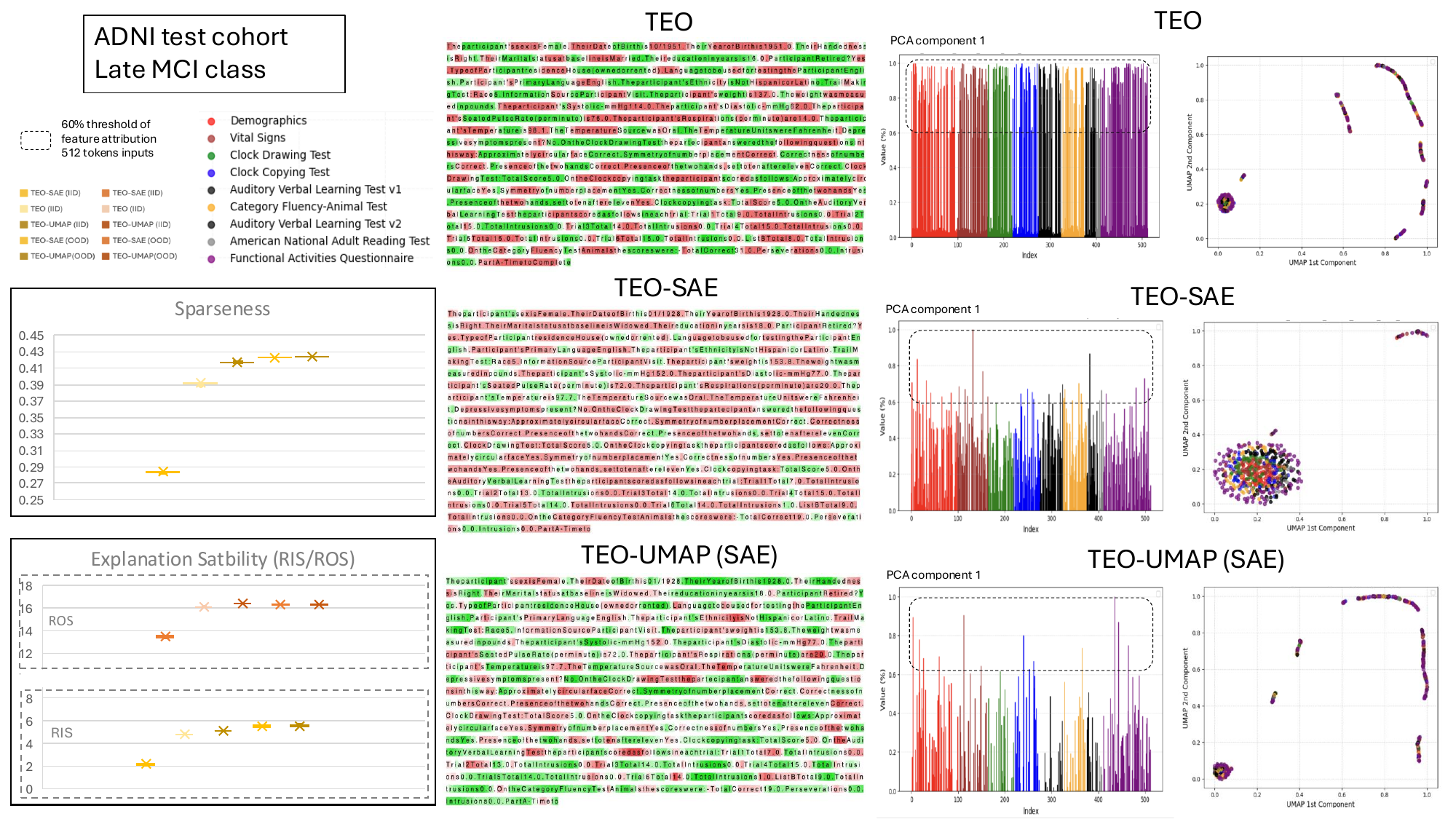}
\caption{{\textcolor{black}{\textbf{Left:} Stability–sparsity frontier for explanation optimizers on ADNI (Late MCI) in the testing cohort. Scatter points show TEO, TEO–SAE, and TEO–UMAP (SAE) on ADNI (IID) and BrainLat (OOD). Metrics: Sparseness (higher is better) vs.\ RIS/ROS stability (lower is better).  
\textbf{Middle:} Token-level heatmap produced by the proposed framework, with feature-attribution scale (green: positive relevance; red: negative relevance; white: neutral).  
\textbf{Right:} $1^\text{st}$ PCA and 2D UMAP projections of the full testing cohort. Thresholding uses 60\% feature attribution over 512 tokens. The generated input text is split into nine subgroups (different colours) based on input modality, as detailed in Supplementary Material~1.1.5, for pattern analysis and biomarker identification.}}}
\label{abst}
\end{figure*}

Figure~\ref{abst} presents qualitative local-level and cohort-level attributions for the LMCI class in the three-class classification task on both IID and OOD settings using our proposed optimisers TEO, TEO-SAE, and TEO-UMAP (SAE). At the local level, the heatmaps illustrate feature importance, with tokens colour-coded to indicate relevance (green = positive relevance; red = negative relevance). Visually, TEO-SAE produces the tightest, least noisy explanations—fewer spurious highlights and clearer token groupings—consistent with its lowest RIS/ROS in Table~\ref{sample-table}. Adding the UMAP constraint restores higher sparseness while preserving much of TEO-SAE  stability: TEO-UMAP (SAE) yields compact, well-separated patterns that remain clinically interpretable across IID and OOD. Across classes and datasets, higher Sparseness corresponds to less diffuse maps with balanced positive/negative highlights, whereas low Sparseness with high RIS/ROS manifests as saturated red/green patches and unstable saliency (see Supplementary Figures 7–16). Among the six classical attribution methods (Activation, Layer Conductance, Feature Ablation, Gradient SHAP, Gradient Activation, Integrated Gradient), Feature Ablation attains the highest Sparseness but exhibits poorer stability (elevated RIS/ROS), a gap that worsens with SAE due to decoder-driven “decompression” (Supplementary Figures  7–8). Layer Conductance shows the opposite trade-off: SAE reduces Sparseness but improves stability (lower RIS/ROS), with similar stability gains observed for Gradient Activation, Integrated Gradient, and Gradient SHAP (Supplementary Figures 9–10). Overall, none of these classical methods match the proposed framework as TEO-SAE is consistently most stable, and TEO-UMAP (SAE) offers a tunable sparsity–stability compromise that generalises from ADNI to BrainLat (for extended analyses see Supplementary §1.6).

At the cohort level, we extracted UMAP embeddings to visualise patterns and text–category clusters in 2D space, observing any spreading effects or homogeneous clustering. We also applied PCA to identify high-contributing features (threshold 0.6 on the first component; Figure~\ref{abst}). Moving from TEO to TEO–SAE produces tighter, more homogeneous low-to-high attribution and the lowest RIS/ROS (highest stability), but also a marked reduction in sparseness, evident as broader token spread in the 2D manifold (see Figure~\ref{abst}). In some cases, this stabilisation concentrates signals so strongly that few features exceed the significance threshold (square box in the 1D scatter plot where PCA first component $\geq 0.6$), and not all subgroups are represented (Figure~\ref{abst}, Supplementary Figure 32). Imposing a linear UMAP constraint (TEO–UMAP) mitigates this effect by restoring sparsity in significant attributions while retaining stability, yielding compact, clinically interpretable maps with more uniform subgroup coverage (Figure~\ref{abst}; Supplementary Figures 33–35). The behaviour of the proposed framework (TEO, TEO–SAE, TEO–UMAP) shows that higher sparseness corresponds to less diffuse, more balanced highlights, whereas lower sparseness with higher RIS/ROS results in saturated red/green patches. This mirrors the patterns observed across the six classical methods (Activation, Layer Conduction, Feature Ablation, Gradient SHAP, Gradient Activation, Integrated Gradient; Apendix A.2). With SAE, feature-learning explainers such as Layer Conduction generally gain stability (lower RIS/ROS) at some sparsity cost, while Feature Ablation maintains high sparsity but remains unstable. None, however, match the stability–sparsity trade-off achieved by TEO–SAE and TEO–UMAP (box plots in Figure~\ref{abst}; Table~\ref{sample-table}). Supplementary §1.7 (Figures 22–31) provides more details about cohort-level attributions.
\subsection{Are Monosemantic Representation--Based Attribution Methods Statistically Distinct from Standard Attribution Techniques?}
A statistical evaluation of interpretability metrics (sparseness, RIS, and ROS, see Supplementary~\S 1.5) across methods with and without the SAE layer was computed. In the binary ADNI task, \textit{paired $t$-tests} with FDR correction showed that adding an SAE bottleneck significantly reduced Complexity ($p<10^{-10}$) and RIS ($p<10^{-4}$) in both groups, while ROS changes were small and inconsistent (marginal for Control, non-significant for Alzheimer’s). The strongest SAE effects appeared in attribution metrics, with Gradient SHAP ($p<10^{-45}$), Layer Conduction ($p=3.2\times10^{-7}$), Integrated Gradients ($p<10^{-55}$), and the TEO ($p<10^{-95}$) all showing decisive reductions, confirming robust stability gains under SAE. In the three-class ADNI task, \textit{paired $t$-tests and Wilcoxon signed-rank tests} (BH-FDR) indicated that the MCI group showed the clearest improvement: ROS decreased strongly ($t(17)=-10.12,\; p=1.30\times10^{-8};\; W=0,\; p=8.0\times10^{-6},\; q=2.3\times10^{-5}$), RIS showed a smaller reduction detected non-parametrically ($W=19,\; p=0.00117$), and Complexity increased modestly by Wilcoxon ($W=17,\; p=7.9\times10^{-4}$) while paired $t$-tests were non-significant. Control and LMCI had incomplete pairs, preventing matched testing with correction. Overall, SAE reliably improves attribution stability (lower RIS/ROS) and increase sparseness in binary tasks, with the three-class MCI group showing the most consistent ROS gains.

\subsection{Clinical Impact and Diagnostic Outcomes in Alzheimer’s Disease}
\textbf{Biomarker Identification:} Our findings demonstrate that both TEO-SAE and TEO-UMAP yield the most reliable and consistent identification of informative sources across the nine multimodal subgroups. Using a threshold of 0.6 on the first principal component (PCA; Figure~\ref{abst}), we observe systematic and class-specific biomarker patterns in the IID binary task (ADNI). For the Control group, TEO-SAE is primarily driven by FAQ, whereas TEO-UMAP shifts emphasis toward DEM, AVLT2, and FAQ. For the Alzheimer’s group, TEO prioritises FAQ, AVLT1, and CFA, while TEO-UMAP highlights ANART, FAQ, and DEM. In the three-class setting, TEO identifies AVLT1, CDT, and ANART as most influential for Controls, while TEO-UMAP selects AVLT1, CDT, and CFA. For MCI, TEO assigns highest relevance to CCT, AVLT2, and FAQ, whereas TEO-UMAP favours AVLT2, ANART, and CFA. For LMCI, TEO elevates AVLT1, FAQ, and CDT, while TEO-UMAP elevates FAQ, ANART, and AVLT2. These trends are summarised in Appendix Table~2 (Section~B.2), with acronym definitions provided in Sections~1.1.5 (Table 1).

Importantly, these results illustrate how our proposed mechanistic attribution framework can disentangle which demographic/vital features and, crucially, which cognitive assessments contribute most strongly to each diagnostic category. This is particularly relevant in clinical neuroscience, where cognitive tests are often time-consuming, resource-intensive, and susceptible to demographic or cultural bias. Our framework provides a principled approach for identifying the most informative cognitive biomarkers-such as AVLT, FAQ, CDT, and ANART-which have been repeatedly linked to early Alzheimer’s progression in the literature \cite{petersen1999mci}, \cite{bondi2019mci}. By pinpointing the minimal set of high-yield assessments for each classification task, the method supports more efficient screening pipelines, reduces clinical burden, and enables scalable application in larger cohorts where repeated or comprehensive testing is impractical.

\section{LM-Based External Semantic Evaluation of Attribution-Derived Tokens from the Proposed Framework}
\textbf{Setup.} To assess the clinical relevance of our attribution framework and, in particular, the contribution of the SAE layer in producing monosemantic and diagnostically coherent explanations, we performed an auxiliary evaluation across three cohorts (Binary ADNI, Binary BrainLAT, and three-class ADNI). For each test sample, we extracted the top 50\% most influential token attributions from TEO without SAE, TEO with SAE, and TEO-UMAP, and constructed class-specific CSV files in which highlighted characters for each attribution method were arranged column-wise, with the full character sequence (including the CLS token) provided in the first column to ensure clear sample-level distinction. These CSVs were then presented to a large language model (ChatGPT-5.1~\cite{gpt}) under a fixed prompting protocol to yield an external, model-agnostic assessment of the interpretability structure encoded by each explanation space. The goal of this experiment is to assess whether the selected high-importance tokens preserve sufficient clinically meaningful structure to support consistent interpretation. While language models can infer meaning from partial inputs, this evaluation does not assume faithful reconstruction; instead, it measures the relative interpretability of different explanation methods.

\textbf{Results.} Across both binary tasks, all three attribution methods correctly identified the pathological Alzheimer’s class; however, TEO without SAE consistently fixated on task artefacts (e.g., instruction counts or label-like patterns) rather than clinically meaningful biomarkers, whereas TEO-SAE and TEO-UMAP surfaced coherent neurocognitive indicators, demographic risk factors, and processing-speed impairments. In the more challenging three-class ADNI setting, the SAE-enabled variants again produced clearer diagnostic separation and more structured biomarker profiles, while TEO without SAE failed to identify pathology or highlight relevant clinical features. These results show that using an SAE-driven monosemantic representation substantially strengthens the reliability, diagnostic validity, and clinical interpretability of the resulting attribution signals (see Appendix~B.1.; Figure 4).

\section{Conclusion}
We introduced a unified interpretability framework that combines monosemantic feature extraction with learning-based explanation optimization, optionally augmented with a geometry-aware constraint. Across IID and OOD settings and multiple classification tasks, the proposed methods yield more stable and robust explanations than classical attribution approaches, defining a tunable sparsity–stability trade-off that generalizes under distribution shift. Clinically, the framework identifies coherent diagnostic markers and enables low-dimensional cohort-level explanations that support efficient assessment. Overall, this approach advances robust and trustworthy interpretability for Transformer-Based LM-based Alzheimer’s disease analysis.

\subsubsection*{Impact Statement}
This paper presents work whose goal is to advance the field of machine learning. There are many potential societal consequences of our work, none of which we feel must be specifically highlighted here.
The paper discusses several potential positive societal impacts, particularly emphasizing its relevance to clinical applications such as the early diagnosis and treatment planning of Alzheimer’s Disease. By proposing a unified interpretability framework that combines attributional and mechanistic techniques, the authors aim to enhance
the trustworthiness, consistency, and human alignment of large language model (LLM) outputs. This improved interpretability is presented as a means to support safer and more effective integration of LLMs and Transformer-Based LMs into cognitive health and clinical decision-making, with
the potential to uncover clinically meaningful patterns and ultimately improve patient outcomes. However, the paper does not explicitly address possible negative societal impacts of the work. It does not discuss risks such as the misinterpretation of model ex-677
planations, over-reliance on machine-generated insights in high-stakes medical contexts, or the potential for the framework to inadvertently reinforce biases embedded in training data. Societal impacts can be better established through future work, in which we plan to incorporate clinician-in-the-loop evaluation and patients.
\subsubsection*{Acknowledgments}

This acknowledgment complies with the dataset license and does not affect the anonymization of the authors.

Data collection and sharing for the Alzheimer's Disease Neuroimaging Initiative (ADNI) is funded by the National Institute on Aging (National Institutes of Health Grant U19AG024904). The grantee organization is the Northern California Institute for Research and Education. Past funding was obtained from the National Institute of Biomedical Imaging and Bioengineering, the Canadian Institutes of Health Research, and private sector contributions through the Foundation for the National Institutes of Health (FNIH), including generous contributions from the following organizations: AbbVie; Alzheimer's Association; Alzheimer's Drug Discovery Foundation; Araclon Biotech; BioClinica, Inc.; Biogen; Bristol Myers Squibb Company; CereSpir, Inc.; Cogstate; Eisai Inc.; Elan Pharmaceuticals, Inc.; Eli Lilly and Company; EuroImmun; F. Hoffmann-La Roche Ltd and its affiliated company Genentech, Inc.; Fujirebio; GE Healthcare; IXICO Ltd.; Janssen Alzheimer Immunotherapy Research \& Development, LLC.; Johnson \& Johnson Pharmaceutical Research \& Development LLC.; Lumosity; Lundbeck; Merck \& Co., Inc.; Meso Scale Diagnostics, LLC.; NeuroRx Research; Neurotrack Technologies; Novartis Pharmaceuticals Corporation; Pfizer Inc.; Piramal Imaging; Servier; Takeda Pharmaceutical Company; and Transition Therapeutics.
\bibliography{arXiv}

@article{petersen1999mci,
  title={Mild cognitive impairment: clinical characterization and outcome},
  author={Petersen, Ronald C. and Smith, Glenn E. and Waring, Susan C. and Ivnik, Robert J. and Tangalos, Eric G. and Kokmen, Emre},
  journal={Archives of Neurology},
  volume={56},
  number={3},
  pages={303--308},
  year={1999},
  publisher={American Medical Association}
}

@article{pca,
author = { Karl   Pearson   F.R.S. },
title = {LIII. On lines and planes of closest fit to systems of points in space},
journal = {The London, Edinburgh, and Dublin Philosophical Magazine and Journal of Science},
volume = {2},
number = {11},
pages = {559-572},
year  = {1901},
publisher = {Taylor & Francis},
doi = {10.1080/14786440109462720},
}

@misc{toy,
      title={Toy Models of Superposition}, 
      author={Nelson Elhage and Tristan Hume and Catherine Olsson and Nicholas Schiefer and Tom Henighan and Shauna Kravec and Zac Hatfield-Dodds and Robert Lasenby and Dawn Drain and Carol Chen and Roger Grosse and Sam McCandlish and Jared Kaplan and Dario Amodei and Martin Wattenberg and Christopher Olah},
      year={2022},
      eprint={2209.10652},
      archivePrefix={arXiv},
      primaryClass={cs.LG},
      url={https://arxiv.org/abs/2209.10652}, 
}

@article{gornotempini2011ppa,
  title={Classification of primary progressive aphasia and its variants},
  author={Gorno-Tempini, Maria Luisa and Hillis, Argye E and Weintraub, Sandra and Kertesz, Andrew and Mendez, Mario and Cappa, Stefano F and Ogar, Jennifer M and Rohrer, Jonathan D and Black, Sandra and Boeve, Bradley F and others},
  journal={Neurology},
  volume={76},
  number={11},
  pages={1006--1014},
  year={2011},
  publisher={Wiley Online Library}
}

@article{jack2018niaaa,
  title={NIA-AA Research Framework: Toward a biological definition of Alzheimer's disease},
  author={Jack, Clifford R and Bennett, David A and Blennow, Kaj and Carrillo, Maria C and Dunn, Bruce and Haeberlein, Susan B and Holtzman, David M and Jagust, William and Jessen, Frank and Karlawish, Jason and others},
  journal={Alzheimer's \& Dementia},
  volume={14},
  number={4},
  pages={535--562},
  year={2018},
  publisher={Elsevier}
}

@article{devlin2018bert,
  title={Bert: Pre-training of deep bidirectional transformers for language understanding},
  author={Devlin, Jacob and Chang, Ming-Wei and Lee, Kenton and Toutanova, Kristina},
  journal={arXiv preprint arXiv:1810.04805},
  year={2018}
}

@article{umap,
  title={UMAP: Uniform Manifold Approximation and Projection for Dimension Reduction},
  author={McInnes, Leland and Healy, John and Melville, James},
  journal={arXiv preprint arXiv:1802.03426},
  year={2018}
}

@misc{modernbert,
      title={Smarter, Better, Faster, Longer: A Modern Bidirectional Encoder for Fast, Memory Efficient, and Long Context Finetuning and Inference}, 
      author={Benjamin Warner and Antoine Chaffin and Benjamin Clavié and Orion Weller and Oskar Hallström and Said Taghadouini and Alexis Gallagher and Raja Biswas and Faisal Ladhak and Tom Aarsen and Nathan Cooper and Griffin Adams and Jeremy Howard and Iacopo Poli},
      year={2024},
      eprint={2412.13663},
      archivePrefix={arXiv},
      primaryClass={cs.CL},
      url={https://arxiv.org/abs/2412.13663}, 
}

@inproceedings{brown2020language,
  title={Language models are few-shot learners},
  author={Brown, Tom B and Mann, Benjamin and Ryder, Nick and Subbiah, Melanie and Kaplan, Jared and Dhariwal, Prafulla and Neelakantan, Arvind and Shyam, Pranav and Sastry, Girish and Askell, Amanda and others},
  booktitle={Advances in neural information processing systems},
  volume={33},
  pages={1877--1901},
  year={2020}
}

@article{doshi2017towards,
  title={Towards a Rigorous Science of Interpretable Machine Learning},
  author={Doshi-Velez, Finale and Kim, Been},
  journal={arXiv preprint arXiv:1702.08608},
  year={2017},
  url={https://arxiv.org/abs/1702.08608}
}

@article{adni,
title = {The Alzheimer's Disease Neuroimaging Initiative},
journal = {Neuroimaging Clinics of North America},
volume = {15},
number = {4},
pages = {869-877},
year = {2005},
note = {Alzheimer's Disease: 100 Years of Progress},
issn = {1052-5149},
doi = {https://doi.org/10.1016/j.nic.2005.09.008},
url = {https://www.sciencedirect.com/science/article/pii/S1052514905001024},
author = {Susanne G. Mueller and Michael W. Weiner and Leon J. Thal and Ronald C. Petersen and Clifford Jack and William Jagust and John Q. Trojanowski and Arthur W. Toga and Laurel Beckett}
}

@ARTICLE{loniida,
AUTHOR={Neu, Scott C.  and Crawford, Karen L.  and Toga, Arthur W. },
TITLE={The image and data archive at the laboratory of neuro imaging},
JOURNAL={Frontiers in Neuroinformatics},
VOLUME={Volume 17 - 2023},
YEAR={2023},
URL={https://www.frontiersin.org/journals/neuroinformatics/articles/10.3389/fninf.2023.1173623},
DOI={10.3389/fninf.2023.1173623},
ISSN={1662-5196}}

@article{brainlat,
  author = {Prado, Pavel and Medel, Vicente and Sainz-Ballesteros, Agust{\'\i}n and Santamar{\'\i}a-Garc{\'\i}a, Hernando and Moguilner, Sebastián and Mejía, Jhony and González-Gómez, Raúl and Slachevsky, Andrea and Behrens, María Isabel and Aguillón, David and Lopera, Francisco and Parra, Mario A. and Matallana, Diana and Maito, Marcelo Adrián and García, Adolfo M. and Custodio, Nilton and Ávila Funes, Alberto and Piña-Escudero, Stefanie and Birba, Agustina and Fittipaldi, Sol and Legaz, Agustina and Ibáñez, Agustín},
  title = {The BrainLat project: a multimodal neuroimaging dataset of neurodegeneration from underrepresented backgrounds},
  journal = {Scientific Data},
  year = {2023},
  volume = {10},
  pages = {889},
  doi = {10.1038/s41597-023-02806-8},
  url = {https://www.nature.com/articles/s41597-023-02806-8}
}

@misc{lime,
      title={"Why Should I Trust You?": Explaining the Predictions of Any Classifier}, 
      author={Marco Tulio Ribeiro and Sameer Singh and Carlos Guestrin},
      year={2016},
      eprint={1602.04938},
      archivePrefix={arXiv},
      primaryClass={cs.LG},
      url={https://arxiv.org/abs/1602.04938}, 
}

@inproceedings{spar,
  title        = {Concise Explanations of Neural Networks using Adversarial Training},
  author       = {Chalasani, Prasad and Chen, Jiefeng and Chowdhury, Amrita Roy and Wu, Xi and Jha, Somesh},
  booktitle    = {Proceedings of the 37th International Conference on Machine Learning (ICML)},
  editor       = {Hal Daumé III and Aarti Singh},
  series       = {Proceedings of Machine Learning Research},
  volume       = {119},
  pages        = {1383--1391},
  publisher    = {PMLR},
  address      = {Virtual Event, online},
  month        = {July 13--18},
  year         = {2020},
  note         = {Originally released as arXiv:1810.06583 (2018)},
}

@misc{att,
    title   = {Attention Is All You Need},
    author  = {Ashish Vaswani and Noam Shazeer and Niki Parmar and Jakob Uszkoreit and Llion Jones and Aidan N. Gomez and Lukasz Kaiser and Illia Polosukhin},
    year    = {2017},
    eprint  = {1706.03762},
    archivePrefix = {arXiv},
    primaryClass = {cs.CL}
}

@article{ref11,
  title={Zoom In: An Introduction to Circuits},
  author={Olah, Chris and Cammarata, Nick and Schubert, Ludwig and Goh, Gabriel and Petrov, Michael and Carter, Shan},
  journal={Distill},
  volume={5},
  number={3},
  pages={e00024.001},
  year={2020},
  doi={10.23915/distill.00024.001},
  url={https://distill.pub/2020/circuits/zoom-in}
}

@article{ref14,
  title={Towards Monosemanticity: Decomposing Language Models With Dictionary Learning},
  author={Bricken, Trenton and Templeton, Adly and Batson, Joshua and Chen, Brian and Jermyn, Adam and Conerly, Tom and Turner, Nick and Anil, Cem and Denison, Carson and Askell, Amanda and others},
  journal={Transformer Circuits Thread},
  year={2023},
  url={https://transformer-circuits.pub/2023/monosemanticity/index.html}
}

@article{ref15,
  title={The Missing Curve Detectors of InceptionV1: Applying Sparse Autoencoders to InceptionV1 Early Vision},
  author={Gorton, Liv},
  journal={arXiv preprint arXiv:2406.03662},
  year={2024},
  url={https://arxiv.org/abs/2406.03662}
}

@article{ref16,
  title={In-context Learning and Induction Heads},
  author={Olsson, Catherine and Elhage, Nelson and Nanda, Neel and Joseph, Nicholas and DasSarma, Nova and Henighan, Tom and Mann, Ben and Askell, Amanda and Bai, Yuntao and Chen, Anna and others},
  journal={Transformer Circuits Thread},
  year={2022},
  url={https://transformer-circuits.pub/2022/in-context-learning/index.html}
}

@article{ref17,
  title={Sparse Feature Circuits: Discovering and Editing Interpretable Causal Graphs in Language Models},
  author={Marks, Samuel and Rager, Can and Michaud, Eric J and Belinkov, Yonatan and Bau, David and Mueller, Aaron},
  journal={arXiv preprint arXiv:2403.19647},
  year={2024},
  url={https://arxiv.org/abs/2403.19647}
}

@article{olah2020zoom,
  title={Zoom In: An Introduction to Circuits},
  author={Olah, Chris and Satyanarayan, Arvind and Wusser, Ludwig Schubert and Carter, Shan},
  journal={Distill},
  year={2020},
  doi={10.23915/distill.00024.001}
}

@article{elhage2022superposition,
  title={A Mechanistic Interpretability Analysis of Superposition in Neural Networks},
  author={Elhage, Nelson and Nanda, Neel and others},
  journal={Transformer Circuits Thread},
  year={2022},
  url={https://transformer-circuits.pub/2022/superposition/}
}

@misc{bills2023language,
      title={Language Models Represent Space and Time}, 
      author={Wes Gurnee and Max Tegmark},
      year={2024},
      eprint={2310.02207},
      archivePrefix={arXiv},
      primaryClass={cs.LG},
      url={https://arxiv.org/abs/2310.02207}, 
}

@article{quiroga2005invariant,
  title={Invariant visual representation by single neurons in the human brain},
  author={Quiroga, Rodrigo and others},
  journal={Nature},
  volume={435},
  number={7045},
  pages={1102--1107},
  year={2005}
}

@article{rigotti2013mixed,
  title={The importance of mixed selectivity in complex cognitive tasks},
  author={Rigotti, Mattia and Barak, Omri and Warden, Melissa and others},
  journal={Nature},
  volume={497},
  number={7451},
  pages={585--590},
  year={2013}
}

@article{bondi2019mci,
author = {Mark W. Bondi and Emily C. Edmonds and Amy J. Jak and Lindsay R. Clark and Lisa Delano-Wood and Carrie R. McDonald and Daniel A. Nation and David J. Libon and Rhoda Au and Douglas Galasko and David P. Salmon and },
title ={Neuropsychological Criteria for Mild Cognitive Impairment Improves Diagnostic Precision, Biomarker Associations, and Progression Rates},

journal = {Journal of Alzheimer’s Disease},
volume = {42},
number = {1},
pages = {275-289},
year = {2014},
doi = {10.3233/JAD-140276},
    note ={PMID: 24844687},

URL = { 
    
    
        https://journals.sagepub.com/doi/abs/10.3233/JAD-140276
    

},
eprint = { 
    
    
        https://journals.sagepub.com/doi/pdf/10.3233/JAD-140276
    

}
}

@misc{ref18,
  title={SAE Visualizer},
  author={McDougall, Callum},
  year={2024},
  url={https://github.com/callummcdougall/SAE-Visualizer}
}

@article{ref19,
  title={Language models can explain neurons in language models},
  author={Bills, Steven and Cammarata, Nick and Mossing, Dan and Tillman, Henk and Gao, Leo and Goh, Gabriel and Sutskever, Ilya and Leike, Jan and Wu, Jeff and Saunders, William},
  journal={openaipublic},
  year={2023},
  url={https://openaipublic.blob.core.windows.net/neuron-explainer/paper/index.html}
}

@misc{gate,
      title={Improving Dictionary Learning with Gated Sparse Autoencoders}, 
      author={Senthooran Rajamanoharan and Arthur Conmy and Lewis Smith and Tom Lieberum and Vikrant Varma and János Kramár and Rohin Shah and Neel Nanda},
      year={2024},
      eprint={2404.16014},
      archivePrefix={arXiv},
      primaryClass={cs.LG},
      url={https://arxiv.org/abs/2404.16014}, 
}

@misc{ros,
      title={Rethinking Stability for Attribution-based Explanations}, 
      author={Chirag Agarwal and Nari Johnson and Martin Pawelczyk and Satyapriya Krishna and Eshika Saxena and Marinka Zitnik and Himabindu Lakkaraju},
      year={2022},
      eprint={2203.06877},
      archivePrefix={arXiv},
      primaryClass={cs.LG},
      url={https://arxiv.org/abs/2203.06877}, 
}

@article{mx1,
title = {Explainable artificial intelligence (XAI) in deep learning-based medical image analysis},
journal = {Medical Image Analysis},
volume = {79},
pages = {102470},
year = {2022},
issn = {1361-8415},
doi = {https://doi.org/10.1016/j.media.2022.102470},
url = {https://www.sciencedirect.com/science/article/pii/S1361841522001177},
author = {Bas H.M. van der Velden and Hugo J. Kuijf and Kenneth G.A. Gilhuijs and Max A. Viergever},

}

@article{mx2,
title = {ExplAIn: Explanatory artificial intelligence for diabetic retinopathy diagnosis},
journal = {Medical Image Analysis},
volume = {72},
pages = {102118},
year = {2021},
issn = {1361-8415},
doi = {https://doi.org/10.1016/j.media.2021.102118},
url = {https://www.sciencedirect.com/science/article/pii/S136184152100164X},
author = {Gwenolé Quellec and Hassan {Al Hajj} and Mathieu Lamard and Pierre-Henri Conze and Pascale Massin and Béatrice Cochener}
}

@misc{sael,
   title = {SAELens},
   author = {Bloom, Joseph and Tigges, Curt and Duong, Anthony and Chanin, David},
   year = {2024},
   howpublished = {\url{https://github.com/jbloomAus/SAELens}},
}

@article{quan,
  author  = {Anna Hedstr{\"{o}}m and Leander Weber and Daniel Krakowczyk and Dilyara Bareeva and Franz Motzkus and Wojciech Samek and Sebastian Lapuschkin and Marina Marina M.{-}C. H{\"{o}}hne},
  title   = {Quantus: An Explainable AI Toolkit for Responsible Evaluation of Neural Network Explanations and Beyond},
  journal = {Journal of Machine Learning Research},
  year    = {2023},
  volume  = {24},
  number  = {34},
  pages   = {1--11},
  url     = {http://jmlr.org/papers/v24/22-0142.html}
}

@article{mine,
  author    = {Michail Mamalakis and Krit Dwivedi and Michael Sharkey and Samer Alabed and David Kiely and Andrew J. Swift},
  title     = {A transparent artificial intelligence framework to assess lung disease in pulmonary hypertension},
  journal   = {Scientific Reports},
  volume    = {13},
  number    = {1},
  pages     = {3812},
  year      = {2023},
  publisher = {Springer Nature},
  doi       = {10.1038/s41598-023-30503-4},
  url       = {https://doi.org/10.1038/s41598-023-30503-4}
}

@misc{mi,
      title={Mechanistic Interpretability for AI Safety -- A Review}, 
      author={Leonard Bereska and Efstratios Gavves},
      year={2024},
      eprint={2404.14082},
      archivePrefix={arXiv},
      primaryClass={cs.AI},
      url={https://arxiv.org/abs/2404.14082}, 
}

@misc{sae,
      title={Sparse Autoencoders Find Highly Interpretable Features in Language Models}, 
      author={Hoagy Cunningham and Aidan Ewart and Logan Riggs and Robert Huben and Lee Sharkey},
      year={2023},
      eprint={2309.08600},
      archivePrefix={arXiv},
      primaryClass={cs.LG},
      url={https://arxiv.org/abs/2309.08600},
}

@misc{super,
      title={Toy Models of Superposition}, 
      author={Nelson Elhage and Tristan Hume and Catherine Olsson and Nicholas Schiefer and Tom Henighan and Shauna Kravec and Zac Hatfield-Dodds and Robert Lasenby and Dawn Drain and Carol Chen and Roger Grosse and Sam McCandlish and Jared Kaplan and Dario Amodei and Martin Wattenberg and Christopher Olah},
      year={2022},
      eprint={2209.10652},
      archivePrefix={arXiv},
      primaryClass={cs.LG},
      url={https://arxiv.org/abs/2209.10652}, 
}

@article{manif2021trust,
author = {Liu, Haochen and Wang, Yiqi and Fan, Wenqi and Liu, Xiaorui and Li, Yaxin and Jain, Shaili and Liu, Yunhao and Jain, Anil and Tang, Jiliang},
title = {Trustworthy AI: A Computational Perspective},
year = {2022},
issue_date = {February 2023},
publisher = {Association for Computing Machinery},
address = {New York, NY, USA},
volume = {14},
number = {1},
issn = {2157-6904},
url = {https://doi.org/10.1145/3546872},
doi = {10.1145/3546872},
journal = {ACM Trans. Intell. Syst. Technol.},
month = nov,
articleno = {4},
numpages = {59}
}

@article{jack2018nia,
  title={NIA-AA Research Framework: Toward a biological definition of Alzheimer’s disease},
  author={Jack, Clifford R and others},
  journal={Alzheimer's \& Dementia},
  volume={14},
  number={4},
  pages={535--562},
  year={2018}
}

@misc{gpt,
  author       = {OpenAI},
  title        = {{ChatGPT-4o}},
  year         = {2024},
  howpublished = {\url{https://openai.com/chatgpt}},
  note         = {Accessed May 2025}
}

@article{touvron2023llama,
  title={LLaMA 2: Open foundation and fine-tuned chat models},
  author={Touvron, Hugo and others},
  journal={arXiv preprint arXiv:2307.09288},
  year={2023}
}

@misc{int,
      title={Axiomatic Attribution for Deep Networks}, 
      author={Mukund Sundararajan and Ankur Taly and Qiqi Yan},
      year={2017},
      eprint={1703.01365},
      archivePrefix={arXiv},
      primaryClass={cs.LG},
      url={https://arxiv.org/abs/1703.01365}, 
}

@article{tahami,
  author  = {Tahami Monfared, Alireza A. and Byrnes, Melissa J. and White, Lauren A. and Zhang, Jianwei and Yu, Eric and Lin, Jeng},
  title   = {Alzheimer’s Disease: Epidemiology and Clinical Progression},
  journal = {Neurology and Therapy},
  volume  = {11},
  pages   = {553--569},
  year    = {2022},
  doi     = {10.1007/s40120-022-00338-8}
}

@article{lee2019predicting,
  title={Predicting Alzheimer’s disease progression using multi-modal deep learning approach},
  author={Lee, Garam and Nho, Kwangsik and Kang, Byungkon and Sohn, Kyung-Ah and Kim, Dokyoon and Weiner, Michael W. and Alzheimer’s Disease Neuroimaging Initiative},
  journal={Scientific Reports},
  volume={9},
  number={1},
  pages={1952},
  year={2019},
  publisher={Nature Publishing Group},
  doi={10.1038/s41598-018-37769-z},
  url={https://doi.org/10.1038/s41598-018-37769-z}
}

@article{lundberg2017unified,
  title={A unified approach to interpreting model predictions},
  author={Lundberg, Scott M and Lee, Su-In},
  journal={Advances in neural information processing systems},
  volume={30},
  year={2017}
}

@article{ref20,
  title={Scaling Monosemanticity: Extracting Interpretable Features from Claude 3},
  author={Templeton, Adly and Conerly, Tom and Marcus, Jonathan and Lindsey, Jack and Bricken, Trenton and Chen, Brian and Pearce, Adam and Citro, Craig and Ameisen, Emmanuel and Jones, Andy and others},
  journal={Transformer Circuits Thread},
  year={2024},
  url={https://transformer-circuits.pub/2024/scaling-monosemanticity/index.html}
}

@article{mamalakis,
   title={Solving the enigma: Enhancing faithfulness and comprehensibility in explanations of deep networks},
   volume={6},
   ISSN={2666-6510},
   url={http://dx.doi.org/10.1016/j.aiopen.2025.02.001},
   DOI={10.1016/j.aiopen.2025.02.001},
   journal={AI Open},
   publisher={Elsevier BV},
   author={Mamalakis, Michail and Mamalakis, Antonios and Agartz, Ingrid and Mørch-Johnsen, Lynn Egeland and Murray, Graham K. and Suckling, John and Lio, Pietro},
   year={2025},
   pages={70–81} }

@article{self,
    author  = {Toan Q. Nguyen and Julian Salazar},
    title   = {Transformers without Tears: Improving the Normalization of Self-Attention},
    year    = {2019},
    eprint  = {arXiv:1910.05895},
    doi     = {10.5281/zenodo.3525484},
}

@article{ho2020denoising,
    title={Denoising Diffusion Probabilistic Models},
    author={Jonathan Ho and Ajay Jain and Pieter Abbeel},
    year={2020},
    journal={arXiv preprint arxiv:2006.11239}
}

@inproceedings{unet,
  title={U-Net: Convolutional Networks for Biomedical Image Segmentation},
  author={Ronneberger, Olaf and Fischer, Philipp and Brox, Thomas},
  booktitle={International Conference on Medical Image Computing and Computer-Assisted Intervention (MICCAI)},
  pages={234--241},
  year={2015},
  publisher={Springer},
  doi={10.1007/978-3-319-24574-4_28}
}

@article{1,
title = {Using high spatial resolution fMRI to understand representation in the auditory network},
journal = {Progress in Neurobiology},
volume = {207},
pages = {101887},
year = {2021},
note = {How high spatiotemporal resolution fMRI can advance neuroscience},
issn = {0301-0082},
doi = {https://doi.org/10.1016/j.pneurobio.2020.101887},
url = {https://www.sciencedirect.com/science/article/pii/S0301008220301428},
author = {Michelle Moerel and Essa Yacoub and Omer Faruk Gulban and Agustin Lage-Castellanos and Federico {De Martino}}
}

@incollection{2,
title = {Chapter 8 - Bridging the gap between system and cell: The role of ultra-high field MRI in human neuroscience},
editor = {Tara Mahfoud and Sam McLean and Nikolas Rose},
series = {Progress in Brain Research},
publisher = {Elsevier},
volume = {233},
pages = {179-220},
year = {2017},
booktitle = {Vital Models},
issn = {0079-6123},
doi = {https://doi.org/10.1016/bs.pbr.2017.05.005},
url = {https://www.sciencedirect.com/science/article/pii/S0079612317300493},
author = {Robert Turner and Daniel {De Haan}}
}

@article{3,
	author = {Feinberg, David A. and Beckett, Alexander J. S. and Vu, An T. and Stockmann, Jason and Huber, Laurentius and Ma, Samantha and Ahn, Sinyeob and Setsompop, Kawin and Cao, Xiaozhi and Park, Suhyung and Liu, Chunlei and Wald, Lawrence L. and Polimeni, Jonathan R. and Mareyam, Azma and Gruber, Bernhard and Stirnberg, R{\"u}diger and Liao, Congyu and Yacoub, Essa and Davids, Mathias and Bell, Paul and Rummert, Elmar and Koehler, Michael and Potthast, Andreas and Gonzalez-Insua, Ignacio and Stocker, Stefan and Gunamony, Shajan and Dietz, Peter},
	date = {2023/12/01},
	doi = {10.1038/s41592-023-02068-7},
	id = {Feinberg2023},
	isbn = {1548-7105},
	journal = {Nature Methods},
	number = {12},
	pages = {2048--2057},
	title = {Next-generation MRI scanner designed for ultra-high-resolution human brain imaging at 7 Tesla},
	url = {https://doi.org/10.1038/s41592-023-02068-7},
	volume = {20},
	year = {2023}}

@article{4,
	author = {Markello, Ross D. and Hansen, Justine Y. and Liu, Zhen-Qi and Bazinet, Vincent and Shafiei, Golia and Su{\'a}rez, Laura E. and Blostein, Nadia and Seidlitz, Jakob and Baillet, Sylvain and Satterthwaite, Theodore D. and Chakravarty, M. Mallar and Raznahan, Armin and Misic, Bratislav},
	date = {2022/11/01},
	doi = {10.1038/s41592-022-01625-w},
	id = {Markello2022},
	isbn = {1548-7105},
	journal = {Nature Methods},
	number = {11},
	pages = {1472--1479},
	title = {neuromaps: structural and functional interpretation of brain maps},
	url = {https://doi.org/10.1038/s41592-022-01625-w},
	volume = {19},
	year = {2022}}

@misc{svarm,
      title={Approximating the Shapley Value without Marginal Contributions}, 
      author={Patrick Kolpaczki and Viktor Bengs and Maximilian Muschalik and Eyke Hüllermeier},
      year={2024},
      eprint={2302.00736},
      archivePrefix={arXiv},
      primaryClass={cs.LG}
}

@article{lrp,
    doi = {10.1371/journal.pone.0130140},
    author = {Bach, Sebastian AND Binder, Alexander AND Montavon, Grégoire AND Klauschen, Frederick AND Müller, Klaus-Robert AND Samek, Wojciech},
    journal = {PLOS ONE},
    publisher = {Public Library of Science},
    title = {On Pixel-Wise Explanations for Non-Linear Classifier Decisions by Layer-Wise Relevance Propagation},
    year = {2015},
    month = {07},
    volume = {10},
    url = {https://doi.org/10.1371/journal.pone.0130140},
    pages = {1-46},
    number = {7},

}

@article{shap,
  author       = {Scott M. Lundberg and
                  Su{-}In Lee},
  title        = {A unified approach to interpreting model predictions},
  journal      = {CoRR},
  volume       = {abs/1705.07874},
  year         = {2017},
  url          = {http://arxiv.org/abs/1705.07874},
  eprinttype    = {arXiv},
  eprint       = {1705.07874},
  bibsource    = {dblp computer science bibliography, https://dblp.org}
}

@article{xai,

  author={Samek, Wojciech and Montavon, Grégoire and Lapuschkin, Sebastian and Anders, Christopher J. and Müller, Klaus-Robert},

  journal={Proceedings of the IEEE}, 

  title={Explaining Deep Neural Networks and Beyond: A Review of Methods and Applications}, 

  year={2021},

  volume={109},

  number={3},

  pages={247-278},

  doi={10.1109/JPROC.2021.3060483}}

@misc{adam,
  doi = {10.48550/ARXIV.1412.6980},
  
  url = {https://arxiv.org/abs/1412.6980},
  
  author = {Kingma, Diederik P. and Ba, Jimmy},
  
  title = {Adam: A Method for Stochastic Optimization},
  
  publisher = {arXiv},
  
  year = {2014},
  
  copyright = {arXiv.org perpetual, non-exclusive license}
}

@misc{xaimi,
      title={Explainable deep learning models in medical image analysis}, 
      author={Amitojdeep Singh and Sourya Sengupta and Vasudevan Lakshminarayanan},
      year={2020},
      eprint={2005.13799},
      archivePrefix={arXiv},
      primaryClass={cs.CV}
}

@article{f,
  doi = {10.48550/ARXIV.2005.00631},
  
  url = {https://arxiv.org/abs/2005.00631},
  
  author = {Bhatt, Umang and Weller, Adrian and Moura, José M. F.},
  
  title = {Evaluating and Aggregating Feature-based Model Explanations},
  
  publisher = {arXiv},
  
  year = {2020},
  
  copyright = {arXiv.org perpetual, non-exclusive license}
}
\bibliographystyle{unsrt}  

\newpage
\appendix
\onecolumn

\section{Related Work}
\subsubsection{Attributional Interpretabillity}
Attributional interpretability (AtI), a branch of explainable AI (XAI), focuses on explaining model outputs by tracing predictions back to individual input contributions, often using gradient-based methods \cite{mi}. While gradients provide insights into the relationship between inputs and outputs, they can be sensitive to perturbations or discontinuities, posing challenges for reliable interpretation.
\par AtI encompasses various methods for interpreting complex, nonlinear models, including techniques like Local Interpretable Model-agnostic Explanations (LIME; \cite{lime}) and SHapley Additive exPlanations (SHAP; \cite{shap}). In medical imaging, popular attribution techniques include SHAP, Layer-wise Relevance Propagation (LRP; \cite{lrp}), and gradient-based methods like GRAD-CAM (\cite{xaimi}). These methods aim to enhance trust in models and provide valuable insights into decision-making processes. However, they face limitations. For instance, LRP emphasizes positive preactivations, often yielding less precise explanations, while SHAP is computationally intensive due to the complexity of calculating Shapley values \cite{shap}. Adaptations like Monte Carlo methods and stratified sampling (e.g., SVARM) have improved the efficiency and precision of certain techniques \cite{svarm}.

\subsubsection{Mechanistic Interpretability and Sparse Autoencoder}
\par Mechanistic interpretability (MI), a key area of explainable AI (XAI), focuses on understanding the internal activation patterns of AI models by analyzing their fundamental components, such as features, neurons, layers, and connections. Unlike AtI, MI takes a bottom-up approach, aiming to uncover the causal relationships and precise computations that transform inputs into outputs. This method identifies specific neural circuits driving behavior and provides a reverse-engineering perspective. Insights from fields like physics, neuroscience, and systems biology further guide the development of transparent and value-aligned AI systems.
\par A core principle of MI is the concept of polysemanticity, where individual neurons encode multiple concepts, contrasted with monosemanticity, where neurons correspond to a single semantic concept. Polysemanticity reduces interpretability, as neurons represent overlapping features. Structures like sparse autoencoders (SAEs) address this by leveraging the superposition hypothesis, which posits that neural networks use high-dimensional spaces to represent more features than the number of neurons, encoding them in nearly orthogonal directions. SAEs decompose embeddings from deep layers, such as MLPs or transformer attention layers, into higher-dimensional monosemantic representations, aligning activation patterns with specific concepts of interest \cite{sae,super}.

Sparse Autoencoder architectures have significantly advanced our understanding of feature representations in language and vision models~\cite{ref15}. Neural network behavior is often interpreted through \textit{computational circuits}—groups of neurons that compute meaningful functions, such as edge detectors~\cite{ref11} or word-copying units~\cite{ref16}. Leveraging SAE-derived features instead of raw neurons has improved the interpretability of circuits associated with complex behaviors~\cite{ref17}. This shift enables clearer mappings between neuron activations and high-level functions, facilitating validation of model behavior~\cite{mi}. By aligning internal representations with privileged basis directions—distinct semantic vectors within network layers—researchers further enhance monosemanticity and advance the interpretability of deep models.

\section{Mathematical Background}
\subsection{Attributional theory and methods}

Attribution explainability methods follow the framework of additive feature attribution, where the explanation model \( g(f, \mathbf{x}) \) is represented as a linear function of simplified input features:

\begin{equation}
    g(f, \mathbf{x}) = \phi_0 + \sum_{i=1}^M \phi_i x_i
    \label{eq:additive1}
\end{equation}

Here, \( f \) is the predictive model, \( \phi_i \in \mathbb{R} \) is the attribution (importance) assigned to feature \( x_i \), and \( M \) is the number of simplified input features.

For this study, we employed six well-established attributional interpretability methods applied to Transformer-Based language models and LLMs, denoted as \( K = 6 \): \textit{Feature Ablation}, \textit{Layer Activations} (which capture the embedding activation space of a specific layer of interest within the LM), \textit{Layer DeepLIFT SHAP}, \textit{Layer Gradient SHAP} \cite{shap}, \textit{Layer Integrated Gradients} \cite{int}, and \textit{Layer Gradient $\times$ Activation}. 

To align these layer-wise interpretability methods with the additive feature attribution framework, we reinterpret the internal activations (i.e., latent units) of a network layer \( L \) as simplified input features. The objective is to estimate an attribution score \( \phi_i \) for each unit, where \( \phi_i \in \mathbb{R} \) quantifies the contribution of the corresponding neuron to the model’s prediction.

\noindent \textbf{ Layer SHAP implementations:} This directly corresponds to the Shapley formulation:
\begin{equation}
    \phi_i = \sum_{S \subseteq F \setminus \{i\}} \frac{|S|!(|F|-|S|-1)!}{|F|!} \left[ f_{S \cup \{i\}}(x_{S \cup \{i\}}) - f_S(x_S) \right]
    \label{eq:shap}
\end{equation}

In practice, Deep SHAP approximates this using sampling and a chain-rule based linearization over network layers \cite{shap}.
\textit{Gradient SHAP} assumes that input features are independent and that the explanation model is linear, allowing explanations to be expressed as an additive composition of feature contributions. Under these assumptions, SHAP values \cite{shap} can be approximated by computing the expected gradients over a distribution of perturbed inputs. Specifically, Gaussian noise is added to each input feature to generate multiple baseline samples, and the resulting gradients are averaged to approximate SHAP attributions.

\vspace{1em}
\noindent \textbf{ Activation Attribution:} This method treats the raw activation \( a^L_i(\mathbf{x}) \) as proportional to its importance in the output. In the additive form:
\begin{equation}
    \phi_i = a^L_i(\mathbf{x})
\end{equation}
Assuming linearity between layer \( L \) and the output, activations themselves serve as proxy contributions.

\vspace{1em}

\noindent \textbf{Gradient × Activation Attribution:} This method computes the element-wise product between the activation values and the gradients of the model output with respect to those activations, thereby capturing the first-order sensitivity of the output to the neurons in the layer. 
To this end, the method estimates the first-order sensitivity of the output with respect to the activation:
\begin{equation}
    \phi_i = a^L_i(\mathbf{x}) \cdot \frac{\partial f}{\partial a^L_i}(\mathbf{x})
\end{equation}
This corresponds to a local linear approximation (first-order Taylor expansion) of the model at \( \mathbf{x} \), akin to DeepLIFT and the SHAP linearization used in DeepLift SHAP \cite{shap}.

\vspace{1em}

\noindent \textbf{ Feature Ablation Attribution:} This attributional interpretability technique is a perturbation-based approach to estimating attributions. It involves replacing the input or output values of a selected layer with a given baseline or reference value and computing the resulting change in the model’s output. By default, each neuron (i.e., scalar input or output value) within the layer is ablated independently. For neuron group \( S \subseteq \{1, \dots, d_L\} \), the perturbed activation is:

\begin{equation}
    \tilde{a}^L_i =
    \begin{cases}
        b^L_i & \text{if } i \in S, \\
        a^L_i(\mathbf{x}) & \text{otherwise},
    \end{cases}
\end{equation}

and the attribution is the marginal effect:

\begin{equation}
    \phi_S = f\left(\mathbf{x}; \tilde{\mathbf{a}}^L_S\right) - f(\mathbf{x})
\end{equation}
\bigskip

All attribution methods were applied to the final (22nd) layer of the \textsc{Modern-BERT} LM—the model variant that achieved the highest classification accuracy in our evaluations (see Suplementary material section 1.1).
These formulations allow us to ground various neural attribution techniques within a unified additive explanation model, facilitating their comparison and hybridization under shared theoretical assumptions. 

\subsection{Attributional explanation optimizer framework}

Let \( \mathcal{A} = \{A_1, A_2, \dots, A_K\} \) denote the set of \( K = 6 \) attribution methods applied to the final layer \( L \) of the model \( f \). Each method \( A_k \) generates an attribution vector \( \boldsymbol{\phi}^{(k)} = [\phi^{(k)}_1, \phi^{(k)}_2, \dots, \phi^{(k)}_M] \), where \( M \) is the number of latent features (neurons) in layer \( L \). The goal is to derive a unified attribution vector \( \bar{\boldsymbol{\phi}} \) that captures the consensus explanation across methods.

\subsubsection{Scoring and Weighting Attribution Methods}

Each attribution vector \( \boldsymbol{\phi}^{(k)} \) is evaluated using the following quality metrics:
\subsubsection{Evaluation Interpretability Metrics}

We evaluate the robustness of each attribution method \( A_k \) using the following stability metrics:

\vspace{0.5em}

\noindent \textbf{Relative Input Stability (RIS):}
\begin{equation}
    M_{\text{RIS}}^{(k)}=\mathrm{RIS}(f, \boldsymbol{\phi}^{(k)}; \mathbf{x}) =
    \frac{ \|\mathbf{x}\|_p }{ \|\boldsymbol{\phi}^{(k)}(\mathbf{x})\|_p }
    \max_{\mathbf{x'} \in \mathcal{N}_{\mathbf{x}},\, \hat{y}_{\mathbf{x'}} = \hat{y}_{\mathbf{x}}}
    \frac{ \| \boldsymbol{\phi}^{(k)}(\mathbf{x}) - \boldsymbol{\phi}^{(k)}(\mathbf{x'}) \|_p }
         { \| \mathbf{x} - \mathbf{x'} \|_p }
    \label{eq:ris}
\end{equation}

\noindent \textbf{Relative Output Stability (ROS):}
\begin{equation}
    M_{\text{ROS}}^{(k)}=\mathrm{ROS}(f, \boldsymbol{\phi}^{(k)}; \mathbf{x}) =
    \frac{ \|f(\mathbf{x})\|_p }{ \|\boldsymbol{\phi}^{(k)}(\mathbf{x})\|_p }
    \max_{\mathbf{x'} \in \mathcal{N}_{\mathbf{x}},\, \hat{y}_{\mathbf{x'}} = \hat{y}_{\mathbf{x}}}
    \frac{ \| \boldsymbol{\phi}^{(k)}(\mathbf{x}) - \boldsymbol{\phi}^{(k)}(\mathbf{x'}) \|_p }
         { \| f(\mathbf{x}) - f(\mathbf{x'}) \|_p }
    \label{eq:ros}
\end{equation}

Here, \( \mathcal{N}_{\mathbf{x}} \) denotes a neighborhood of perturbed inputs \( \mathbf{x'} \) around \( \mathbf{x} \), and \( \hat{y}_{\mathbf{x}} \) is the predicted class label. Both metrics measure the relative sensitivity of the attribution vector \( \boldsymbol{\phi}^{(k)} \) to perturbations in the input or output space.

\vspace{0.5em}

\noindent \textbf{Sparseness Metric:}
We quantify the \textbf{sparseness} of the attribution vector \( \boldsymbol{\phi}^{(k)} \in \mathbb{R}^d \) using the \textit{Gini Index}, a measure of inequality that has been shown to satisfy several desirable properties for evaluating sparseness \cite{spar}. This formulation is adopted in the context of explaining neural network predictions \cite{spar}.

Let \( v \in \mathbb{R}^d_{\geq 0} \) be a non-negative vector. Denote by \( v_{(k)} \) the \( k \)-th smallest element in \( v \) after sorting it in non-decreasing order. Then, the \textbf{Gini Index} \( G(v) \in [0,1] \) is defined as:

\begin{equation}
G(v) = 1 - 2 \sum_{k=1}^{d} \frac{v_{(k)}}{\|v\|_1} \cdot \left( \frac{d - k + 0.5}{d} \right),
\label{eq:gini}
\end{equation}

where \( \|v\|_1 = \sum_{i=1}^d v_i \) is the \( \ell_1 \)-norm of \( v \).
To evaluate the sparseness of an attribution vector \( \boldsymbol{\phi}^{(k)} \), we apply the Gini Index to the vector of its absolute values:

\[
\text{Sparseness}\left( \boldsymbol{\phi}^{(k)} \right) = G\left( \left| \boldsymbol{\phi}^{(k)} \right| \right),
\]

where \( \left| \boldsymbol{\phi}^{(k)} \right| = \left( |\phi^{(k)}_1|, |\phi^{(k)}_2|, \ldots, |\phi^{(k)}_d| \right) \).

Higher values of \( G\left( \left| \boldsymbol{\phi}^{(k)} \right| \right) \) indicate greater sparseness. In the extreme case, if only one component is non-zero, the Gini Index reaches its maximum value of 1, indicating perfect sparseness. If all components are equal, the Gini Index is 0.

\subsubsection{Aggregation of Attributions}

The weighted average attribution vector \( \bar{\boldsymbol{\phi}} \) is calculated as:

\begin{equation}
\bar{\boldsymbol{\phi}} = \sum_{k=1}^K w_k \cdot \boldsymbol{\phi}^{(k)}
\end{equation}

This vector serves as the target explanation for the optimization process.

\subsubsection{Explanation Reconstruction via Encoder–Decoder Models}

An encoder–decoder model is trained to generate a reconstructed explanation \( \hat{\boldsymbol{\phi}} \) from the original input \( \boldsymbol{x} \). Two architectures are considered the Diffusion UNet1D \cite{unet} and the x-transformer autoencoder \cite{att,self}.

\paragraph{Diffusion model: } The diffusion model follows the basic structure of a 1-dimensional U-Net and is trained using diffusion principles. In this framework, diffusion models \cite{ho2020denoising} are latent variable models in which the observed data $\mathbf{\phi}^{(k)}_0$ is gradually corrupted through a forward noising process, producing a sequence of latent variables $\mathbf{\phi}^{(k)}_{1:T}$. A corresponding reverse process is then learned to recover the original data from noise. The mathematical formulation is as follows:

\textsc{Forward Process:} A fixed Markov chain progressively adds Gaussian noise to the data:
\begin{equation}
q(\mathbf{\phi}^{(k)}_{1:T}|\mathbf{\phi}^{(k)}_0) := \prod_{t=1}^T q(\mathbf{\phi}^{(k)}_t|\mathbf{\phi}^{(k)}_{t-1}), \quad q(\mathbf{\phi}^{(k)}_t|\mathbf{\phi}^{(k)}_{t-1}) := \mathcal{N}(\mathbf{\phi}^{(k)}_t; \sqrt{1 - \beta_t} \mathbf{\phi}^{(k)}_{t-1}, \beta_t \mathbf{I})
\label{eq:forward}
\end{equation}

Alternatively, sampling from the forward process at an arbitrary timestep $t$ is possible in closed form:
\begin{equation}
q(\mathbf{\phi}^{(k)}_t|\mathbf{\phi}^{(k)}_0) = \mathcal{N}(\mathbf{\phi}^{(k)}_t; \sqrt{\bar{\alpha}_t} \mathbf{\phi}^{(k)}_0, (1 - \bar{\alpha}_t) \mathbf{I}),
\label{eq:closed_form_forward}
\end{equation}
where $\alpha_t := 1 - \beta_t$ and $\bar{\alpha}_t := \prod_{s=1}^t \alpha_s$.

\textsc{Reverse Process:} A learned time-reversal model with Gaussian transitions:
\begin{equation}
p_\theta(\mathbf{\phi}^{(k)}_{0:T}) := p(\mathbf{\phi}^{(k)}_T) \prod_{t=1}^T p_\theta(\mathbf{\phi}^{(k)}_{t-1}|\mathbf{\phi}^{(k)}_t), \quad p_\theta(\mathbf{\phi}^{(k)}_{t-1}|\mathbf{\phi}^{(k)}_t) := \mathcal{N}(\mathbf{\phi}^{(k)}_{t-1}; \boldsymbol{\mu}_\theta(\mathbf{\phi}^{(k)}_t, t), \Sigma_\theta(\mathbf{\phi}^{(k)}_t, t)),
\label{eq:reverse}
\end{equation}
where $p(\mathbf{\phi}^{(k)}_T) := \mathcal{N}(\mathbf{\phi}^{(k)}_T; \mathbf{0}, \mathbf{I})$.

\textsc{Training Objective:} The training objective of diffusion models is based on a variational bound, which includes Kullback–Leibler (KL) divergence terms. The KL term comparing the true posterior from the forward process and the model's learned reverse process is written as:
\begin{equation}
    \mathrm{KL}\left(q({\phi}^{(k)}_{t-1} \mid {\phi}^{(k)}_t, {\phi}^{(k)}_0) \,\|\, p_\theta({\phi}^{(k)}_{t-1} \mid {\phi}^{(k)}_t)\right)
\end{equation}
Both distributions are Gaussian:
\begin{align}
    q({\phi}^{(k)}_{t-1} \mid {\phi}^{(k)}_t, {\phi}^{(k)}_0) &= \mathcal{N}({\phi}^{(k)}_{t-1}; \tilde{\mu}_t({\phi}^{(k)}_t, {\phi}^{(k)}_0), \tilde{\beta}_t \mathbf{I}) \\
    p_\theta({\phi}^{(k)}_{t-1} \mid {\phi}^{(k)}_t) &= \mathcal{N}({\phi}^{(k)}_{t-1}; \mu_\theta({\phi}^{(k)}_t, t), \sigma_t^2 \mathbf{I})
\end{align}

The closed-form KL divergence between two Gaussians \( \mathcal{N}(\mu_1, \sigma_1^2 \mathbf{I}) \) and \( \mathcal{N}(\mu_2, \sigma_2^2 \mathbf{I}) \) in \( d \)-dimensions is:

\begin{equation}
    \mathrm{KL} = \frac{1}{2} \left[
        \log\left( \frac{\sigma_2^2}{\sigma_1^2} \right)
        + \frac{\sigma_1^2 + \|\mu_1 - \mu_2\|^2}{\sigma_2^2}
        - d
    \right]
\end{equation}

In our setting, this term is computed for each timestep \( t \) and summed across all steps:

\begin{equation}
    \mathcal{L}_{1:T-1} = \sum_{t=2}^T \mathbb{E}_{q({\phi}^{(k)}_0, {\phi}^{(k)}_t)} \left[
        \mathrm{KL}\left(q({\phi}^{(k)}_{t-1} \mid {\phi}^{(k)}_t, {\phi}^{(k)}_0) \,\|\, p_\theta({\phi}^{(k)}_{t-1} \mid {\phi}^{(k)}_t)\right)
    \right]
\end{equation}

This forms a core part of the evidence lower bound (ELBO) optimized during training.
Using variational inference, we minimize the negative ELBO:
\begin{equation}
\mathcal{L} = \mathbb{E}_q \left[ -\log p(\mathbf{\phi}^{(k)}_T) + \sum_{t=1}^T \mathrm{KL}\left(q(\mathbf{\phi}^{(k)}_{t-1}|\mathbf{\phi}^{(k)}_t, \mathbf{\phi}^{(k)}_0) \Vert p_\theta(\mathbf{\phi}^{(k)}_{t-1}|\mathbf{\phi}^{(k)}_t)\right) - \log p_\theta(\mathbf{\phi}^{(k)}_0|\mathbf{\phi}^{(k)}_1) \right].
\label{eq:elbo}
\end{equation}

Each KL term compares Gaussian distributions and can be computed in closed form. The posterior $q(\mathbf{\phi}^{(k)}_{t-1}|\mathbf{\phi}^{(k)}_t, \mathbf{\phi}^{(k)}_0)$ is also Gaussian:
\begin{equation}
q(\mathbf{\phi}^{(k)}_{t-1}|\mathbf{\phi}^{(k)}_t, \mathbf{\phi}^{(k)}_0) = \mathcal{N}(\mathbf{\phi}^{(k)}_{t-1}; \tilde{\boldsymbol{\mu}}_t(\mathbf{\phi}^{(k)}_t, \mathbf{\phi}^{(k)}_0), \tilde{\beta}_t \mathbf{I}),
\label{eq:posterior}
\end{equation}
with:
\begin{align}
\tilde{\boldsymbol{\mu}}_t(\mathbf{\phi}^{(k)}_t, \mathbf{\phi}^{(k)}_0) &= \frac{\sqrt{\bar{\alpha}_{t-1}} \beta_t}{1 - \bar{\alpha}_t} \mathbf{\phi}^{(k)}_0 + \frac{\sqrt{\alpha_t}(1 - \bar{\alpha}_{t-1})}{1 - \bar{\alpha}_t} \mathbf{\phi}^{(k)}_t, \\
\tilde{\beta}_t &= \frac{1 - \bar{\alpha}_{t-1}}{1 - \bar{\alpha}_t} \beta_t.
\end{align}

\textsc{Simplified Training Loss:} The common parameterization rewrites the objective as denoising score matching:
\begin{equation}
\mathcal{L}_{\text{simple}}(\theta) := \mathbb{E}_{t, \mathbf{\phi}^{(k)}_0, \boldsymbol{\epsilon}} \left[ \left\| \boldsymbol{\epsilon} - \boldsymbol{\epsilon}_\theta(\sqrt{\bar{\alpha}_t} \mathbf{\phi}^{(k)}_0 + \sqrt{1 - \bar{\alpha}_t} \boldsymbol{\epsilon}, t) \right\|^2 \right],
\label{eq:simple_loss}
\end{equation}
where $\boldsymbol{\epsilon} \sim \mathcal{N}(0, \mathbf{I})$ and $\boldsymbol{\epsilon}_\theta$ is the neural network trained to predict noise.

In our implementation we compute the total loss for the diffusion model as: 
\begin{equation}
    \mathcal{L}_{\text{similarity}}(\hat{\boldsymbol{\phi}},\bar{\boldsymbol{\phi}}) = \mathcal{L}_{\text{similarity}}(\theta)= \frac{1}{K+1} \sum_{l=0}^{K} \mathcal{L}_{\text{simple}}^{(l)}(\theta)
\end{equation}

\paragraph{x-Transformer: }
Let the input sequence be:
\[
\boldsymbol{\phi}^{(k)} = [\boldsymbol{\phi}^{(k)}_1, \boldsymbol{\phi}^{(k)}_2, \dots, \boldsymbol{\phi}^{(k)}_T] \in \mathbb{R}^{T \times d_{\text{in}}}
\]
where \( d_{\text{in}} = 7 \) is the input dimensionality and \( T = 512 \) is the sequence length. We consider a Transformer-based encoder-decoder architecture operating on input sequences $\mathbf{\Phi}^{(k)} \in \mathbb{R}^{B \times T \times d_{\text{in}}}$ at diffusion step $k$, where:
$B$ is the batch size, $T$ is the sequence length, $d_{\text{in}}$ is the input feature dimension, and $\mathbf{\Phi}^{(k)}$ is the input sequence at step $k$.

The processing pipeline is mathematically formulated as follows:

\textsc{Input Projection and Positional Encoding:} We first project the input to the model dimension $d$ and add positional encodings:
\begin{equation}
\mathbf{X}_0 = \mathbf{W}_{\text{in}} \mathbf{\Phi}^{(k)} + \mathbf{P}, \quad \mathbf{X}_0 \in \mathbb{R}^{B \times T \times d}
\end{equation}
where: $\mathbf{W}_{\text{in}} \in \mathbb{R}^{d_{\text{in}} \times d}$ is a learnable linear projection matrix, and $\mathbf{P} \in \mathbb{R}^{1 \times T \times d}$ is a learnable positional embedding matrix.

\textsc{Encoder: Multi-Head Self-Attention Layers:} The encoder consists of $L_e$ stacked multi-head self-attention (MHSA) layers:
\begin{equation}
\mathbf{H}_{\text{enc}} = \text{MHSA}_{L_e} \circ \cdots \circ \text{MHSA}_{1}(\mathbf{X}_0)
\end{equation}
where each MHSA layer performs:
\begin{equation}
\text{MHSA}(\mathbf{X}) = \text{Softmax} \left( \frac{\mathbf{Q} \mathbf{K}^\top}{\sqrt{d_h}} \right) \mathbf{V}
\end{equation}
with: $\mathbf{Q}, \mathbf{K}, \mathbf{V}$: Query, Key, and Value matrices obtained via learned linear projections, and $d_h$: the dimensionality of each attention head.

\textsc{Decoder Input Projection:} During training, the decoder may receive the ground-truth output $\mathbf{\Phi}^{(k)}_{\text{target}} \in \mathbb{R}^{B \times T \times 1}$:
\begin{equation}
\mathbf{Y}_0 = \mathbf{W}_{\text{dec}} \mathbf{\Phi}^{(k)}_{\text{target}} + \mathbf{P}
\end{equation}
where $\mathbf{W}_{\text{dec}} \in \mathbb{R}^{1 \times d}$ is a projection matrix.

If no decoder input is available (e.g., during inference), $\mathbf{\Phi}^{(k)}_{\text{target}}$ is initialized to a zero tensor.

\textsc{Decoder MHSA + Cross-Attention Layers:} The decoder consists of $L_d$ layers of MHSA followed by cross-attention (CA) using the encoder context:
\begin{equation}
\mathbf{H}_{\text{dec}} = \text{CA}_{L_d} \circ \cdots \circ \text{CA}_{1} \left( \text{MHSA}_{L_d} \circ \cdots \circ \text{MHSA}_{1}(\mathbf{Y}_0) \,\middle|\, \mathbf{H}_{\text{enc}} \right)
\end{equation}
Each cross-attention (CA) layer uses the decoder hidden state as the query and encoder output as the key and value:
\begin{equation}
\text{CA}(\mathbf{Y}, \mathbf{H}_{\text{enc}}) = \text{Softmax} \left( \frac{\mathbf{Q}_{\text{dec}} \mathbf{K}_{\text{enc}}^\top}{\sqrt{d_h}} \right) \mathbf{V}_{\text{enc}}
\end{equation}

\textsc{Output Projection:} Finally, the decoder output is projected back to the target dimension:
\begin{equation}
\hat{\mathbf{\Phi}}^{(k)} = \mathbf{W}_{\text{out}} \mathbf{H}_{\text{dec}}, \quad \hat{\mathbf{\Phi}}^{(k)} \in \mathbb{R}^{B \times T \times 1}
\end{equation}
where $\mathbf{W}_{\text{out}} \in \mathbb{R}^{d \times 1}$ is a linear projection matrix.

The similarity cost function is given by the Mean Squared Error (MSE) loss between the predicted output of the x-Transformer and the target weighted attribution vector as follow:
\begin{equation}
    \mathcal{L}_{\text{similarity}}(\hat{\boldsymbol{\phi}},\bar{\boldsymbol{\phi}}) = \mathcal{L}_{\text{MSE}} = \frac{1}{T} \sum_{t=1}^{T} \left\| \hat{{\boldsymbol{\phi}}}_t - \bar{\boldsymbol{\phi}}_t \right\|^2,
\end{equation}

\subsection{The total cost function of the optimizer}
As previously highlighted, the reconstruction of the optimal explanation and the associated cost function adhere to the same principles and architectural design outlined in \cite{mamalakis}. The cost function consists of three key components: sparseness, as defined in \cite{spar}; ROS and RIS scores \cite{ros}; and similarity. The integration of these components ensures a robust and interpretable evaluation. 
The total cost function for training the reconstruction model is:
\begin{align}
\mathcal{L}_{\text{total}}(\boldsymbol{\phi}^{(k)}, \hat{\boldsymbol{\phi}}) = & 
\lambda_1 \cdot \frac{1}{M_{\text{RIS}}(f, \hat{\boldsymbol{\phi}})} + 
\lambda_2 \cdot \frac{1}{M_{\text{ROS}}(f, \hat{\boldsymbol{\phi}})} \notag \\
& + \lambda_3 \cdot M_{\text{sparse}}(f, \hat{\boldsymbol{\phi}}) + 
\lambda_4 \cdot \mathcal{L}_{\text{similarity}}(\hat{\boldsymbol{\phi}}, \bar{\boldsymbol{\phi}})
\end{align}

where:\( \lambda_1, \lambda_2, \lambda_3, \lambda_4 \) are hyperparameters controlling the influence of each loss term. This formulation enables a principled and quantitative integration of multiple attribution methods, optimizing toward a robust and interpretable explanation.

\subsection{The UMAP extraction and the linear constrain}

Given a dataset \(\hat{\boldsymbol{\Phi}} = \{ \hat{\boldsymbol{\phi}}_1, \hat{\boldsymbol{\phi}}_2, \dots, \hat{\boldsymbol{\phi}}_n \} \subset \mathbb{R}^D \), UMAP aims to find a low-dimensional embedding \( U = \{ u_1, u_2, \dots, u_n \} \subset \mathbb{R}^d \) where typically \( d = 2 \) or \( d = 3 \), such that the local topological structure of the data in \( \hat{\boldsymbol{\Phi}} \) is preserved in \( U \).

\textsc{High-Dimensional Graph Construction:} First, the algorithm constructs a k-nearest neighbors graph in the high-dimensional space \(\hat{\boldsymbol{\Phi}} \). The distance metric used to calculate the pairwise distances is typically Euclidean:

\[
d(\hat{\boldsymbol{\phi}}_i, \hat{\boldsymbol{\phi}}_j) = \| \hat{\boldsymbol{\phi}}_i - \hat{\boldsymbol{\phi}}_j \|_2
\]

Next, a conditional probability is defined between points \( \hat{\boldsymbol{\phi}}_i \) and \( \hat{\boldsymbol{\phi}}_j \) using a Gaussian distribution:

\[
p_{ij} = \exp\left( -\frac{\| \hat{\boldsymbol{\phi}}_i - \hat{\boldsymbol{\phi}}_j \|^2}{\sigma_i^2} \right)
\]

where \( \sigma_i \) is the bandwidth for the Gaussian distribution, determined through a binary search to match a fixed perplexity. 

The graph is symmetrized:

\[
P_{ij} = \frac{p_{ij} + p_{ji}}{2}
\]

\textsc{Low-Dimensional Embedding Graph:} In the low-dimensional space, a similar probability is defined between points \( u_i \) and \( u_j \):

\[
q_{ij} = \frac{1}{1 + a \| u_i - u_j \|^{2b}}
\]

where \( a \) and \( b \) are hyperparameters that control the shape of the distribution, and \( \| u_i - u_j \|_2 \) is the Euclidean distance between points in the low-dimensional embedding.

\textsc{Objective Function:} The optimization process involves minimizing the cross-entropy between the high-dimensional and low-dimensional probability distributions:

\[
\mathcal{L} = \sum_{i < j} \left[ P_{ij} \log(Q_{ij}) + (1 - P_{ij}) \log(1 - Q_{ij}) \right]
\]

This loss function encourages points that are close in the high-dimensional space to be close in the low-dimensional space, and points that are distant to remain distant.

\textsc{Optimization Process:} The optimization is carried out using stochastic gradient descent (SGD), updating the embedding points \( \{ u_i \} \) iteratively based on the gradient of the loss function \( \mathcal{L} \). The gradient updates for the low-dimensional embedding \( u_i \) are computed as follows:

\[
\frac{\partial \mathcal{L}}{\partial u_i} = -\sum_{j \neq i} \left( P_{ij} - Q_{ij} \right) \frac{u_i - u_j}{\| u_i - u_j \|_2^2}
\]

\textsc{Regularization Constraint:} To prevent the embedding from collapsing to a single point, we introduce a variance constraint to ensure that the variance of the embedding does not approach zero:

\[
\text{Var}(U) = \frac{1}{n} \sum_{i=1}^n \| u_i - \bar{u} \|_2^2 \geq \epsilon
\]

where \( \bar{U} = \frac{1}{n} \sum_{i=1}^n u_i \) is the mean of the embeddings, and \( \epsilon > 0 \) is a small constant that enforces a lower bound on the variance.

\textcolor{black}{{\textsc{Application of UMAP in our problem:}
To obtain a comparable low-dimensional representation of the attribution scores across
all tokenizer features, we applied a feature-wise UMAP projection procedure to the
normalized attribution matrix. For each attribution method, the attribution tensor has
shape $\mathbb{R}^{M \times T}$, where $M$ denotes the number of test samples in the
evaluation cohort and $T$ corresponds to the dimensionality of the tokenizer embedding
space of the input text. For each feature 
$j \in \{1,\dots,T\}$, we first applied min--max normalization to the
feature-specific attribution vector}
\[
\mathbf{x}^{(j)} \in \mathbb{R}^{M},
\]
and subsequently performed a one-dimensional UMAP projection to obtain a two-dimensional
embedding
\[
\mathbf{y}^{(j)} \in \mathbb{R}^{M \times 2}.
\]
The resulting coordinates were then normalized to the interval $[0,1]$ to ensure that all
feature-wise embeddings share a common bounded range. This procedure preserves the
relative neighborhood structure of the $M$-sample attribution distribution for each
feature while mapping all $T$ features into a comparable two-dimensional representation
space.}

\textcolor{black}{The motivation for applying UMAP independently to each of the $T$ tokenizer features is
to ensure that all attribution methods are projected into an aligned and comparable
representation space. Since each attribution method produces values defined over the same
token embedding dimensions, a feature-wise nonlinear projection enables consistent
cross-method comparison of attribution patterns within the shared tokenizer feature
space.}

\bigskip

\textsc{Linear Constraint for Equal Components in UMAP:}
Let \( u_i = (u_{i1}, u_{i2}, \dots, u_{id}) \) denote the embedding of the \( i \)-th data
point in a \( d \)-dimensional space. The requirement that the first and second embedding
components are equal can be written as:
\[
u_{i1} = u_{i2} \quad \forall i \in \{1, 2, \dots, n\}.
\]
Equivalently, this can be expressed as the linear equality constraint:
\[
u_{i1} - u_{i2} = 0 \quad \forall i \in \{1, 2, \dots, n\}.
\]
This constraint enforces that, for each data point \( i \), the first and second coordinates
of the embedding vector \( u_i \) are identical.

Within the total objective 
$\mathcal{L}_{\text{total}}(\boldsymbol{\phi}^{(k)}, \hat{\boldsymbol{\phi}})$ of Eq.~35, an
additional penalty term may be introduced to enforce this constraint. The penalty can be
written as:
\[
\lambda_5 \sum_{i=1}^{n} \left( u_{i1} - u_{i2} \right)^2,
\]
where \( \lambda_5 \) is a regularization parameter controlling the strength of the
constraint. This term encourages the first and second components of each reconstructed
embedding point from the optimizer $(\hat{\boldsymbol{\phi}})$ to be equal, while still
allowing flexibility depending on the value of \( \lambda_5 \).

\subsection{The superposition and the monosemantic representations}
We model an embedding space as a real vector space $\mathbb{R}^d$, where a hidden activation vector $\mathbf{h}\in\mathbb{R}^d$ represents a combination of underlying semantic features. By the linear representation hypothesis, each interpretable feature corresponds to a fixed direction in $\mathbb{R}^d$ \cite{olah2020zoom,elhage2022superposition}.

Let $\mathbf{a}\in\mathbb{R}^F$ be a sparse feature activation vector and $W\in\mathbb{R}^{d\times F}$ be a linear transformation such that:
\[
  \mathbf{h} = W\mathbf{a} = \sum_{i=1}^{F} a_i \mathbf{w}_i,
\]
where $\mathbf{w}_i$ denotes the $i$-th column of $W$, corresponding to the direction of the feature $i$.

If $F > d$, the map $W$ cannot be invertible, and thus different combination of characteristics can map to the same embedding. This gives rise to superposition, where multiple semantic features are embedded into shared subspaces or overlapping neuron activations \cite{elhage2022superposition}.

\textsc{Monosemantic Representations:} A representation is called monosemantic when each neuron corresponds to a single interpretable feature \cite{olah2020zoom}. Mathematically, this corresponds to the case where $W$ is full-rank and aligned with the identity matrix (or a rotation of it):
\[
  W = I \Rightarrow \mathbf{h} = \mathbf{a}.
\]

This implies that each feature $a_i$ is represented by a unique dimension $h_i$, with no overlap. Each neuron responds to a single, isolated concept, akin to “grandmother cells” in neuroscience \cite{quiroga2005invariant}.

\textsc{Polysemantic Representations:} In contrast, polysemantic neurons represent multiple, distinct concepts. Formally, if neuron $h_j$ computes:
\[
  h_j = \sum_{i=1}^{F} W_{j,i} a_i,
\]
and two or more $W_{j,i} \neq 0$, then neuron $j$ encodes multiple features simultaneously, exhibiting polysemanticity \cite{elhage2022superposition, bills2023language}.

More generally, a polysemantic embedding may be viewed as a mixture:
\[
  \mathbf{h} = \sum_{k=1}^{K} \alpha_k \mathbf{c}_k, \quad K > 1,
\]
where $\mathbf{c}_k$ are concept vectors and $\alpha_k$ are scalar weights.

This behavior is prevalent in both neural network activations and in biological neurons that exhibit mixed selectivity \cite{rigotti2013mixed}.

Monosemantic representations arise from disentangled bases, where neurons correspond to isolated features. Superposition emerges from dimensionality compression and necessarily leads to polysemantic neurons, each encoding a combination of features. Spare auto-encoder is a way to try to solve the polysemantic neurons—each encoding problem.

\subsection{The SAE approach and architectures}
Sparse Autoencoder (SAE) architectures have advanced our understanding of how language and vision models represent features \cite{ref15}. Neural network behavior is often explained via \textit{computational circuits}—collections of neurons that together compute meaningful functions. Classical circuit analysis has identified key components such as edge detectors \cite{ref11} or word-copying units \cite{bills2023language}. By using features derived from SAEs rather than raw neurons, researchers have improved the interpretability of circuits related to complex behaviors \cite{ref17}.

Feature discovery can involve visual analysis \cite{ref18}, manual inspection \cite{ref14}, and even assistance from large language models \cite{ref19}. Their causal role is often validated via activation interventions: modifying a feature activation vector $\mathbf{a}$ and observing predictable changes in model output \cite{ref20}.

The mathematical formulation situates SAE architectures within the theoretical framework of superposition and semantic disentanglement. By expressing hidden states as sparse linear combinations of interpretable features, SAEs bridge the gap between low-level activations and human-understandable concepts. 

\textsc{Linear Formulation of SAEs:} Let $\mathbf{x} \in \mathbb{R}^d$ denote a layer's neuron activation vector in a pretrained model. A Sparse Autoencoder learns a sparse feature representation $\mathbf{a} \in \mathbb{R}^F$ such that:

\begin{equation}
\hat{\mathbf{x}} = W \mathbf{a} + \mathbf{b}, \label{eq:reconstruction}
\end{equation}

where $W \in \mathbb{R}^{d \times F}$ is the decoder (dictionary) matrix and $\mathbf{b} \in \mathbb{R}^d$ is a learned bias term. Each column $W_{\cdot,i}$ represents the direction of feature $i$ in neuron space, and $a_i$ is its activation. This linear mapping enables complex activations to be expressed as combinations of more interpretable features.

If $F > d$, then the feature space is overcomplete, and $W$ cannot be full-rank. This leads to superposition, where multiple features overlap in the same subspace, and individual neurons encode multiple unrelated concepts \cite{elhage2022superposition}.
If $W$ is invertible and aligned to a basis, each neuron corresponds to a single feature. The representation is monosemantic and disentangled \cite{olah2020zoom}.
When $W$ has overlapping columns, neurons can respond to multiple features, yielding polysemantic behavior. That is, for some $j$, $x_j = \sum_i W_{j,i} a_i$ involves multiple nonzero terms \cite{bills2023language}.

\textsc{Variants of SAEs:} Variants of SAEs like TopK, JumpReLU, and Gated-SAEs offer increasingly precise control over the mapping between low-level activations and human-understandable concepts, enabling fine-grained analysis and intervention.

\textbf{TopK-SAEs:} Instead of using a soft sparsityconstraint (e.g., L1 regularization), TopK-SAEs enforce hard sparsity using a top-$K$ activation function:
\begin{equation}
\mathbf{a} = \text{TopK}(W_{\text{enc}}(\mathbf{x} - \mathbf{b}_{\text{dec}})), \label{eq:topk}
\end{equation}
which retains only the $K$ largest entries of the preactivation and zeros out the rest. This promotes discrete sparsity and avoids complex hyperparameter tuning.

\textbf{JumpReLU-SAEs:} JumpReLU replaces ReLU with a thresholded step function:
\begin{equation}
\text{JumpReLU}_\theta(x) = x \cdot H(x - \theta),
\end{equation}
where $H(\cdot)$ is the Heaviside step function and $\theta$ is a learnable threshold. This allows neurons to activate only above a semantic threshold, aligning with binary behavior observed in some interpretable features. However, the discontinuity makes training difficult due to non-differentiability.

\textbf{Gated-SAEs:} Gated-SAEs introduce a gating mechanism that decouples activation magnitude and presence. Let $W_{\text{mag}}$ and $W_{\text{gate}}$ be two encoders. Then the feature activation is computed as:

\begin{equation}
\mathbf{a} = \left(W_{\text{mag}}(\mathbf{x})\right) \odot H(W_{\text{gate}}(\mathbf{x}) - \theta),
\end{equation}
where $\odot$ denotes elementwise multiplication. This enables better control over when and how strongly a feature activates, making them easier to train than JumpReLU-SAEs \cite{gate}.

In this study we utilize two different architectures of SAEs the standard SAE and TopK-SAE.

\subsection{Attribution from Sparse Feature Space to Input Tokens}
Let \( \mathbf{x}_{\text{input}} \in \mathbb{R}^{d_{\text{input}}} \) 
denote the input embedding vector (e.g., LM token embeddings), 
\( \mathbf{x} = f(\mathbf{x}_{\text{input}}) \in \mathbb{R}^d \) 
the hidden layer activation of the LM, 
\( \mathbf{a} = \text{Encoder}(\mathbf{x}) \in \mathbb{R}^F \) 
the SAE sparse feature vector, and 
\( \hat{\mathbf{x}} = W \mathbf{a} + \mathbf{b} \) 
the reconstructed activation from the SAE decoder. Now suppose we have a sparse attribution vector \( \psi_i \) over features \( \mathbf{a} \), i.e., \( \psi \in \mathbb{R}^F \), where each \( \psi_i \) reflects the importance of SAE feature \( a_i \). We aim to assign importance \( \Phi_k \) to each input token dimension \( x_{\text{input},k} \).

\vspace{1em}

\textsc{Attribution flow through the encoder: } We propagate the feature attributions backward through the encoder to the input. Using the chain rule:

\begin{equation}
\Phi_k = \sum_{i=1}^F \psi_i \cdot \frac{\partial a_i}{\partial x_{\text{input},k}} = \sum_{i=1}^F \psi_i \cdot \frac{\partial a_i}{\partial \mathbf{x}} \cdot \frac{\partial \mathbf{x}}{\partial x_{\text{input},k}}
\label{eq:backprop-attribution2}
\end{equation}

where \( \frac{\partial a_i}{\partial \mathbf{x}} \) is the encoder Jacobian (SAE layer), and \( \frac{\partial \mathbf{x}}{\partial x_{\text{input},k}} \) is the LM gradient from input token to hidden layer.

This gives us a scalar attribution \( \Phi_k \in \mathbb{R} \) for each token/input embedding dimension \( k \).

This represents how much each input token contributes to the sparse SAE features that have been identified as important. In this way, we evaluate the contribution of input features based on the monosemantic behavior of the trained network's mechanism. Based on our study thus far, we will apply the six attribution methods previously discussed at two levels: from the SAE feature space to the encoder layer, and from the encoder layer to the input embedding space. This dual-level attribution analysis enables us to investigate how interpretable sparse features relate to model internals and ultimately influence the input-level representations.

To this end, we define a two-step attribution mechanism:

\subsection*{Step 1: Attribution from Sparse Features to Encoder Layer}

Let \( \boldsymbol{\psi} \in \mathbb{R}^F \) represent the importance scores of sparse features (obtained via attribution methods). We propagate these to the encoder layer as:

\begin{equation}
\boldsymbol{\phi}^{\text{enc}} = W \boldsymbol{\psi} \in \mathbb{R}^d,
\label{eq:feature-to-encoder}
\end{equation}

where \( \boldsymbol{\phi}^{\text{enc}} \) quantifies the contribution of each encoder neuron to the important SAE features.

\subsection*{Step 2: Attribution from Encoder Layer to Input}

To assign attribution scores to input dimensions, we propagate \( \boldsymbol{\phi}^{\text{enc}} \) to the input embedding via the gradient of the encoder:

\begin{equation}
\boldsymbol{\phi}^{\text{input}} = \left( \frac{\partial \mathbf{x}}{\partial \mathbf{x}_{\text{input}}} \right)^\top \boldsymbol{\phi}^{\text{enc}} \in \mathbb{R}^{d_{\text{input}}}.
\label{eq:encoder-to-input}
\end{equation}

\noindent
Alternatively, attribution methods (e.g., Integrated Gradients, SHAP) can directly estimate:
\[
\boldsymbol{\phi}^{\text{input}} = \text{AttributionMethod}(f, \mathbf{x}_{\text{input}}, \boldsymbol{\phi}^{\text{enc}})
\]

This dual-level attribution analysis allows us to connect semantically meaningful sparse features to the raw input representation space.

\section{Technical Appendix}
\subsection{Alzheimer Dataset and Preprocessing}

\subsubsection{Preprocessing}

The ADNI data ~\cite{adni} was downloaded from the Image \& Data Archive (IDA) \cite{loniida}, run by the Laboratory of Neuro Imaging (LONI) at the USC Mark and Mary Stevens Neuroimaging and Informatics Institute. The download comprised folders including information about participants' enrollment, biospecimen, assessments, medical history, imaging and study information. 
In this work, only baseline ('bl') visit data was extracted, that is - the first visit the patient underwent when joining each study. The number of unique participant's RIDs (subject's roster ID) was then recorded, and the intersection of such identifiers across the baseline datasets was calculated through an overlap matrix assessing participant coverage by considering datasets symmetrically. The obtained result, underwent precise analysis and filtering. Non-informative and administrative columns (i.e.: SOURCE, update\_stamp, SITEID, etc.) were removed across all datasets, to then perform a column-wise completeness check to retain only variables with at least 80\% of values present and to balance data availability with feature retention. By prioritizing datasets with the highest number of unique RIDs at baseline, pairwise merging based on shared RIDs was performed (i.e.: inner joins), considering the following files: ADAS, NEUROBAT, FAQ, VITALS, DXSUM. Diagnosis data was sorted chronologically according to EXAMDATE and de-duplicated so as to obtain the first - baseline - diagnosis per subject. Moreover, to ensure robust classification, this was complemented by matching data from adni\_diagnosisDXSUM files. For data augmentation purposes, demographics data was obtained from adni\_demographic\_PTDEMOG and merged according to matching RIDs. Biospecimen and medication data were filtered, cleaned and aggregated by participant - however, due to high sparsity and no adherence of column data to the completeness threshold, such information was not included in the final merge. Similarly, no genetic data was included, due to the lack of relevant biological variables with enough completeness, as remaining columns were primarily collection metadata. The final merged dataset - after excluding administrative columns - comprised 2791 unique participant RIDs with comprehensive neuropsychological, clinical, biospecimen, vital sign, and demographic data at baseline, with the following diagnosis count: 1207 patients diagnosed with Early Mild Cognitive Impairment (EMCI), 441 with Late Mild Cognitive Impairment (LMCI), and 1143 control subjects. 
\textcolor{black}{For the binary classification task, EMCI and LMCI subjects were unified into a  unique MCI cohort - mimicking AD vs CN classification, while for the three-class task, all  three subsets were retained, considering only 440 subjects per class, for balancing purposes.}
Variables from the obtained merged dataset, were mapped to their descriptions and categorical values, according to the DATADIC\_adni123GO dictionary from ADNI ~\cite{adni}. 
Text was then generated by iterating through each subject row, replacing column names with their description and appending the corresponding column value for the specific patient. Whereby  categorical values were present, they were replaced with their corresponding textual value (i.e.: " 'sex': 0 " - was transformed into "The patient's sex is: male"). 
Two distinct datasets - one for training and one for testing - were generated from the obtained final datasets, and they were split into training, testing and validation sets. 

Another dataset was utilized for further model refinement and finetuning. Specifically, the additional data was extrapolated from MRI files from the Latin American Brain Health Institute (BrainLat) dataset, a multi-site initiative that provides neuroimaging, cognitive, and clinical data across several countries in the Latin American region ~\cite{brainlat}. The data included cognition, demographic and records information of 780 subjects. A pre-processing pipeline similar to that employed for ADNI, was followed. Namely, after filtering throughout all MRI files, 760 unique and common MRI IDs - representing each subject - were identified. After dropping subjects with a higher proportion of data missing, and columns not fulfilling the completeness threshold, median imputation based on diagnosis group mean was applied for variables with less than 30\% of data missing (such as 'Age' and 'years of education' for example) with the goal of obtaining a more complete dataset. After dropping administrative and non-informative columns, the final merged dataset comprised variables deriving from cognitive tests (MOCA - Montreal Cognitive Assessment test and the IFS - INECO Frontal Screening) and participants' demographics. The diagnosis distribution of the obtained dataset was the following: 101 control subjects (CN), 109 diagnosed with Fronto-Temporal Dementia (FTD), and 118 subjects with AD. \textcolor{black}{ For the binary classification task, here AD and Fronto-Temporal dementia were unified into a unique cognitively impaired cohort, similar as to what obtained for ADNI, while for the three-class task, the original labels were retained.} The same process as for ADNI was followed to obtain textual descriptions of BrainLat patients' data, considering the related dictionary from ~\cite{brainlat}. Finally, training and testing files where obtained, whereby each class had 50 representative samples each, both for the binary and for the three-class classification. The handling of the final split into training, testing, and validation sets was handled as for ADNI. \textcolor{black}{Throughout the manuscript, the label ‘AD’ is used for convenience to denote the MCI
cohort in ADNI (in both the binary and three-class settings), and the AD+FTD cohort in
BrainLat (in the binary setting). This choice is purely notational, as the term functions as a
class label rather than a clinical diagnosis, and the emphasis is on the model’s ability to
discriminate between the defined classes.} 

\subsubsection{Demographic Comparison of Alzheimer’s Cohorts and Matched Controls}
\label{sec:demographics}

To ensure demographic comparability and reduce confounding in downstream analyses, we examined age and sex distributions across each Alzheimer’s disease (AD) cohort and control groups.

\begin{figure*}
\centering
\subfigure[{\scriptsize Age Distribution for binary classification. }\label{fig:age_distr_binary}]
{\includegraphics[width=0.375\textwidth]{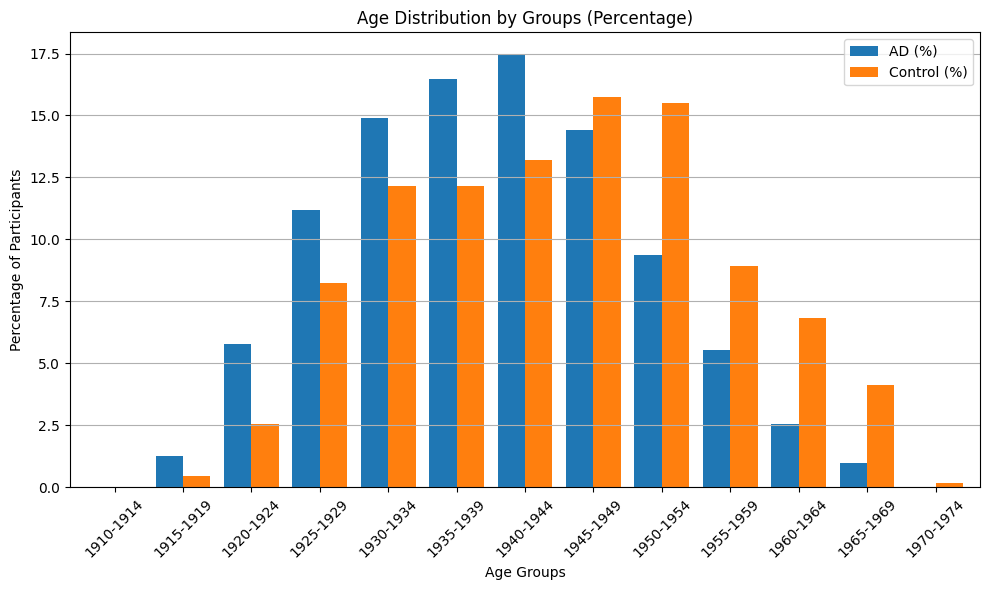}}
\hspace{0.1cm}
\subfigure[{\scriptsize Sex Distribution for binary classification. }\label{fig:sex_distr_binary}]
{\includegraphics[width=0.3\textwidth]{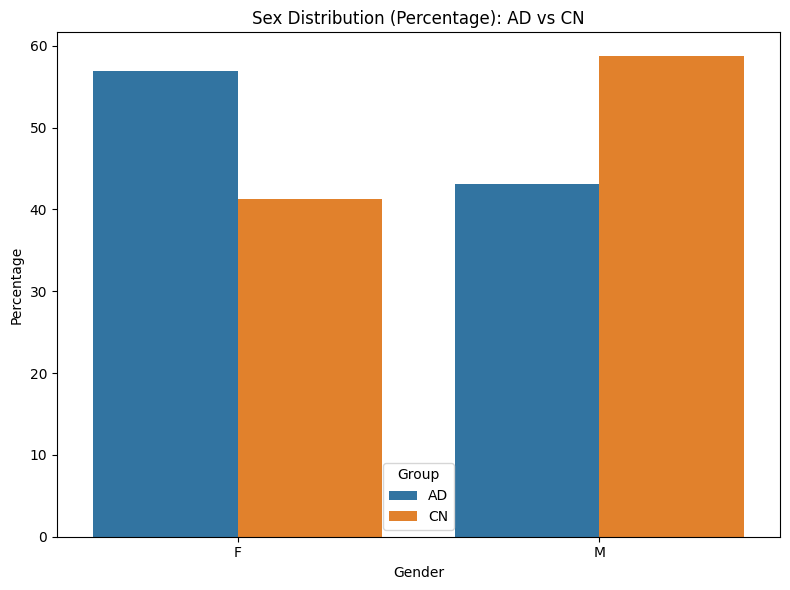}}\\[1ex]

\subfigure[{\scriptsize Age Distribution for ternary classification. }\label{fig:age_distr_ternary}]
{\includegraphics[width=0.375\textwidth]{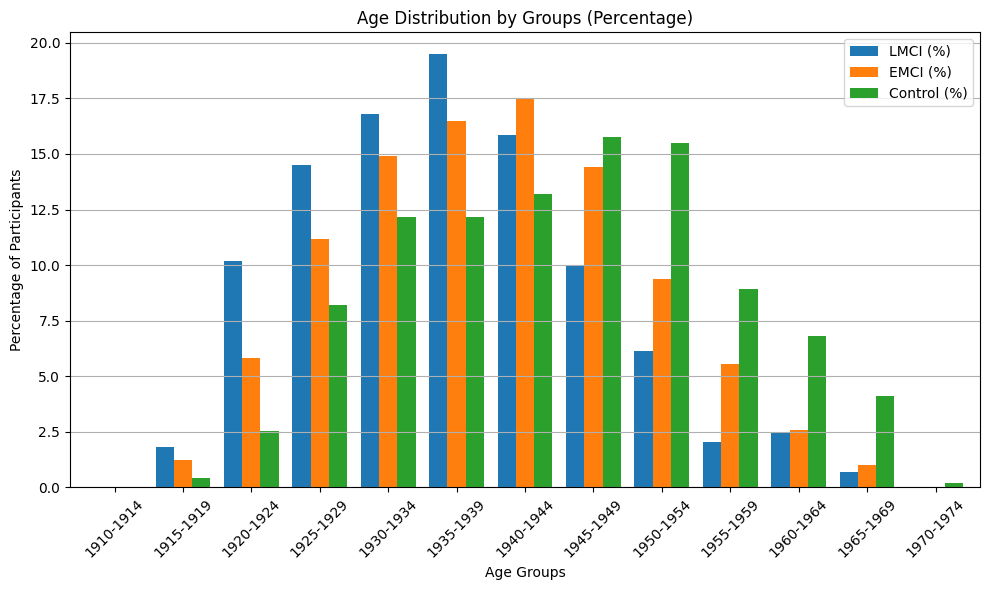}}\hspace{0.1cm}
\subfigure[{\scriptsize  Sex Distribution for three-class classification. }\label{fig:sex_distr_ternary}]
{\includegraphics[width=0.3\textwidth]{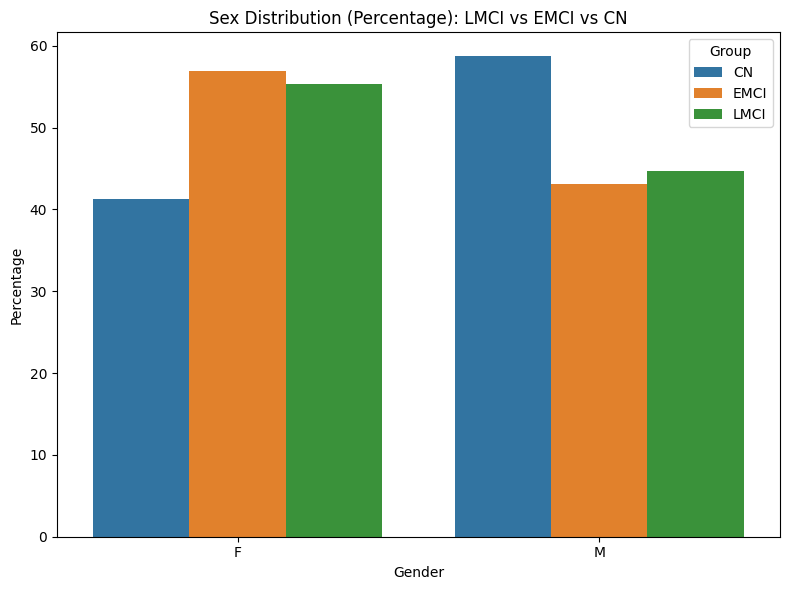}}
\caption{Demographic distributions (age and gender) for Alzheimer's cohorts and control groups for both binary and ternary classification tasks. The top row refers to the binary task, while the bottom row analyzes cohorts for the ternary classification task.}
\label{fig:demo_compare_all}
\end{figure*}

Considering the cohorts for the binary classification from ADNI ~\cite{adni}, AD subjects  ($n = 1207$) and the control group  ($n = 1143$), it is worth noting that both groups consider subjects who were born between a range that goes from the 1930s to the 1960s with comparable distributions. The AD group exhibits sharper age peaks, 
(Figure~\ref{fig:age_distr_binary}) , while the control group shows a more uniform spread. A similar pattern is evident from the three-class classification cohorts (Figure~\ref{fig:age_distr_ternary}), whereby patients diagnosed with LMCI and MCI tend to be demonstrate higher density at certain points, whereas healthy subjects' birth year distribution tends to be flatter. 

The gender distribution is uniform across groups, both in binary and three-class classification (Figures~\ref{fig:sex_distr_binary} and~\ref{fig:sex_distr_ternary}), with a slight predominance of female participants in AD groups, but overall disparity suggests minimal risk of demographic bias.

Regarding the BrainLat dataset ~\cite{brainlat}, similar patterns are evident. Control subjects are, on average, younger than subjects diagnosed with AD by 4 years, although the distribution for AD tends to be more coherently spread than the one for CN (AD cohort mean age: 71, with a standard deviation of 8.7, CN cohort mean age: 67, with standard deviation of 8.5). In the cohorts obtained for the three-class classification task, the age difference remains the same - as AD subjects tend to be the oldest, followed by those belonging to the FTD cohort and CN cohort respectively. Age variability in this case, becomes more comparable between the different diagnoses. 
Similarly to what was found for ADNI, gender-wise, the data distribution tends to be more skewed toward female participants, both in the AD and in the CN cohorts. The same is found for the subsets obtained for the three-class classification task, whereby female patients diagnosed with AD and FTD represent a higher number than male ones.

 \subsubsection{Phenotypic and Lifestyle Profiling}
\label{sec:profiling}
To characterize the ADNI  cohorts beyond age and sex, we analyzed phenotypic and lifestyle variables spanning physical health (e.g., systolic and diastolic blood pressure, respiratory and pulse rate, height, weight,  body temperature, dominant hand) and behavioral and lifestyle factors (e.g., living situation, marital status, primary language). These features were compared across all four groups to identify significant inter-group differences  ~\cite{adni}.  

In the comparison between AD and CN cohorts for the binary classification, a significant difference was found in subjects' pulse rate ($p < 0.05$) based on independent samples t-test - consistent with the nervous system dysfunction that Alzheimer's involves.   
Instead, no significance was found for systolic and diastolic blood pressure, respiratory rate, body temperature and weight.  
In terms of behavioral and lifestyle factors, a significant difference in marital status - based on Fisher's exact test - was observed between the two groups. Although most of subjects in the AD and CN groups were married, widowed individuals made up a larger proportion than divorced individuals in the AD group, while the opposite was true for CN subjects. Moreover, the CN group had a higher percentage of individuals who had never been married.
Subjects also differed for living situation (Fisher's exact test). Most subjects diagnosed with AD, lived in a house and smaller proportions lived in - respectively - a condo, an apartment, and a mobile home, with the lowest percentages residing in a retirement community and in an assisted living facility. Although CN subjects also predominantly lived in a house, they were more likely than AD subjects to live in an apartment or a condo, followed by a mobile home, an assisted living facility and lastly, a retirement community. 

\subsubsection{Assessing Compatibility Between IID and OOD Cohorts}
\label{sec:asses}

\textcolor{black}{The selection of ADNI (IID) and BrainLat (OOD) cohorts was motivated by their demographic comparability and complementary clinical profiles. As described in Section~\ref{sec:demographics}, both datasets show overlapping age and sex distributions, with balanced ratios and only minor female predominance. These similarities minimize confounding, ensuring that performance differences reflect domain shifts rather than demographic bias.}

\textcolor{black}{Phenotypic and lifestyle profiling (Section~\ref{sec:profiling}) revealed moderate inter-group differences, such as in pulse rate and marital status, consistent with disease-specific traits. ADNI primarily represents the Alzheimer’s continuum (CN, MCI, LMCI), whereas BrainLat includes FTD, AD, and controls. Despite differing diagnostic labels, these groups share clinical overlap: FTD often exhibits MCI-like cognitive decline, and LMCI represents a prodromal AD stage~\cite{petersen1999mci,jack2018niaaa,gornotempini2011ppa}.}

\textcolor{black}{This overlap establishes a natural testbed for generalization, challenging models trained on IID data to transfer to OOD settings with related but non-identical diagnoses. The IID/OOD pairing thus provides a rigorous, clinically meaningful framework to evaluate the adaptability and robustness of LM-based diagnostic systems.}

\subsubsection{Modalities subgroup extractions}
\begin{table*}
\centering
\small
\setlength{\tabcolsep}{4pt}
\renewcommand{\arraystretch}{1.05}
\caption{Variables with character counts, generation order, and estimated token usage (tokens $\approx \lceil \text{chars}/4 \rceil$).}
\label{tab:variables_tokens}
\resizebox{\textwidth}{!}{%
\begin{tabular}{@{}l p{8.0cm} r r r @{}} 
\toprule
\textbf{Variable} & \textbf{Description} & \textbf{Chars} & \textbf{Order} & \textbf{Tokens (est.)} \\
\midrule
PTGENDER & Participant sex & 30 & 1 & 8 \\
PTDOB & Date of birth & 31 & 2 & 8 \\
PTDOBYY & Year of birth & 28 & 3 & 7 \\
PTHAND & Handedness & 26 & 4 & 7 \\
PTMARRY & Marital status at baseline & 44 & 5 & 11 \\
PTEDUCAT & Education (years) & 31 & 6 & 8 \\
PTNOTRT & Retired status & 25 & 7 & 7 \\
PTHOME & Residence type & 54 & 8 & 14 \\
PTTLANG & Language used for testing & 56 & 9 & 14 \\
PTPLANG & Primary language & 39 & 10 & 10 \\
PTETHCAT & Ethnicity & 54 & 11 & 14 \\
PTRACCAT & Race (TMT) & 28 & 12 & 7 \\
PTSOURCE & Information source & 37 & 13 & 10 \\
VSWEIGHT & Weight & 32 & 14 & 8 \\
VSWTUNIT & Weight unit & 34 & 15 & 9 \\
VSBPSYS & Systolic blood pressure (mmHg) & 40 & 16 & 10 \\
VSBPDIA & Diastolic blood pressure (mmHg) & 40 & 17 & 10 \\
VSPULSE & Pulse rate (per minute) & 56 & 18 & 14 \\
VSRESP & Respiratory rate (per minute) & 51 & 19 & 13 \\
VSTEMP & Body temperature & 37 & 20 & 10 \\
VSTMPSRC & Temperature source & 32 & 21 & 8 \\
VSTMPUNT & Temperature unit & 38 & 22 & 10 \\
DXDEP & Depressive symptoms & 32 & 23 & 8 \\
CLOCKCIRC & Clock Drawing Test: circularity & 126 & 24 & 32 \\
CLOCKSYM & Clock number symmetry & 39 & 25 & 10 \\
CLOCKNUM & Clock number correctness & 31 & 26 & 8 \\
CLOCKHAND & Presence of hands & 34 & 27 & 9 \\
CLOCKTIME & Hands set to ten past eleven & 59 & 28 & 15 \\
CLOCKSCOR & Clock Drawing total score & 36 & 29 & 9 \\
COPYCIRC & Clock copying circularity & 95 & 30 & 24 \\
COPYSYM & Copy symmetry & 39 & 31 & 10 \\
COPYNUM & Copy number correctness & 31 & 32 & 8 \\
COPYHAND & Copy hand presence & 34 & 33 & 9 \\
COPYTIME & Copy time setting & 59 & 34 & 15 \\
COPYSCOR & Copy total score & 36 & 35 & 9 \\
AVTOT1 & AVLT Trial 1 total & 104 & 36 & 26 \\
AVERR1 & AVLT intrusions (Trial 1) & 19 & 37 & 5 \\
AVTOT2 & AVLT Trial 2 total & 16 & 38 & 4 \\
AVERR2 & AVLT intrusions (Trial 2) & 19 & 39 & 5 \\
AVTOT3 & AVLT Trial 3 total & 16 & 40 & 4 \\
AVERR3 & AVLT intrusions (Trial 3) & 19 & 41 & 5 \\
AVTOT4 & AVLT Trial 4 total & 16 & 42 & 4 \\
AVERR4 & AVLT intrusions (Trial 4) & 19 & 43 & 5 \\
AVTOT5 & AVLT Trial 5 total & 16 & 44 & 4 \\
AVERR5 & AVLT intrusions (Trial 5) & 19 & 45 & 5 \\
AVTOT6 & AVLT Trial 6 total & 16 & 46 & 4 \\
AVERR6 & AVLT intrusions (Trial 6) & 19 & 47 & 5 \\
AVTOTB & AVLT List B total & 15 & 48 & 4 \\
AVERRB & AVLT intrusions (List B) & 19 & 49 & 5 \\
\bottomrule
\end{tabular}}
\end{table*}
\begin{table*}
\centering
\small
\setlength{\tabcolsep}{4pt}
\renewcommand{\arraystretch}{1.05}
\ContinuedFloat
\resizebox{\textwidth}{!}{%
\begin{tabular}{@{}l p{8.0cm} r r r @{}} 
\toprule
\textbf{Variable} & \textbf{Description} & \textbf{Chars} & \textbf{Order} & \textbf{Tokens (est.)} \\
\midrule
CATANIMSC & Category fluency (animals) & 73 & 50 & 19 \\
CATANPERS & Perseverations & 17 & 51 & 5 \\
CATANINTR & Intrusions & 13 & 52 & 4 \\
TRAASCOR & Trail Making A time & 29 & 53 & 8 \\
TRAAERRCOM & TMT-A commission errors & 23 & 54 & 6 \\
TRAAERROM & TMT-A omission errors & 21 & 55 & 6 \\
TRABSCOR & Trail Making B time & 30 & 56 & 8 \\
TRABERRCOM & TMT-B commission errors & 23 & 57 & 6 \\
TRABERROM & TMT-B omission errors & 21 & 58 & 6 \\
AVDEL30MIN & AVLT 30-min delay total & 96 & 59 & 24 \\
AVDELERR1 & Delay intrusions & 19 & 60 & 5 \\
AVDELTOT & Recognition score & 20 & 61 & 5 \\
AVDELERR2 & Recognition intrusions & 19 & 62 & 5 \\
ANARTERR & ANART total errors & 81 & 63 & 21 \\
FAQFINAN & FAQ: financial management & 151 & 64 & 38 \\
FAQFORM & FAQ: forms & 58 & 65 & 15 \\
FAQSHOP & FAQ: shopping & 64 & 66 & 16 \\
FAQGAME & FAQ: games/hobbies & 68 & 67 & 17 \\
FAQBEVG & FAQ: beverage preparation & 60 & 68 & 15 \\
FAQMEAL & FAQ: meal preparation & 26 & 69 & 7 \\
FAQEVENT & FAQ: current events & 32 & 70 & 8 \\
FAQTV & FAQ: TV/book comprehension & 70 & 71 & 18 \\
FAQREM & FAQ: remembering events & 66 & 72 & 17 \\
FAQTRAVL & FAQ: travel & 121 & 73 & 31 \\
FAQTOTAL & FAQ total score & 26 & 74 & 7 \\
\bottomrule
\end{tabular}}
\end{table*}

Based on Table~\ref{tab:variables_tokens}, we extracted nine subgroups as follows: Demographics, Vital Signs, Clock Drawing Test, Clock Copying Test, Auditory Verbal Learning Test (version 1), Category Fluency – Animal Test, Auditory Verbal Learning Test (version 2), American National Adult Reading Test, and Functional Activities Questionnaire.

\subsubsection{Datasets claims}
Data used in the preparation of this article was obtained from the Alzheimer's Disease Neuroimaging Initiative (ADNI) database (adni.loni.usc.edu) on August 8th 2025 (version: "08Aug2025") and it included all ADNI phases. 
ADNI was launched in 2003 as a public-private partnership, led by Principal Investigator Michael W. Weiner, MD. It aimed at testing whether cognitive, imaging, genetic, clinical, neuropsychological assessment and other biological markers, can be combined to measure the progression of mild cognitive impairment (MCI) and early Alzheimer's disease (AD). The goals also include the validation of biomarkers for clinical trials, and the provision of data concerning the diagnosis and progression of Alzheimer’s disease to the scientific community. For up-to-date information, see adni.loni.usc.edu.

\subsection{ Summary of Training Outcomes for LLM Encoders on IID and OOD datasets}

\textcolor{black}{We systematically compared the performance of different \emph{fine-tuned encoder models} (BERT, RoBERTa, DistilBERT, ALBERT, BioBERT, ModernBERT) on \textsc{ADNI} (in-domain, IID) and evaluated cross-dataset generalization to \textsc{BrainLat} (out-of-domain, OOD). On \textsc{ADNI}, ModernBERT is the strongest encoder across all metrics: \emph{Binary}—Acc: 0.7237, F1: 0.7589, ROC-AUC: 0.8395, AUC-PR: 0.8641. \emph{Three-class}— Acc: 0.6505, F1: 0.6880, ROC-AUC: 0.7867, AUC-PR: 0.7848. BioBERT and RoBERTa are the most competitive baselines but remain below ModernBERT. In the zero-shot transfer from \textsc{ADNI} to \textsc{BrainLat} and the binary classification task, ModernBERT achieved modest performance with an average accuracy of approximately 0.55. In a representative run, the model reached 0.53 accuracy, 0.52 precision, 0.70 recall, an F1 score of 0.60, and both \textit{ROC-AUC} and \textit{AUC-PR} near 0.58. These results highlight a conservative decision threshold and the difficulty of domain transfer without adaptation. Introducing few-shot supervision improved performance moderately. In the $K$-shot regime, accuracy increased by up to 0.10 compared to zero-shot, reaching approximately 0.62 at $K=10$, with parallel gains in F1. The ROC-AUC and AUC-PR metrics remained high and stable, suggesting that limited supervision can partially mitigate domain shift but does not fully bridge the gap. LoRA-based parameter-efficient adaptation produced results comparable to few-shot training, offering efficiency in training without substantial additional gains in predictive performance.
By contrast, full fine-tuning of all pretrained weights on \textsc{BrainLat} yielded the strongest improvements, with accuracy rising to 0.84 and consistent gains across F1, ROC-AUC, and AUC-PR. These results demonstrate that full supervised adaptation remains the most effective approach to address domain shift when sufficient labeled data are available. In the three-class \textsc{BrainLat} setting, zero-shot transfer from \textsc{ADNI} yielded limited generalization (\textit{Accuracy}=0.40, \textit{F1}=0.41, \textit{ROC-AUC}=0.44), reflecting the challenge of domain and class shifts. Few-shot adaptation ($K=10$) moderately improved performance (\textit{Accuracy}=0.49, \textit{F1}=0.48), while \textit{LoRA}-based fine-tuning achieved comparable results (\textit{Accuracy}=0.50, \textit{F1}=0.48). Full fine-tuning produced the strongest gains, reaching \textit{Accuracy}=0.69, \textit{F1}=0.73, and \textit{ROC-AUC}=0.81. These findings confirm that, although limited supervision aids adaptation, full parameter optimization is essential for robust multi-class generalization across cohorts. However, this setting is outside the scope of this work: we focus on explanation performance under OOD conditions without training on the OOD cohort (i.e., without full fine-tuning). }

Therefore, for all downstream analyses we \emph{stick with} \textit{ModernBERT}: in the IID setting we use \textit{ModernBERT} fine-tuned on \textsc{ADNI} (best overall on in-domain tasks), and in the OOD setting we use \textit{ModernBERT} in a zero-shot configuration on \textsc{BrainLat} (best overall under out-of-domain conditions). All subsequent explainability analyses were conducted using the final (22\textsuperscript{nd}) layer of \textit{ModernBERT}.

\subsection{Hyperparameter tuning for the reconstruction optimizer and SAE models.}

\begin{figure*}
\centering 
\includegraphics[trim={0.0cm 2.75cm 0.0cm 0.0cm}, clip, width=\textwidth]{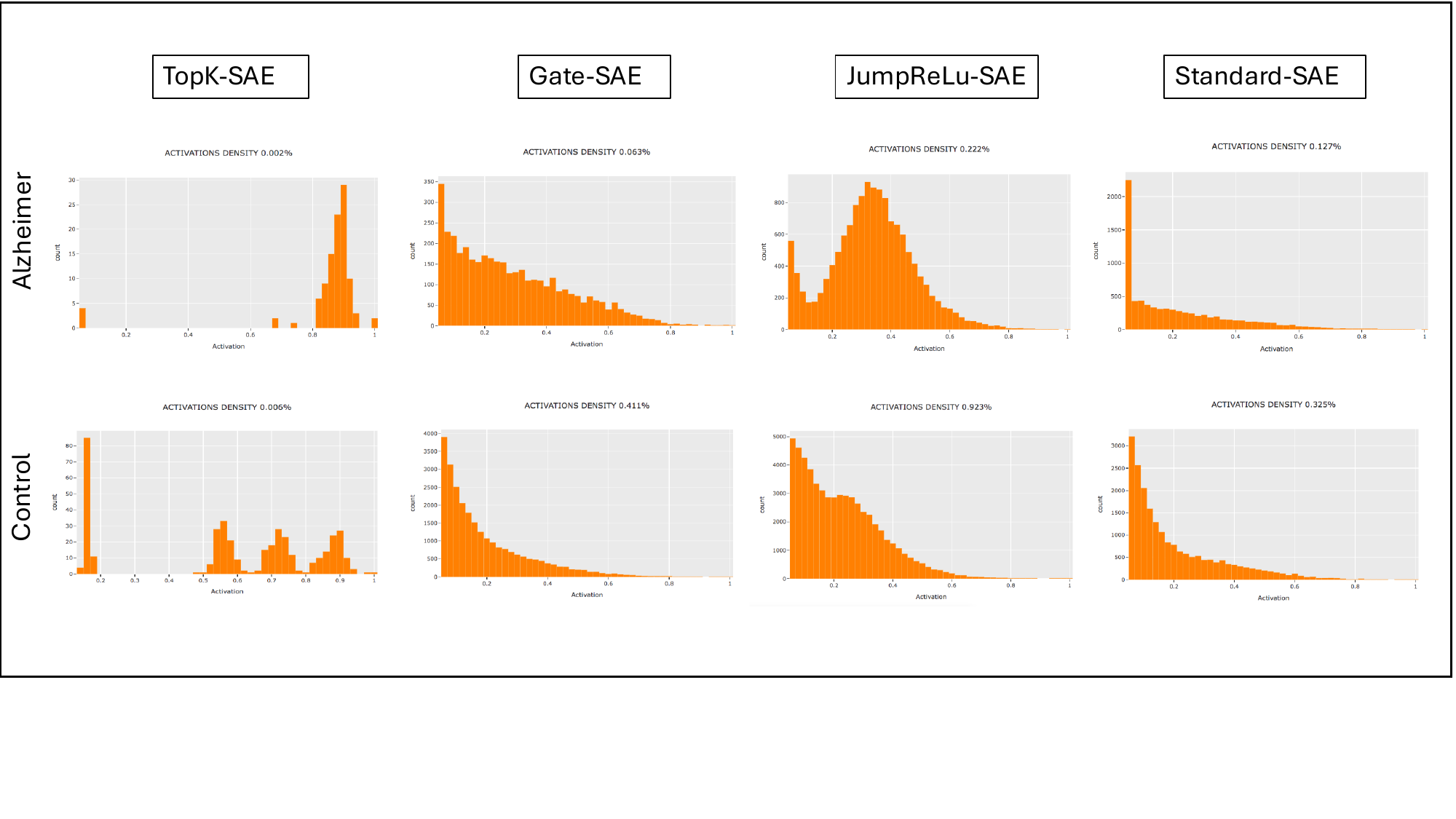}
\caption{Latent space projections from four SAE variants (TopK-SAE, Gate-SAE, JumpReLU-SAE, Standard-SAE) applied to Alzheimer’s and Control groups. TopK-SAE shows the clearest group separation, highlighting its superior ability to extract interpretable, clinically relevant features.}
\label{arcw}
\end{figure*}

\begin{figure*}
\centering 
\includegraphics[width=0.9\textwidth]{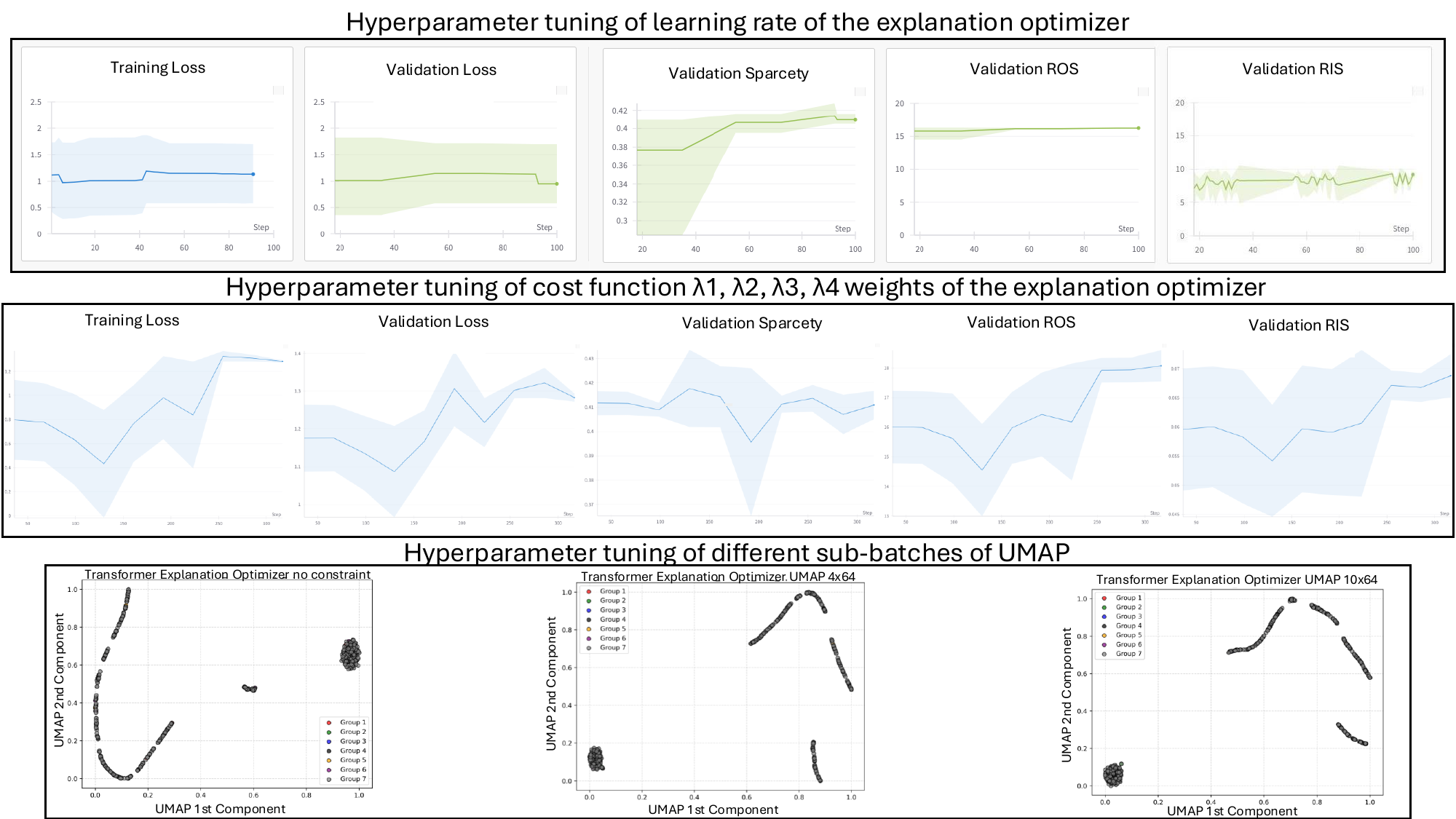}
\caption{
        \textbf{Hyperparameter tuning of the explanation optimizer and UMAP settings.}
        Top row: impact of learning rate on training loss, validation loss, sparseness, ROS, and RIS metrics.
        Middle row: sensitivity analysis of the explanation cost weights $\lambda_1$, $\lambda_2$, $\lambda_3$, and $\lambda_4$,
        showing trade-offs between attribution sparseness and robustness.
        Bottom row: UMAP projections of token-level attribution spaces under different sub-batch configurations,
        revealing how UMAP resolution influences the geometric structure of explanations.%
}
\label{arcw1}
\end{figure*}

\begin{figure*}
\centering 
\includegraphics[width=0.9\textwidth]{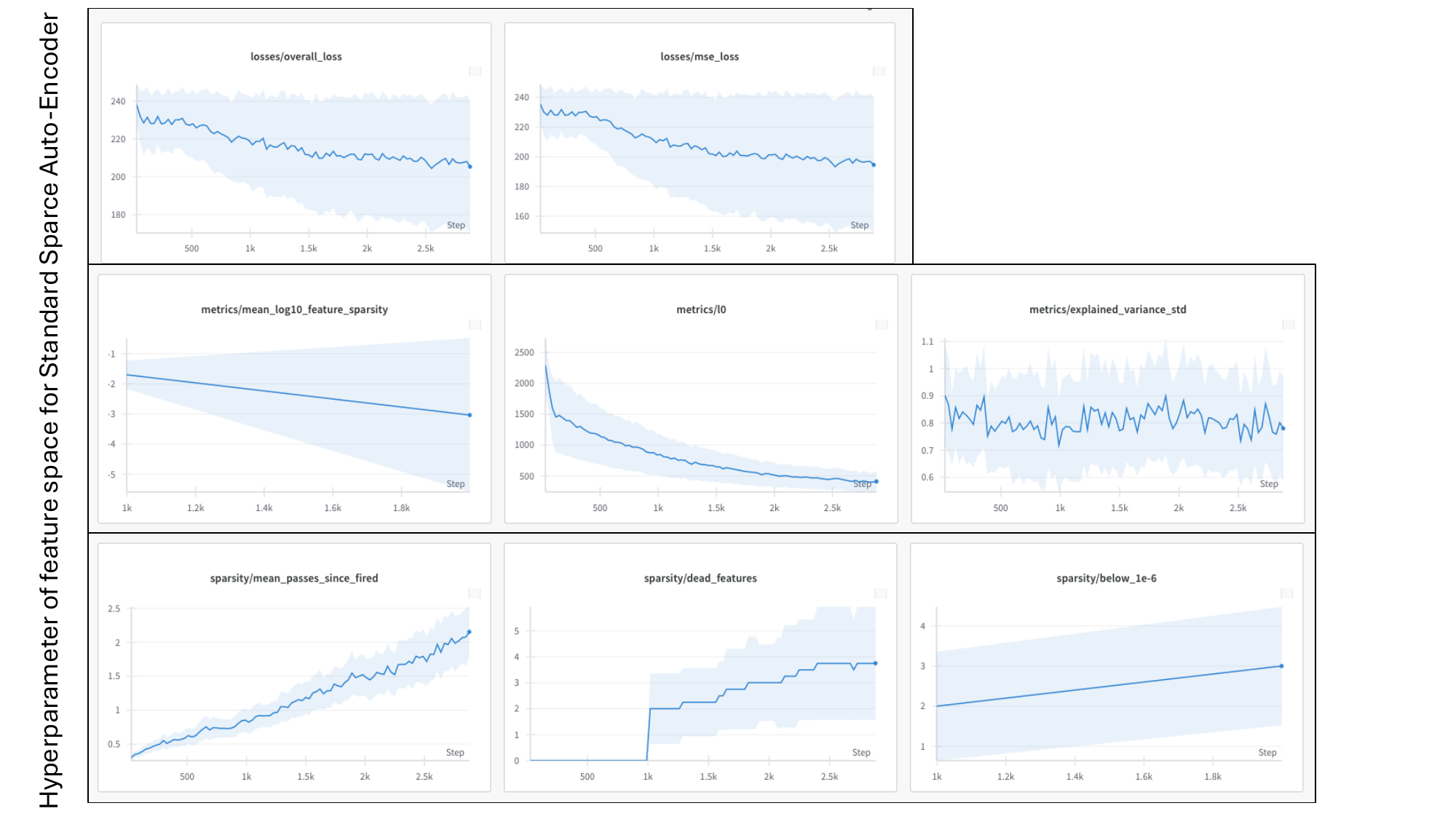}
\caption{Hyperparameter tuning of the feature space for the Standard Sparse Autoencoder (SAE).
        The plots track training dynamics and sparseness characteristics across training steps.
        Top row: loss trends for overall and reconstruction loss.
        Middle row: log-sparsity metric, Kullback–Leibler divergence (KL), and explained variance standard deviation.
        Bottom row: progression of sparsity across mean-poisson stem-freed features, fixed features, and a threshold-based view (1e-6).
        These results guide optimal SAE configurations for producing monosemantic feature representations.
}
\label{arcws}
\end{figure*}

\begin{figure*}
\centering 
\includegraphics[width=0.9\textwidth]{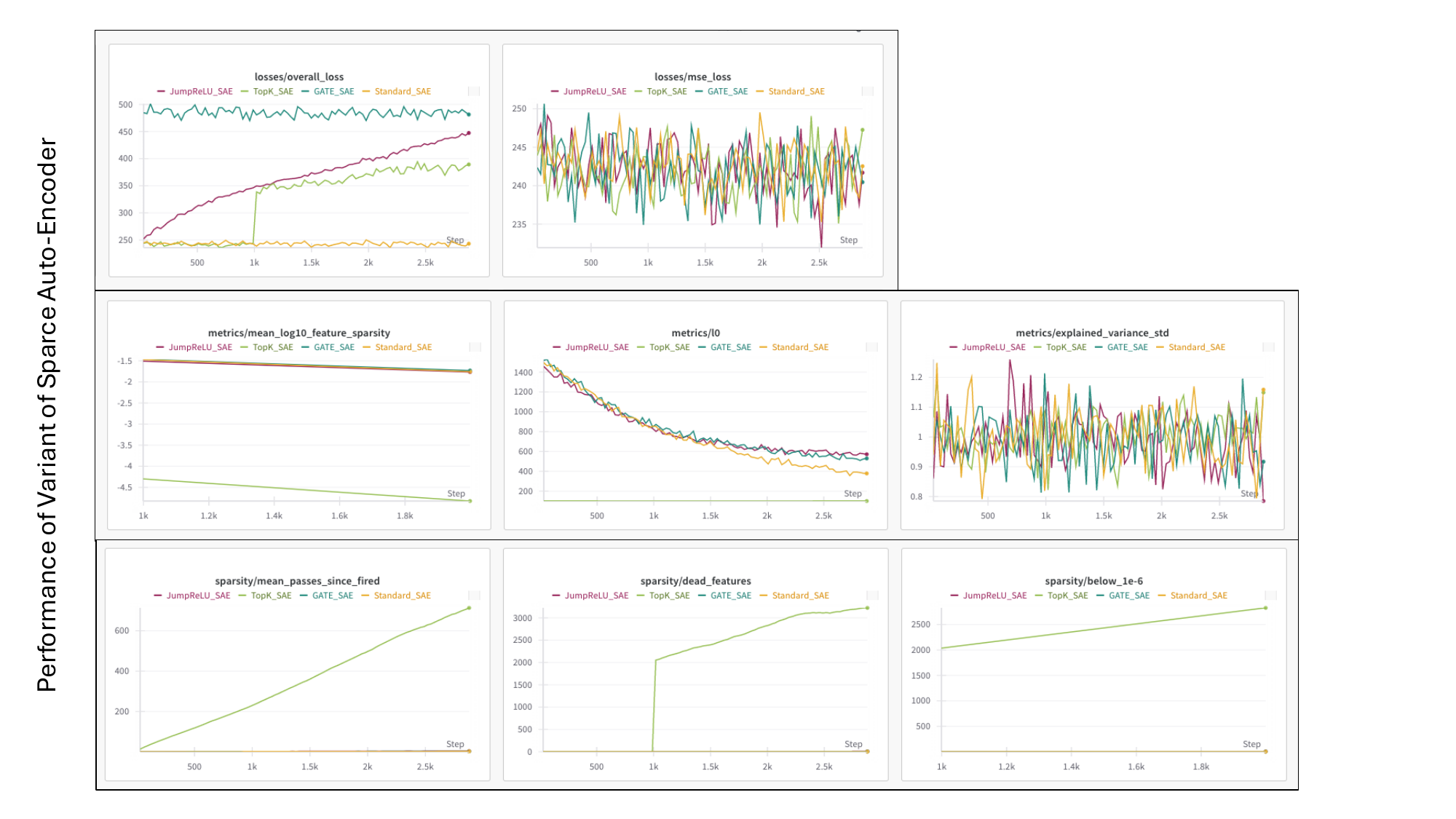}
\caption{Performance comparison of different variants of the Sparse Autoencoder (SAE).
        Top row: overall and reconstruction loss across training steps for JumpInit-SAE, Top-$k$ SAE, Gated-SAE, and Standard-SAE.
        Middle row: log-sparsity metric, KL divergence, and explained variance standard deviation, showing divergence in regularization behavior.
        Bottom row: sparsity progression for mean-poisson stem-freed features, fixed feature count, and a thresholded view (1e-6).
        JumpInit-SAE shows early convergence in sparsity, while Gated-SAE maintains tighter control over variance.
        These results highlight trade-offs between sparsity enforcement mechanisms and attributional stability.
}
\label{arcwssr}
\end{figure*}

A thorough hyperparameter tuning process was conducted for each simulation (Figures ~\ref{arcw}, \ref{arcws}, \ref{arcwssr}). The explanation optimizer was trained with learning rates of 2e\textsuperscript{-2}, 2e\textsuperscript{-3}, 2e\textsuperscript{-4}, and 2e\textsuperscript{-5}, with the best performance observed at 2e\textsuperscript{-4}. Various combinations of the weighting parameters \(\lambda_{1}, \lambda_{2}, \lambda_{3}, \lambda_{4}\) were tested---for example, (0.3, 0.2, 0.25, 0.25)---with the optimal configuration found to be (0.1, 0.3, 0.1, 0.5). For the UMAP constraints, subgroup levels were evaluated across several scales: no UMAP, every 4\(\times\) batch size, 10\(\times\) batch size, and full cohort level. The best performance was achieved at the 4\(\times\) batch size level. Regarding the SAE (Sparse Autoencoder), different model variants were evaluated, including \textit{Standard}, \textit{TopK}, \textit{JumpReLU}, and \textit{GATE}, as described in the Methods section. Among these, the \textit{TopK} variant achieved the best results. Feature space depths of 16\(\times\), 32\(\times\), and 64\(\times\) were tested, with 32\(\times\) providing the best trade-off between sparseness and reconstruction performance. The final simulation and training settings included the Adam optimizer~\cite{adam} with a learning rate of 2e\textsuperscript{-4}, a batch size of 64, and 200 total training steps, using a 50/50 train-validation split. The learning rate schedule followed a fixed-step approach with a step size of 150 and a decay factor (gamma) of 0.95. For the SAE training, we used 6{,}000 training steps, 200{,}000 training tokens, a learning rate of 5e\textsuperscript{-5}, and a model dimension of 768, consistent with the 22-layer \textit{Modern-BERT} architecture. The context size was 512, with warm-up steps of 1{,}000, learning rate decay steps of 1{,}200, and L1 warm-up steps of 300. Finally, explanation metrics such as ROS, RIS, and sparseness were computed using default configurations from the \texttt{quantus} Python package~\cite{quan}.
Figure \ref{arcw1} presents a comparative visualization of activation patterns projections generated by different Sparse Autoencoder (SAE) variants—TopK-SAE, Gate-SAE, JumpReLU-SAE, and Standard-SAE—applied to two subject groups: Alzheimer’s and Control. While the specific axes and metrics are not labeled, the separation between the two groups provides insight into the effectiveness of each SAE in producing disentangled, semantically meaningful representations. Among the models, the TopK-SAE exhibits the clearest separation between the Alzheimer’s and Control cohorts, suggesting superior performance in capturing clinically relevant patterns. This visual evidence supports the paper’s central claim that monosemantic representations enhance interpretability and robustness in clinical applications of LMs.

\subsection{Statistical Analysis}
We conducted both parametric and non-parametric statistical tests on the binary and three-class classification performance of all classes and tasks in the ADNI cohort to assess the significance of differences introduced by the monosemantic bottleneck (SAE) in traditional attribution techniques, focusing on the metrics of Sparseness, RIS, and ROS.

For the binary classification task, in both the Control and Alzheimer’s groups, paired testing demonstrated that SAE produced robust and statistically significant reductions in attribution-based measures and Complexity, while effects on RIS were smaller but still reliable, and changes in ROS were modest and often non-significant after correction. In the Control group, Complexity decreased from $0.3377 \pm 0.0017$ (no-SAE) to $0.3140 \pm 0.0010$ (SAE), yielding $t(29)=64.0$, $p=1.5\times10^{-47}$ (FDR $q<10^{-46}$), and RIS declined from $9.313 \pm 0.143$ to $9.175 \pm 0.109$, $t(29)=4.22$, $p=9.5\times10^{-5}$ (FDR $q=1.9\times10^{-4}$), both clearly rejecting the null hypothesis, whereas ROS decreased slightly from $20.616 \pm 0.203$ to $20.515 \pm 0.131$, $t(29)=2.30$, $p=0.026$ (FDR $q=0.026$), a marginal result that did not withstand correction. Attribution metrics showed the largest SAE effects: Grad-SHAP dropped from $0.4333 \pm 0.0030$ to $0.1339 \pm 0.0099$ ($p<10^{-50}$), Guided Backprop from $0.2500 \pm 0.0230$ to $0.1668 \pm 0.0072$ ($p<10^{-19}$), Integrated Gradients from $0.4304 \pm 0.0066$ to $0.0644 \pm 0.0059$ ($p<10^{-80}$), and Optimizer from $0.4199 \pm 0.0005$ to $0.2682 \pm 0.0007$ ($p<10^{-100}$), all leading to decisive rejection of the null. For the Alzheimer’s group, the same direction of effects was observed: Complexity decreased by $-0.024$ ($p<10^{-10}$), RIS by $-0.12$ ($p=4.6\times10^{-4}$), both rejecting the null, while ROS declined by $-0.09$ but did not reach significance ($p=0.073$, FDR $q=0.11$). Attribution metrics again showed dramatic reductions under SAE, with Grad-SHAP ($p<10^{-45}$), Guided Backprop ($p=3.2\times10^{-7}$), Integrated Gradients ($p<10^{-55}$), and Optimizer ($p<10^{-95}$) all supporting strong rejection of the null. Together these results demonstrate that SAE reliably improves attribution stability and reduces Complexity and RIS in both groups, with ROS showing only weak or inconsistent improvement.

For the three-class classification task, we evaluated whether SAE changed the three target metrics (Complexity, RIS, ROS) relative to no-SAE using paired \textit{t}-tests and Wilcoxon signed-rank tests for each clinical group (Control, MCI, LMCI), applying Benjamini--Hochberg FDR to control multiplicity. For the MCI group, ROS showed the clearest and most consistent improvement with SAE: the paired \textit{t}-test yielded $t(17)=-10.12$, $p=1.30\times 10^{-8}$ (FDR $q=3.90\times 10^{-8}$), and the Wilcoxon test yielded $W=0$, $p=8.0\times 10^{-6}$ (FDR $q=2.3\times 10^{-5}$), with a very large paired Cohen's $d=-2.39$ and rank-biserial correlation $r_\mathrm{rb}=-1.00$, indicating markedly lower ROS under SAE (mean difference $-0.904$; SAE $20.672$ vs.\ no-SAE $21.576$). RIS in MCI also decreased with SAE by non-parametric testing: the paired \textit{t}-test did not reach significance ($t(18)=-0.785$, $p=0.443$, FDR $q=0.443$), whereas the Wilcoxon test detected a reduction ($W=19$, $p=0.00117$, FDR $q=0.00117$), with small effect sizes ($d=-0.18$, $r_\mathrm{rb}=-0.80$; mean difference $-0.481$; SAE $10.528$ vs.\ no-SAE $11.010$). In contrast, Complexity in MCI increased with SAE according to the Wilcoxon test ($W=17$, $p=7.90\times 10^{-4}$, FDR $q=0.00117$), while the paired \textit{t}-test was non-significant ($t(18)=1.112$, $p=0.281$, FDR $q=0.421$); effect sizes were small-to-moderate ($d=0.26$, $r_\mathrm{rb}=0.821$; mean difference $+0.510$; SAE $1.175$ vs.\ no-SAE $0.665$). For the \textbf{Control} and \textbf{LMCI} groups, the pasted records contained incomplete pairs that prevented reliable paired testing and FDR-adjusted inference in the same aggregate framework; consequently, we do not report hypothesis tests for these groups here to avoid bias from unmatched rows. Overall, across the three groups, the most robust and reproducible effect we could quantify was the \emph{reduction in ROS under SAE} (clearly demonstrated in MCI with converged paired comparisons), while RIS showed a smaller SAE-related decrease by non-parametric testing and Complexity tended to increase under SAE for MCI.

\section{LM-Based External Semantic Interpretability Analysis}

\subsection{Evidence that SAE-guided attribution yields more reliable explanations than traditional attribution.}
\textcolor{black}{
To further validate the role of the SAE layer in shaping the attribution space into a more monosemantic and clinically coherent feature representation, we conducted an auxiliary evaluation. Specifically, for each test input, we extracted the top 50\% most influential token attributions produced by our attribution framework—TEO without SAE, TEO with SAE, and TEO-UMAP. We then generated a CSV file for each classification class, in which the highlighted characters for the three attribution methods were organized column-wise. The complete character sequence for each sample, beginning with the CLS token, was included in the first column to ensure clear sample-level distinction. These CSV files were subsequently provided as input to a large language model (ChatGPT-5.1~\cite{gpt}) using a fixed prompt to obtain an external, model-agnostic assessment of the interpretability structure encoded by each explanation space.}

\textcolor{black}{Three experiments were performed:  
\begin{enumerate}
    \item \textbf{Binary ADNI} (Control vs.\ Alzheimer’s disease), with each class provided as a separate CSV file. 
    \item \textbf{Binary BrainLAT} (Control vs.\ Alzheimer’s disease), also split into two class-specific CSV files.
    \item \textbf{Three-class ADNI} (Control, MCI, LMCI), with each diagnostic category represented in its own CSV file.
\end{enumerate}}

\textcolor{black}{We evaluated two primary criteria:  
(i) whether the language model could distinguish, based solely on the highlighted features, which CSV corresponded to the pathological versus the healthy control class; and  
(ii) whether the model could identify meaningful pathology-related biomarkers.}

\textcolor{black}{For the first two experiments, the model was prompted with:  
\begin{quote}
\textit{Given the two CSV files, and recognizing that medical biases exist in the \texttt{char} column with each sample beginning with the character sequence [CLS], determine how each of the three attribution methods (attr1, attr2, attr3) highlight features associated with healthy or unhealthy interpretations, and analyze the reasons for these differences. Predict the pathology and specify which of the two CSV files corresponds to the pathological case for each attribution technique.}
\end{quote}}

\textcolor{black}{For the three-class ADNI experiment, we used:  
\begin{quote}
\textit{Given the three CSV files, and recognizing that medical biases exist in the \texttt{char} column with each sample beginning with the character sequence [CLS], determine how each attribution method (attr1, attr2, attr3) highlights features associated with healthy or pathological interpretations. Predict the pathology or pathologies, identify which CSVs correspond to the pathological and healthy groups for each attribution method, and specify the associated conditions.}
\end{quote}}

\textcolor{black}{The resulting GPT-generated interpretations, shown in Figures~\ref{fig:global_explanation}, provide an external linguistic lens on each explanation space.}

\textcolor{black}{In the binary ADNI experiment, all the three attribution frameworks, TEO without SAE, TEO with SAE, and TEO-UMAP, correctly identified the pathological file (CSV1) and associated it with Alzheimer’s disease, noting that the signal was more consistent with late-stage rather than early cognitive impairment. A similar pattern emerged for the binary BrainLAT experiment: the pathological CSV was attributed to Alzheimer’s disease rather than MCI, with clear differentiation from the healthy control (Figures~\ref{fig:global_explanation}a,b). }

\textcolor{black}{Across both binary tasks, TEO without SAE exhibited erratic and clinically uninformative behavior, frequently attending to task labels, instruction counts, or other artefactual patterns rather than neurocognitive biomarkers (Figures~\ref{fig:global_explanation}a,b). In contrast, TEO-SAE and TEO-UMAP consistently highlighted clinically meaningful domains, including demographic risk factors, processing-speed impairments, and neurophysiological indicators.}

\textcolor{black}{In the more complex three-class ADNI experiment, the advantages of the SAE-induced monosemantic structure became even clearer (Figure~\ref{fig:global_explanation}c). TEO without SAE failed to correctly identify pathological classes and did not surface meaningful biomarkers. Conversely, TEO-SAE achieved clearer diagnostic separation and more coherent feature concentrations, while TEO-UMAP further emphasized structured biomarkers, particularly within demographics and vitals, and also provided correct class-level predictions.}

Collectively, these findings demonstrate that incorporating the SAE layer---thereby enforcing a more monosemantic, disentangled attribution representation---substantially enhances the clinical meaningfulness, stability, and diagnostic alignment of the resulting explanations.

\begin{figure*}
    \centering

    \subfigure[Two-class ADNI classification experiment%
    \label{fig:age_distr_binary2}]
    {
        \includegraphics[width=\textwidth]{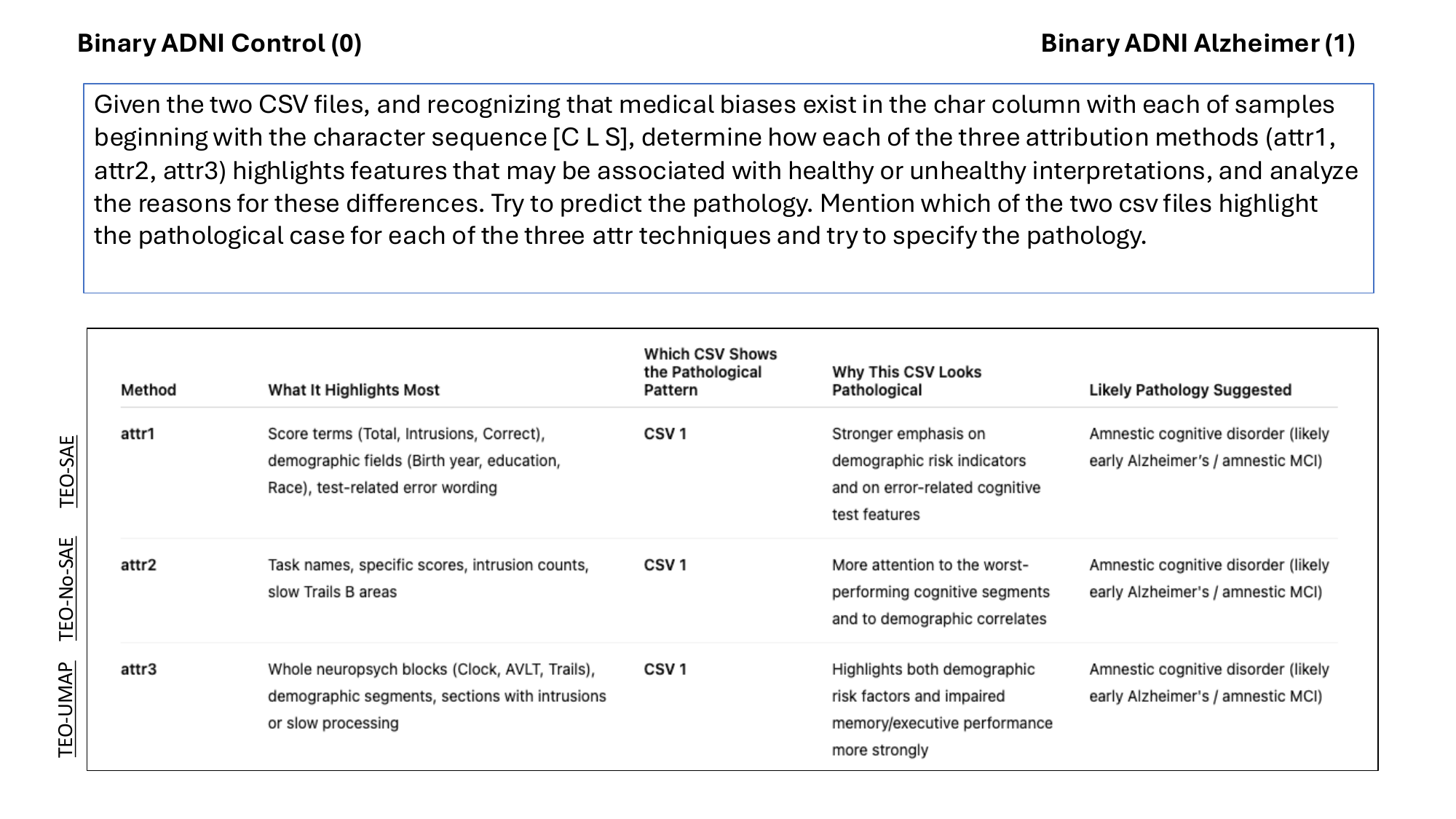}
    }

    \vspace{1ex}

    \subfigure[Two-class BrainLAT classification experiment%
    \label{fig:sex_distr_binary}]
    {
        \includegraphics[width=\textwidth]{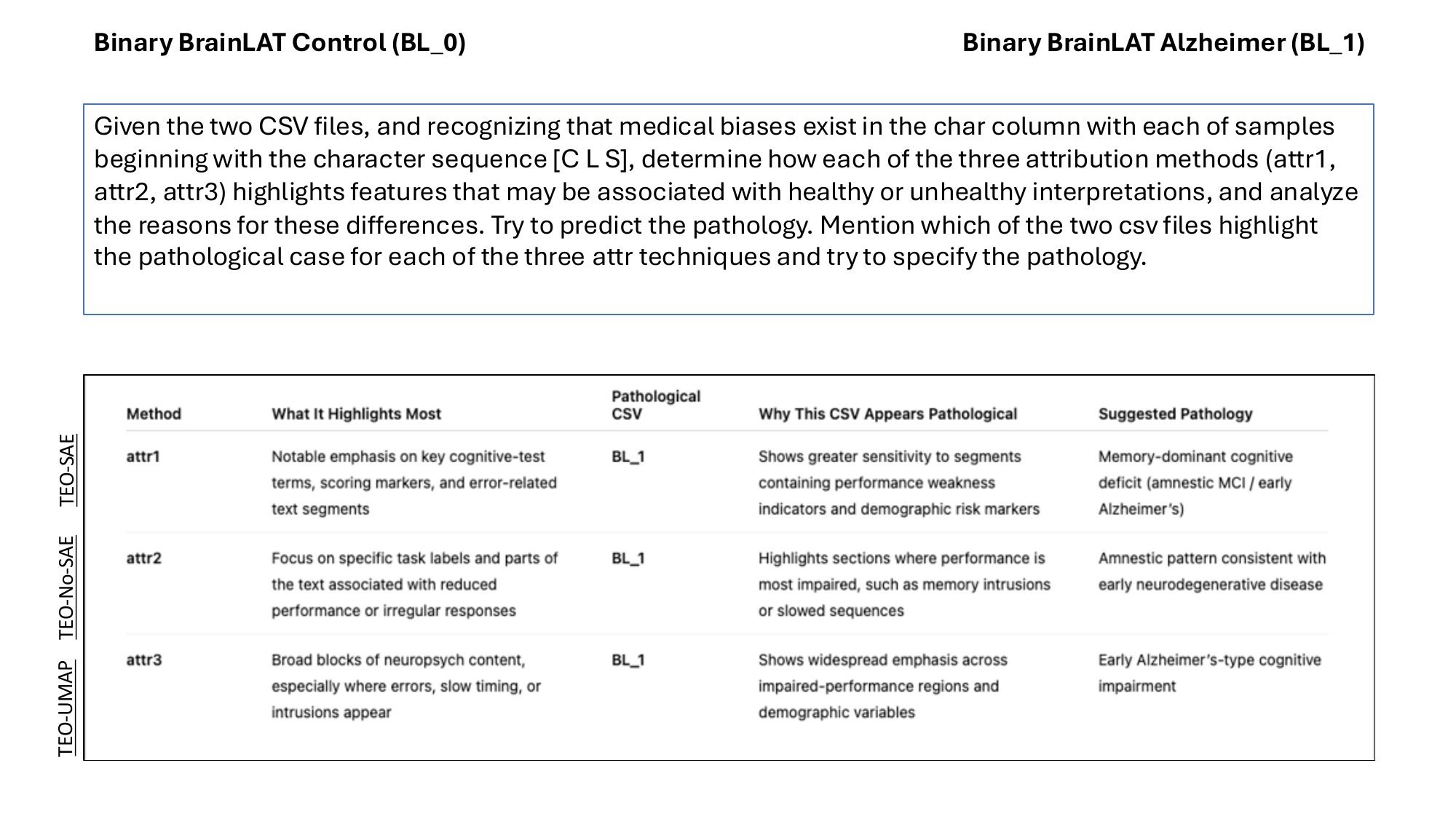}
    }

    \caption{GPT-5.1 generated global explanations characterizing attribution-score performance in biomarker identification for TEO, both with and without the SAE layer, and for the TEO-UMAP model. (a) Binary ADNI: Control vs.\ Alzheimer's disease. (b) Binary BrainLAT: Control vs.\ Alzheimer's disease.}
    \label{fig:global_explanation}
\end{figure*}

\begin{figure*}[t]
    \ContinuedFloat
    \centering

    \subfigure[Three-class ADNI classification experiment%
    \label{fig:age_distr_ternary}]
    {
        \includegraphics[width=\textwidth]{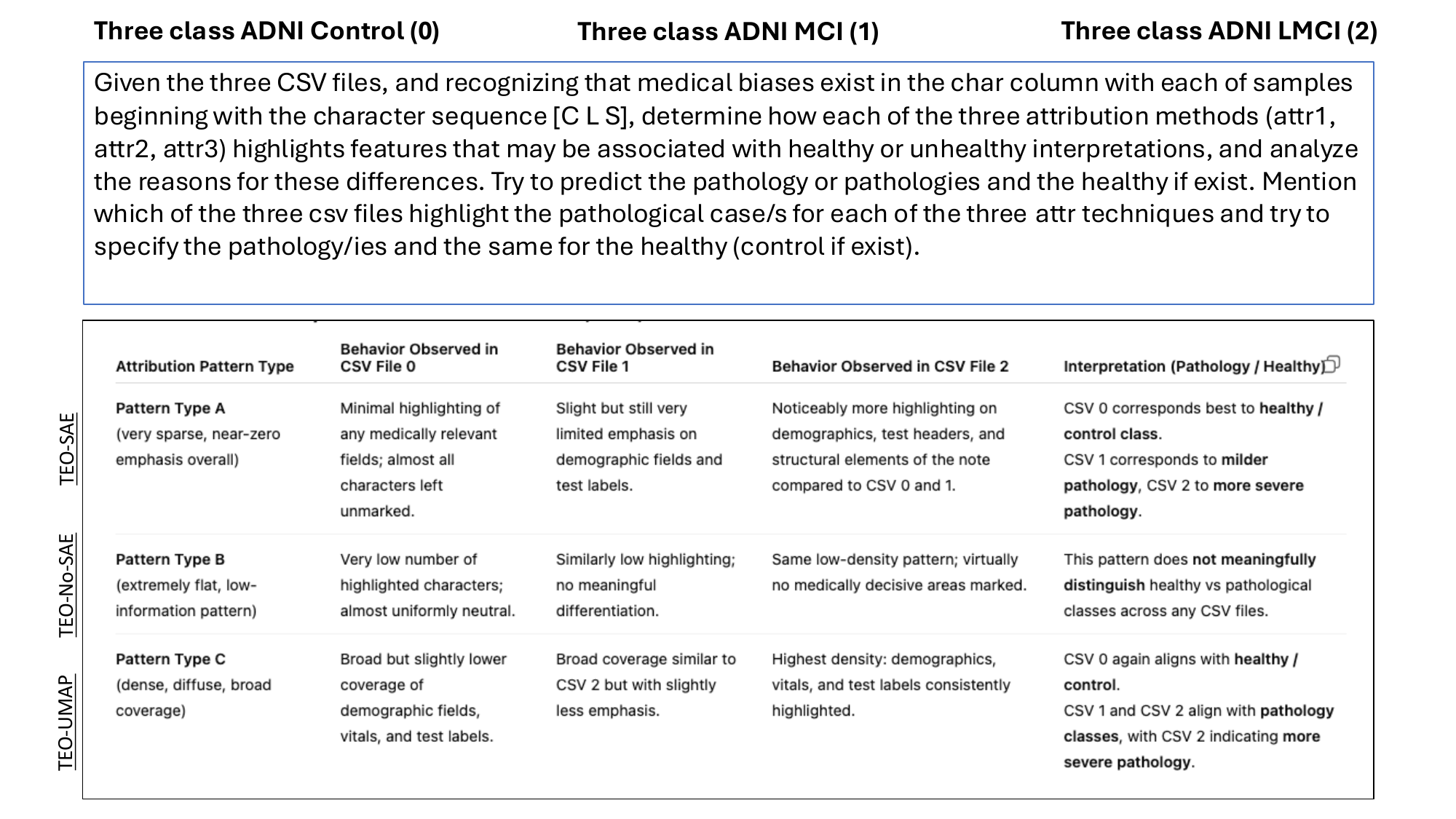}
    }

    \caption[]{GPT-5.1 generated global explanations characterizing attribution-score performance in biomarker identification for TEO, both with and without the SAE layer, and for the TEO-UMAP model. (continued) (c) Three-class ADNI: Control, MCI, and LMCI categories.}
\end{figure*}

\subsection{The clinical impact and outcome in the diagnosis of Alzheimer, early MCI and late MCI.}
This study shows that the Transformer Explanation Optimizer (TEO) with a Sparse Autoencoder (SAE) and TEO-UMAP provide the most reliable identification of informative sources across nine multimodal subgroups: Demographics (DEM), Vital Signs (VS), Clock Drawing Test (CDT), Clock Copying Test (CCT), Auditory Verbal Learning Test v1 (AVLT1), Category Fluency—Animals (CFA), Auditory Verbal Learning Test v2 (AVLT2), American National Adult Reading Test (ANART), and Functional Activities Questionnaire (FAQ). Using a significance threshold of 0.6 on PCA principal components PC1, we observe in the binary task that, for Control, TEO-SAE is dominated by FAQ, whereas TEO-UMAP emphasises DEM, AVLT2, and FAQ; for Alzheimer’s, TEO prioritises FAQ, AVLT1, and CFA, while TEO-UMAP highlights ANART, FAQ, and DEM. In the three-class task, for Control the main contributors are AVLT1, CDT, and ANART under TEO, and AVLT1, CDT, and CFA under TEO-UMAP; for MCI, TEO favours CCT, AVLT2, and FAQ, whereas TEO-UMAP favours AVLT2, ANART, and CFA; and for LMCI, TEO elevates AVLT1, FAQ, and CDT, while TEO-UMAP elevates FAQ, ANART, and AVLT2. These patterns, summarised in Table~\ref{tab:assessments}, support the clinical interpretability of the proposed optimisers.

\begin{table}
\centering
\footnotesize
\setlength{\tabcolsep}{3pt}
\renewcommand{\arraystretch}{1.05}

\caption{Relative contribution of multimodal assessment groups across diagnostic categories and attribution methods.}
\label{tab:assessments}

\begin{tabular}{lccccccccc}
\toprule
\textbf{Group} &
\textbf{Dem} & \textbf{VS} & \textbf{CDT} & \textbf{CCT} & \textbf{AVLT1} &
\textbf{CFA} & \textbf{AVLT2} & \textbf{ANART} & \textbf{FAQ} \\
\midrule
\multicolumn{10}{l}{\textit{Binary classification (Control vs.\ Alzheimer’s disease)}} \\
\midrule
Control (TEO)       & 0.00 & 0.00 & 0.00 & 0.00 & 0.02 & 0.00 & 0.00 & 0.00 & 0.11 \\
Control (TEO-UMAP)  & 0.33 & 0.21 & 0.29 & 0.29 & 0.12 & 0.30 & 0.32 & 0.30 & 0.36 \\
Alzheimer (TEO)     & 0.05 & 0.00 & 0.09 & 0.01 & 0.12 & 0.12 & 0.00 & 0.00 & 0.13 \\
Alzheimer (TEO-UMAP)& 0.33 & 0.29 & 0.30 & 0.32 & 0.22 & 0.30 & 0.32 & 0.50 & 0.35 \\
\midrule
\multicolumn{10}{l}{\textit{Three-class classification (Control / MCI / LMCI)}} \\
\midrule
Control (TEO)       & 0.74 & 0.55 & 0.87 & 0.63 & 0.88 & 0.68 & 0.20 & 0.80 & 0.38 \\
Control (TEO-UMAP)  & 0.51 & 0.61 & 0.67 & 0.60 & 0.72 & 0.66 & 0.56 & 0.60 & 0.45 \\
MCI (TEO)           & 0.23 & 0.30 & 0.29 & 0.38 & 0.22 & 0.30 & 0.36 & 0.10 & 0.31 \\
MCI (TEO-UMAP)      & 0.31 & 0.26 & 0.21 & 0.22 & 0.22 & 0.36 & 0.40 & 0.40 & 0.23 \\
LMCI (TEO)          & 0.19 & 0.19 & 0.20 & 0.10 & 0.22 & 0.16 & 0.24 & 0.10 & 0.22 \\
LMCI (TEO-UMAP)     & 0.32 & 0.27 & 0.23 & 0.25 & 0.24 & 0.28 & 0.40 & 0.40 & 0.43 \\
\bottomrule
\end{tabular}

\vspace{2pt}
\footnotesize\textit{Abbreviations:} Dem = Demographics; VS = Vital Signs; CDT = Clock Drawing Test; CCT = Clock Copying Test; AVLT1/2 = Auditory Verbal Learning Test (v1/v2); CFA = Category Fluency (Animals); ANART = American National Adult Reading Test; FAQ = Functional Activities Questionnaire.
\end{table}

Across ADNI cohorts, the most stable signals for clinical stratification are functional status (FAQ) and memory measures (AVLT1/AVLT2), with visuospatial performance (CDT) recurrent in Control/LMCI. TEO+SAE preferentially elevates neuropsychological performance features (AVLT1/2, CDT, CCT), while TEO-UMAP surfaces complementary contextual and language markers (DEM, ANART, CFA), yielding class-specific, interpretable profiles: Control---FAQ/AVLT1/CDT; Alzheimer’s---FAQ with AVLT1/CFA (TEO) or ANART/DEM (TEO-UMAP); MCI---AVLT2 with CCT/FAQ (TEO) or ANART/CFA (TEO-UMAP); and LMCI---FAQ with AVLT1/CDT (TEO) or ANART/AVLT2 (TEO-UMAP). Using a simple $PC1 \geq 0.6$ significance rule, these optimisers provide actionable attribution maps that can prioritise assessments, reduce testing burden, support trial enrichment, and guide personalised monitoring. Together, they offer a practically deployable and transparent framework for clinically meaningful multimodal reasoning in neurodegenerative disease.

\subsection{Validation of the UMAP Linear Constraint via PCA Structure Analysis}
To verify the claim that the proposed UMAP linear constraint effectively linearizes the majority of the attribution space and yields robust attribution scores within the same tokenized feature, we performed an additional PCA analysis. Specifically, we extracted the top eight principal components from the attribution matrix and visualized the first two components, which capture the highest proportion of variance in the data. This allows us to assess whether the tokenized features exhibit an approximately linear structure in their dominant statistical directions.
The results (Figure \ref{ar12sss}) demonstrate that the proposed method indeed produces an embedding that approaches a linear configuration, thereby supporting our hypothesis that the UMAP linear constraint leads to more stable and structurally consistent attribution representations. 

\section{Extra Results}
\begin{figure*}
\centering 
\includegraphics[width=0.9\textwidth]{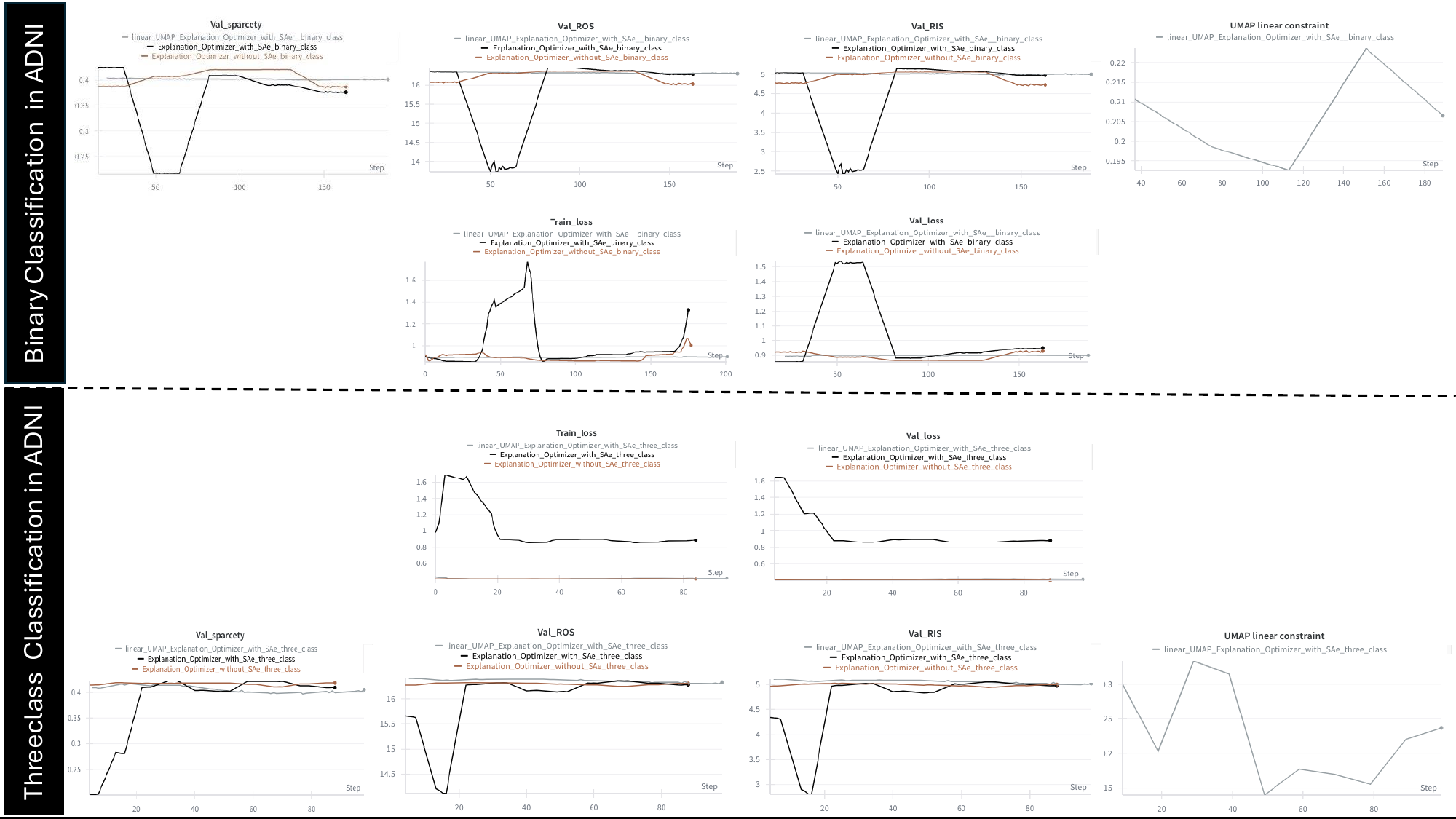}
\caption{Training and interpretability dynamics on \textsc{ADNI} for binary (top) and three-class (bottom) classification. \emph{Each subfigure includes three variants}: Explanation Optimizer \emph{without} SAE (black), \emph{with} SAE (brown), and \emph{with} SAE $+$ linear UMAP constraint (grey. For each variant we plot train loss, validation loss, UMAP reconstruction MSE (linear UMAP constrain), Relative Output Stability (ROS), Relative Input Stability (RIS), and sparcety, cohorts shown separately. SAE reduces volatility and lowers UMAP MSE and RIS/ROS versus the no-SAE baseline; adding a linear UMAP constraint on top of SAE further improves manifold structure and attribution stability, at a minor cost in sparsity.}
\label{arcwsssd}
\end{figure*}

Figure \ref{arcwsssd} compares training dynamics and interpretability metrics on \textsc{ADNI} for binary (top row) and three-class (bottom row) classification. Each subfigure shows three variants of the Explanation Optimizer: \emph{without} SAE (black), \emph{with} SAE (brown), and \emph{with} SAE $+$ linear UMAP constraint (grey). Each row in the figure presents the model’s behavior over training steps across six key metrics: train loss, validation loss, UMAP reconstruction error (MSE), Relative Output Stability (ROS), Relative Input Stability (RIS), and sparsity, . Across both tasks, all training ROS and RIS values for the SAE-based variants (brown/grey) are consistently lower than the no-SAE baseline (black), indicating improved attributional robustness. While sparseness does decrease when introducing SAE, the reduction is modest; adding the linear UMAP constraint (grey) achieves a better balance, maintaining relatively high sparsity while keeping low RIS/ROS. Finally, the training and validation curves track closely and remain smooth for the SAE variants, providing no evidence of overfitting: validation loss follows training loss without widening gaps in either the binary or three-class setting.
Overall, the SAE-enhanced Explanation Optimizer demonstrates significantly improved performance across all interpretability metrics, supporting the hypothesis that enforcing monosemantic representations improves explanation clarity and reliability—especially in high-stakes clinical contexts like Alzheimer’s disease classification.

Across IID (ADNI) and OOD (BrainLat) settings, and for both binary (Alzheimer vs.\ Control) and three-class (Control/LMCI/MCI) tasks, the tables reveal a consistent stability–sparsity frontier driven by the proposed explanation optimizers and the presence of a monosemantic bottleneck (SAE). In the binary IID case (Table~\ref{tab:sae-ordered}), SAE substantially improves stability for explainers that learn features—most notably Layer Conductance and especially TEO—with large drops in RIS/ROS for both Alzheimer and Control, while Activation with SAE increases RIS/ROS and is therefore less robust. In the binary OOD case (Table~\ref{tab:sae-ordered-updated}), these patterns persist and even strengthen: TEO with SAE bottleneck attains the lowest RIS/ROS overall, demonstrating strong cross-dataset stability, whereas TEO–UMAP recovers higher sparseness (>0.40) at the cost of higher RIS/ROS than TEO with SAE, offering a tunable sparsity–stability trade-off. In the three-class IID setting (Table~\ref{tab:three-classes}), Feature Ablation is the sparsity leader across Control/LMCI/MCI (~0.52–0.53) with moderate, steady RIS/ROS; Layer Conductance with SAE markedly reduces RIS/ROS for LMCI/MCI; and TEO with SAE again delivers the most stable attributions across all classes (lowest RIS/ROS), albeit with reduced sparseness. The same rank ordering holds OOD (Table~\ref{tab:three-classes-updated}): TEO with SAE remains the stability winner for Control/LMCI/MCI, TEO–UMAP trades some stability for additional sparsity, and Feature Ablation remains the simplest high-sparsity baseline. Throughout all tables, gradient-formulaic methods (Grad-SHAP, Guided Backprop, Integrated Gradients) show near-invariant RIS/ROS (~5.6/~16.93) regardless of SAE, class, or domain, indicating that SAE chiefly benefits learned-attribution methods. Collectively, Tables~\ref{tab:sae-ordered}, \ref{tab:sae-ordered-updated}, \ref{tab:three-classes}, and \ref{tab:three-classes-updated} support three conclusions: (i) adding an SAE bottleneck reliably lowers RIS/ROS where explanations are learned (Layer Conductance, TEO), (ii) TEO with SAE is the default when stability is paramount, while TEO–UMAP is preferred when higher sparsity is required, and (iii) the class-wise and IID to OOD behaviors are consistent, underscoring the robustness of monosemantic representations for clinical explanation.

\begin{table*}
  \caption{Evaluation scores with and without SAe. Values are mean $\pm$ std. 
  Classes: Alzheimer and Control. Metrics: Sparseness (higher is better), RIS (lower is better), ROS (lower is better).
  Column order: Alzheimer (No SAE), Alzheimer (SAE), Control (No SAE), Control (SAE).  All evaluation metrics were calculated on 200 randomly selected patients from each class (binary-class classification task) in the \textbf{ADNI} testing cohort (IID). Abreviations, DEO: Diffusion Explanation Optimizer, TEO: Transformer Explanation Optimizer, TEO-UMAP: Transformer Explanation Optimizer with UMAP constraint.}
  \label{tab:sae-ordered}
\centering
  \tiny
  \setlength{\tabcolsep}{2.2pt}
  \renewcommand{\arraystretch}{0.9}
  \resizebox{\textwidth}{!}{%
  \begin{tabular}{llcccccc}
    \toprule
    \multirow{2}{*}{Method} & \multirow{2}{*}{Metric} &
    \multicolumn{2}{c}{\textbf{Alzheimer}} & \multicolumn{2}{c}{\textbf{Control}} \\
    \cmidrule(lr){3-4}\cmidrule(lr){5-6}
     & & No SAE & SAE & No SAE & SAE  \\
    \midrule
    \multirow{3}{*}{Activation}
      & Sparseness & $0.316364045 \pm 0.007573187$ & $0.296615553 \pm 0.007063087$ & $0.256150148 \pm 0.01759176$ & $0.251987915 \pm 0.004701258$ \\
      & RIS        & $14.30227 \pm 0.368612837$    & $21.3084024 \pm 0.311021506$  & $14.23653893 \pm 0.338127875$ & $19.32752421 \pm 0.932339244$ \\
      & ROS        & $25.54851914 \pm 0.482826721$ & $32.61738364 \pm 0.307928076$ & $25.54874248 \pm 0.328628877$ & $30.6394217 \pm 0.933402318$ \\
    \midrule
    \multirow{3}{*}{Layer Condact}
      & Sparseness & $0.396588773 \pm 0.026146226$ & $0.391508745 \pm 0.007549659$ & $0.374476778 \pm 0.007092037$ & $0.247974319 \pm 0.007908586$ \\
      & RIS        & $12.39850671 \pm 2.640648847$ & $5.628509596 \pm 0.023609264$ & $5.650216593 \pm 0.0391015$  & $5.614140617 \pm 0.018407327$ \\
      & ROS        & $23.146556 \pm 1.686524073$   & $16.94708243 \pm 0.010343602$ & $16.9614573 \pm 0.031757755$ & $16.93014311 \pm 0.005191254$ \\
    \midrule
    \multirow{3}{*}{Feature Ablation}
      & Sparseness & $\mathbf{0.523581491 \pm 0.009806381}$ & $\mathbf{0.523492234 \pm 0.010441342}$ & $\mathbf{0.525551296 \pm 0.011012696}$ & $\mathbf{0.526520447 \pm 0.008837301}$ \\
      & RIS        & $23.15233791 \pm 0.819793812$ & $23.56094785 \pm 0.103321598$ & $22.56110559 \pm 0.288403561$ & $23.62208862 \pm 0.093292383$ \\
      & ROS        & $33.90884482 \pm 0.161311922$ & $34.92976979 \pm 0.074540131$ & $33.90759033 \pm 0.374703897$ & $34.9726397 \pm 0.10327352$ \\
    \midrule
    \multirow{3}{*}{Gradinet-SHAP}
      & Sparseness & $0.319169309 \pm 0.004303288$ & $0.082047681 \pm 0.015464469$ & $0.433346255 \pm 0.003004736$ & $0.133912362 \pm 0.009943283$ \\
      & RIS        & $5.623104979 \pm 0.023650419$ & $5.621792106 \pm 0.022658996$ & $5.632513362 \pm 0.023548461$ & $5.619618915 \pm 0.019004514$ \\
      & ROS        & $16.93566464 \pm 0.001800535$ & $16.93449562 \pm 1.54604\text{E-}05$ & $16.94606355 \pm 0.002246403$ & $16.93475655 \pm 5.99931\text{E-}06$ \\
    \midrule
    \multirow{3}{*}{Gradient Activation}
      & Sparseness & $0.327713636 \pm 0.03837053$  & $0.203454702 \pm 0.011686877$ & $0.249969957 \pm 0.023039465$ & $0.16678783 \pm 0.00722766$ \\
      & RIS        & $5.614858128 \pm 0.019338754$ & $5.625220574 \pm 0.021340754$ & $5.616992957 \pm 0.021780592$ & $5.617270064 \pm 0.022114697$ \\
      & ROS        & $16.93028085 \pm 0.003427604$ & $16.93433453 \pm 6.78428\text{E-}05$ & $16.934673 \pm 1.43645\text{E-}14$ & $16.93473306 \pm 4.02532\text{E-}05$ \\
    \midrule
    \multirow{3}{*}{Integrated-Gradient}
      & Sparseness & $0.298289818 \pm 0.008006549$ & $0.121161362 \pm 0.005775061$ & $0.430359021 \pm 0.006572262$ & $0.064427234 \pm 0.005909053$ \\
      & RIS        & $5.620585979 \pm 0.018022403$ & $5.622360787 \pm 0.017750318$ & $5.627793468 \pm 0.018989203$ & $5.621391149 \pm 0.016872007$ \\
      & ROS        & $16.93257772 \pm 0.001464829$ & $16.93453232 \pm 1.20129\text{E-}05$ & $16.94336632 \pm 0.002408932$ & $16.93456653 \pm 8.33237\text{E-}06$ \\
    \midrule

    \multirow{3}{*}{DEO}
      & Sparseness & $0.338261111 \pm 0.003260844$ & $0.337375 \pm 0.00290587$     & $0.337742857 \pm 0.001742551$ & $0.314044444 \pm 0.001036523$ \\
      & RIS        & $9.283888889 \pm 0.080010212$ & $9.279 \pm 0.064555158$       & $9.313125 \pm 0.142722049$    & $9.175 \pm 0.108803655$ \\
      & ROS        & $20.63421053 \pm 0.086558637$ & $20.615 \pm 0.088049029$      & $20.61588235 \pm 0.202578961$ & $20.515 \pm 0.129878486$ \\
              \midrule
    \multirow{3}{*}{TEO}
      & Sparseness & $0.421975723 \pm 0.000305212$ & $0.267210213 \pm 0.001025675$ & $0.419939638 \pm 0.00048088$ & $0.268167213 \pm 0.000728522$ \\
      & RIS        & $\mathbf{5.051961362 \pm 0.019221728}$ & $\mathbf{1.622662574 \pm 0.17080061}$  & $\mathbf{5.06881834 \pm 0.01838977}$  & $\mathbf{0.996401319 \pm 0.263922792}$ \\
      & ROS        & $\mathbf{16.35285123 \pm 0.00563874}$  & $\mathbf{12.92504253 \pm 0.170261034}$ & $\mathbf{16.37765691 \pm 0.001096906}$ & $\mathbf{12.29830928 \pm 0.261259725}$ \\
              \midrule
    \multirow{3}{*}{TEO-UMAP}
      & Sparseness & $N/A $ & $0.39891406 \pm 0.000414208$ & $N/A$ & $0.40566988 \pm 0.00031341$ \\
      & RIS        & $N/A$ & $5.439373370 \pm 0.033211570$  & $N/A$  & $5.47087230 \pm 0.17460810$ \\
      & ROS        & $N/A$  & $16.3036705 \pm 0.00333634$ & $N/A$ & $16.21021807 \pm 0.0078926$ \\
    \bottomrule
  \end{tabular}
    }%
  \vspace{2pt}
\end{table*}

\begin{table*}
  \caption{Evaluation scores with and without SAE. Values are mean $\pm$ std. 
  Classes: Alzheimer and Control. Metrics: Sparseness (higher is better), RIS (lower is better), ROS (lower is better). 
  Column order: Alzheimer (No SAE), Alzheimer (SAE), Control (No SAE), Control (SAE). All evaluation metrics were calculated on 50 randomly selected patients from each class (binary-class classification task) in the \textbf{BrainLat} testing cohort (OOD). Abreviations, TEO: Transformer Explanation Optimizer, TEO-UMAP: Transformer Explanation Optimizer with UMAP constraint.}
  \label{tab:sae-ordered-updated}
  \centering
  \scriptsize
  \setlength{\tabcolsep}{3pt}
  \begin{tabular}{llcccc}
    \toprule
    Method & Metric & Alzheimer (No SAE) & Alzheimer (SAE) & Control (No SAE) & Control (SAE) \\
    \midrule
    \multirow{3}{*}{Activation}
      & Sparseness & N/A & 0.1533105 $\pm$ 0.010287697 & N/A & 0.3965415 $\pm$ 0.030322127 \\
      & RIS        & N/A & 19.162526 $\pm$ 0.364196762  & N/A & 18.2411505 $\pm$ 0.539197156 \\
      & ROS        & N/A & 31.28270375 $\pm$ 1.541354976 & N/A & 29.04058325 $\pm$ 0.473071038 \\
    \midrule
    \multirow{3}{*}{Layer Condact}
      & Sparseness & N/A & 0.239227 $\pm$ 0.029777681   & N/A & 0.25431875 $\pm$ 0.020993978 \\
      & RIS        & N/A & 6.1620695 $\pm$ 0.149481666  & N/A & 6.21494775 $\pm$ 0.20764754 \\
      & ROS        & N/A & 16.94383425 $\pm$ 0.007089038 & N/A & 16.9402505 $\pm$ 0.004966063 \\
    \midrule
    \multirow{3}{*}{Feature Ablation}
      & Sparseness & N/A & \textbf{0.5287845 $\pm$ 0.00700955} & N/A & \textbf{0.52849725 $\pm$ 0.004358269} \\
      & RIS        & N/A & 23.5834065 $\pm$ 0.064538961  & N/A & 24.14743175 $\pm$ 0.115957516 \\
      & ROS        & N/A & 34.65309725 $\pm$ 0.252584222 & N/A & 34.961296 $\pm$ 0.220544503 \\
    \midrule
    \multirow{3}{*}{Gradinet-SHAP}
      & Sparseness & N/A & 0.120076 $\pm$ 0.014392456    & N/A & 0.057057 $\pm$ 0.027064338 \\
      & RIS        & N/A & 6.04400225 $\pm$ 0.039573077  & N/A & 6.030265 $\pm$ 0.047136969 \\
      & ROS        & N/A & 16.93474475 $\pm$ 5.76852\text{E-}05 & N/A & 16.93475925 $\pm$ 5.7373\text{E-}06 \\
    \midrule
    \multirow{3}{*}{Gradient Activation}
      & Sparseness & N/A & 0.1139535 $\pm$ 0.01766843    & N/A & 0.062973 $\pm$ 0.006903384 \\
      & RIS        & N/A & 6.032837 $\pm$ 0.027736802    & N/A & 6.0338695 $\pm$ 0.039792395 \\
      & ROS        & N/A & 16.93468825 $\pm$ 3.59398\text{E-}06 & N/A & 16.934848 $\pm$ 3.74789\text{E-}05 \\
    \midrule
    \multirow{3}{*}{Integrated-Gradient}
      & Sparseness & N/A & 0.0642685 $\pm$ 0.005166108   & N/A & 0.0143455 $\pm$ 0.000312693 \\
      & RIS        & N/A & 6.05793275 $\pm$ 0.045559192  & N/A & 6.0275535 $\pm$ 0.033936686 \\
      & ROS        & N/A & 16.9347635 $\pm$ 7.76745\text{E-}06 & N/A & 16.934873 $\pm$ 1.06145\text{E-}05 \\
    \midrule
    \multirow{3}{*}{TEO}
      & Sparseness & N/A & 0.26914625 $\pm$ 0.001645095  & N/A & 0.272516 $\pm$ 0.000382866 \\
      & RIS        & N/A & \textbf{0.683544 $\pm$ 0.667616072} & N/A & \textbf{0.47335295 $\pm$ 0.280125046} \\
      & ROS        & N/A & \textbf{11.52356 $\pm$ 0.659063208} & N/A & \textbf{11.213036 $\pm$ 0.51496551} \\
    \midrule
    \multirow{3}{*}{TEO-UMAP}
      & Sparseness & N/A & 0.398914 $\pm$ 0.00047836     & N/A & 0.40425175 $\pm$ 0.002851775 \\
      & RIS        & N/A & 5.43937375 $\pm$ 0.038349421  & N/A & 5.42815425 $\pm$ 0.194389712 \\
      & ROS        & N/A & 16.303675 $\pm$ 0.003852471   & N/A & 16.157664 $\pm$ 0.105405246 \\
    \bottomrule
  \end{tabular}
\end{table*}

\begin{table*}
  \caption{Evaluation scores with and without SAE. Values are mean $\pm$ std.
  Classes: Control, LMCI, and MCI. Metrics: Sparseness (higher is better), RIS (lower is better), ROS (lower is better). All the evaluation metrics were computed on 100 randomly selected patients from each class (three-class classification task) in the testing cohort in \textbf{ADNI} dataset (IID).
  Column order: Control (No SAE), Control (SAE), LMCI (No SAE), LMCI (SAE), MCI (No SAE), MCI (SAE). Abreviations, TEO: Transformer Explanation Optimizer, TEO-UMAP: Transformer Explanation Optimizer with UMAP constraint.}
  \label{tab:three-classes}
  \centering
  \tiny
  \setlength{\tabcolsep}{2.2pt}
  \renewcommand{\arraystretch}{0.9}
  \resizebox{\textwidth}{!}{%
  \begin{tabular}{llcccccc}
    \toprule
    \multirow{2}{*}{Method} & \multirow{2}{*}{Metric} &
    \multicolumn{2}{c}{\textbf{Control}} & \multicolumn{2}{c}{\textbf{LMCI}} & \multicolumn{2}{c}{\textbf{MCI}} \\
    \cmidrule(lr){3-4}\cmidrule(lr){5-6}\cmidrule(lr){7-8}
     & & No SAE & SAE & No SAE & SAE & No SAE & SAE \\
    \midrule
    \multirow{3}{*}{Activation}
      & Sparseness & 0.302953824 $\pm$ 0.037699004 & 0.345031647 $\pm$ 0.009533629 & 0.271540783 $\pm$ 0.038404292 & 0.264362053 $\pm$ 0.062996342 & 0.262558 $\pm$ 0.03794219 & 0.309052111 $\pm$ 0.060296425 \\
      & RIS        & 14.40424929 $\pm$ 0.165969844 & 18.99683335 $\pm$ 4.483970104 & 15.07860783 $\pm$ 1.975358824 & 18.42309221 $\pm$ 2.351829228 & 16.65684576 $\pm$ 2.82076491 & 19.42272406 $\pm$ 3.474528306 \\
      & ROS        & 25.721685 $\pm$ 0.170545275   & 30.30262794 $\pm$ 0.263575977 & 26.39512961 $\pm$ 1.972727766 & 29.73331658 $\pm$ 4.847156437 & 27.96063871 $\pm$ 2.823294834 & 30.74189461 $\pm$ 6.104484774 \\
    \midrule
    \multirow{3}{*}{Layer Condact}
      & Sparseness & 0.231524647 $\pm$ 0.009587223 & 0.331457882 $\pm$ 0.006159197 & 0.362282261 $\pm$ 0.00636308 & 0.246395316 $\pm$ 0.062800978 & 0.305312706 $\pm$ 0.007627518 & 0.292950278 $\pm$ 0.057776928 \\
      & RIS        & 5.626004529 $\pm$ 0.020886163 & 5.622249412 $\pm$ 0.014860465 & 13.14289435 $\pm$ 0.325483065 & 5.623582526 $\pm$ 0.909878534 & 6.608303235 $\pm$ 2.236340343 & 5.629072111 $\pm$ 1.130689807 \\
      & ROS        & 16.94396553 $\pm$ 0.004532066 & 16.93904894 $\pm$ 0.010478577 & 24.50603352 $\pm$ 0.411948845 & 16.93375379 $\pm$ 2.745745658 & 17.91907076 $\pm$ 2.258068808 & 16.93837556 $\pm$ 3.406698804 \\
    \midrule
    \multirow{3}{*}{Feature Ablation}
      & Sparseness & \textbf{0.523915176 $\pm$ 0.006693817} & \textbf{0.526105 $\pm$ 0.01204565} & \textbf{0.522595565 $\pm$ 0.009666847} & \textbf{0.526753263 $\pm$ 0.083976662} & \textbf{0.522188941 $\pm$ 0.009398367} & \textbf{0.525710278 $\pm$ 0.104819481} \\
      & RIS        & 23.32498194 $\pm$ 0.410939712 & 23.07664553 $\pm$ 0.140278365 & 22.24471861 $\pm$ 0.162941689 & 21.97939553 $\pm$ 3.68022107 & 23.49843053 $\pm$ 0.458719189 & 23.00058106 $\pm$ 4.540229242 \\
      & ROS        & 34.66532071 $\pm$ 0.457995824 & 34.41792582 $\pm$ 0.146270917 & 33.60637357 $\pm$ 0.161507737 & 33.30711989 $\pm$ 5.499686014 & 34.87369265 $\pm$ 0.440653725 & 34.31362572 $\pm$ 6.805012455 \\
    \midrule
    \multirow{3}{*}{Gradient-SHAP}
      & Sparseness & 0.231029588 $\pm$ 0.020644371 & 0.184435333 $\pm$ 0.014766351 & 0.129241348 $\pm$ 0.032633243 & 0.301055444 $\pm$ 0.072119068 & 0.089142118 $\pm$ 0.013138265 & 0.288114875 $\pm$ 0.061769427 \\
      & RIS        & 5.618949294 $\pm$ 0.013910382 & 5.621940533 $\pm$ 0.025299231 & 5.615247522 $\pm$ 0.014396146 & 5.621698722 $\pm$ 0.946218825 & 5.629231 $\pm$ 0.018682697 & 5.618608438 $\pm$ 1.166983293 \\
      & ROS        & 16.93582388 $\pm$ 0.002076107 & 16.934845 $\pm$ 0.0000140509 & 16.92551835 $\pm$ 0.001442338 & 16.93477544 $\pm$ 2.862502663 & 16.93917776 $\pm$ 0.002118976 & 16.93476019 $\pm$ 3.524101451 \\
    \midrule
    \multirow{3}{*}{Guided Backprop}
      & Sparseness & 0.269674235 $\pm$ 0.006145842 & 0.229587733 $\pm$ 0.00357389 & 0.383890696 $\pm$ 0.017652114 & 0.431001667 $\pm$ 0.115600632 & 0.291671824 $\pm$ 0.020033511 & 0.257909625 $\pm$ 0.109537561 \\
      & RIS        & 5.629017941 $\pm$ 0.022520137 & 5.621027267 $\pm$ 0.019431586 & 5.627154783 $\pm$ 0.021213565 & 5.629664833 $\pm$ 0.947801076 & 5.626876882 $\pm$ 0.019349509 & 5.617237063 $\pm$ 1.168361854 \\
      & ROS        & 16.934673 $\pm$ 0              & 16.9348278 $\pm$ 0.0000122544 & 16.93392839 $\pm$ 0.000750952 & 16.93466433 $\pm$ 2.862491655 & 16.93401118 $\pm$ 0.00064598 & 16.93471594 $\pm$ 3.524085042 \\
    \midrule
    \multirow{3}{*}{Integrated Gradient}
      & Sparseness & 0.045146294 $\pm$ 0.007116898 & 0.263864 $\pm$ 0.004189771 & 0.10839887 $\pm$ 0.026164396 & 0.3889465 $\pm$ 0.084086627 & 0.110163824 $\pm$ 0.015744148 & 0.266043 $\pm$ 0.090525251 \\
      & RIS        & 5.620727706 $\pm$ 0.021492346 & 5.6209158 $\pm$ 0.020949259 & 5.609370435 $\pm$ 0.017753051 & 5.628201333 $\pm$ 0.947585532 & 5.628253647 $\pm$ 0.020904475 & 5.620282063 $\pm$ 1.168468694 \\
      & ROS        & 16.93310612 $\pm$ 0.001097405 & 16.93475353 $\pm$ 0.0000121647 & 16.92756339 $\pm$ 0.000606244 & 16.93434683 $\pm$ 2.862457357 & 16.93584506 $\pm$ 0.001969998 & 16.93460281 $\pm$ 3.524039 \\
    \midrule
    \multirow{3}{*}{TEO}
      & Sparseness & 0.391835667 $\pm$ 0.000814648 & 0.268163938 $\pm$ 0.064942517 & 0.413087063 $\pm$ 0.000325772 & 0.285971625 $\pm$ 0.037421241 & 0.390886118 $\pm$ 0.004742559 & 0.283782105 $\pm$ 0.052291259 \\
      & RIS        & \textbf{4.807986067 $\pm$ 0.018432249} & \textbf{1.546787813 $\pm$ 0.11712595} & \textbf{5.093836 $\pm$ 0.018806243} & \textbf{2.264221 $\pm$ 0.487706388} & \textbf{4.828254294 $\pm$ 0.037727688} & \textbf{2.161698368 $\pm$ 0.454717751} \\
      & ROS        & \textbf{16.1172194 $\pm$ 0.008995191}  & \textbf{12.856954 $\pm$ 0.117943866}  & \textbf{16.40431106 $\pm$ 0.002382358} & \textbf{13.56455925 $\pm$ 2.274547046} & \textbf{16.13535541 $\pm$ 0.032411959} & \textbf{13.46760021 $\pm$ 2.764087874} \\
              \midrule
    \multirow{3}{*}{TEO-UMAP}
      & Sparseness & $N/A $ & $ 0.39734881 \pm 0.07492051 $ & $N/A$ &  $0.41611163 \pm 0.08698112$ & $N/A$ & $0.41715421 \pm 0.237175073$\\
       
      & RIS        & $N/A$ & $5.10862522 \pm 0.20827341$  & $N/A$  & $ 5.10165242 \pm 0.16974677 $ & $N/A$ & $ 5.11160575 \pm 0.1072146 $\\

      & ROS        & $N/A$  & $16.412262 \pm 6.84387466  $ & $N/A$ & $16.4031485 \pm 3.86158978$ & $N/A$ & $ 16.4088329 \pm 0.492439623$\\
    \bottomrule
  \end{tabular}
    }%
  \vspace{2pt}
\end{table*}

\begin{table*}
  \caption{Evaluation scores with and without SAE. Values are mean $\pm$ std.
  Classes: Control, LMCI, and MCI. Metrics: Sparseness (higher is better), RIS (lower is better), ROS (lower is better).
  Column order: Control (No SAE), Control (SAE), LMCI (No SAE), LMCI (SAE), MCI (No SAE), MCI (SAE). All evaluation metrics were calculated on 50 randomly selected patients from each class (three-class classification task) in the \textbf{BrainLat} testing cohort (OOD). Abreviations, TEO: Transformer Explanation Optimizer, TEO-UMAP: Transformer Explanation Optimizer with UMAP constraint.}
  \label{tab:three-classes-updated}
  \centering
  \tiny
  \setlength{\tabcolsep}{2.2pt}
  \renewcommand{\arraystretch}{0.9}
  \resizebox{\textwidth}{!}{%
  \begin{tabular}{llcccccc}
    \toprule
    \multirow{2}{*}{Method} & \multirow{2}{*}{Metric} &
    \multicolumn{2}{c}{\textbf{Control}} & \multicolumn{2}{c}{\textbf{LMCI}} & \multicolumn{2}{c}{\textbf{MCI}} \\
    \cmidrule(lr){3-4}\cmidrule(lr){5-6}\cmidrule(lr){7-8}
     & & No SAE & SAE & No SAE & SAE & No SAE & SAE \\
    \midrule
    \multirow{3}{*}{Activation}
      & Sparseness & N/A & 0.4504596 $\pm$ 0.037517308 & N/A & 0.1907182 $\pm$ 0.001584043 & N/A & 0.140032667 $\pm$ 0.012393238 \\
      & RIS        & N/A & 19.0939584 $\pm$ 0.179649958 & N/A & 18.6406378 $\pm$ 0.911739674 & N/A & 18.07866133 $\pm$ 0.031343749 \\
      & ROS        & N/A & 29.9240256 $\pm$ 0.253797244 & N/A & 29.5628004 $\pm$ 0.89780058 & N/A & 28.858315 $\pm$ 0.045164984 \\
    \midrule
    \multirow{3}{*}{Layer Condact}
      & Sparseness & N/A & 0.3252352 $\pm$ 0.014541625 & N/A & 0.185706 $\pm$ 0.007343355 & N/A & 0.200561 $\pm$ 0.011873024 \\
      & RIS        & N/A & 6.2120432 $\pm$ 0.245036193 & N/A & 6.054585 $\pm$ 0.047710839 & N/A & 6.268400333 $\pm$ 0.040128533 \\
      & ROS        & N/A & 16.9581856 $\pm$ 0.017257696 & N/A & 16.9636788 $\pm$ 0.007069011 & N/A & 17.01458633 $\pm$ 0.020606806 \\
    \midrule
    \multirow{3}{*}{Feature Ablation}
      & Sparseness & N/A & \textbf{0.5280516 $\pm$ 0.00583783} & N/A & \textbf{0.526218 $\pm$ 0.005664214} & N/A & \textbf{0.529347333 $\pm$ 0.011164491} \\
      & RIS        & N/A & 22.640559 $\pm$ 0.033054016 & N/A & 23.5692968 $\pm$ 0.057715633 & N/A & 23.59159233 $\pm$ 0.111776179 \\
      & ROS        & N/A & 33.485311 $\pm$ 0.233803422 & N/A & 34.5168758 $\pm$ 0.073725495 & N/A & 34.37202467 $\pm$ 0.080383672 \\
    \midrule
    \multirow{3}{*}{Gradinet-SHAP}
      & Sparseness & N/A & 0.195138 $\pm$ 0.026410904 & N/A & 0.0637004 $\pm$ 0.026472282 & N/A & 0.113682167 $\pm$ 0.042008189 \\
      & RIS        & N/A & 6.122703167 $\pm$ 0.161879331 & N/A & 6.0314946 $\pm$ 0.029788567 & N/A & 6.127377667 $\pm$ 0.120700168 \\
      & ROS        & N/A & 16.93462983 $\pm$ 4.27103\text{E-}05 & N/A & 16.9348698 $\pm$ 7.57377\text{E-}05 & N/A & 16.93466933 $\pm$ 8.86649\text{E-}05 \\
    \midrule
    \multirow{3}{*}{Gradient Activation}
      & Sparseness & N/A & 0.177242 $\pm$ 0.011243388 & N/A & 0.1835882 $\pm$ 0.001632398 & N/A & 0.430289167 $\pm$ 0.00215046 \\
      & RIS        & N/A & 6.123393833 $\pm$ 0.191171312 & N/A & 6.0269246 $\pm$ 0.030177701 & N/A & 6.144976167 $\pm$ 0.120931658 \\
      & ROS        & N/A & 16.93457917 $\pm$ 1.47434\text{E-}05 & N/A & 16.9347678 $\pm$ 2.58844\text{E-}06 & N/A & 16.934534 $\pm$ 2.79285\text{E-}05 \\
    \midrule
    \multirow{3}{*}{Integrated-Gradient}
      & Sparseness & N/A & 0.067058 $\pm$ 0.012083245 & N/A & 0.0071952 $\pm$ 0.000900684 & N/A & 0.036059 $\pm$ 0.004866926 \\
      & RIS        & N/A & 6.1224575 $\pm$ 0.150190804 & N/A & 6.035594 $\pm$ 0.01894045 & N/A & 6.147797667 $\pm$ 0.09243145 \\
      & ROS        & N/A & 16.93462633 $\pm$ 1.30486\text{E-}05 & N/A & 16.9347694 $\pm$ 1.14018\text{E-}06 & N/A & 16.934618 $\pm$ 8.89944\text{E-}06 \\
    \midrule
    \multirow{3}{*}{TEO}
      & Sparseness & N/A & 0.416191667 $\pm$ 0.002863111 & N/A & 0.3715978 $\pm$ 0.000948703 & N/A & 0.42242125 $\pm$ 0.000173513 \\
      & RIS        & N/A & 5.752004667 $\pm$ 0.364536772 & N/A & 4.9396128 $\pm$ 0.014829469 & N/A & \textbf{5.5420845 $\pm$ 0.061116734} \\
      & ROS        & N/A & \textbf{16.379228 $\pm$ 0.003422144} & N/A & \textbf{15.8121004 $\pm$ 0.00994492} & N/A & \textbf{16.277279 $\pm$ 0.001043319} \\
    \midrule
    \multirow{3}{*}{TEO-UMAP}
      & Sparseness & N/A & 0.423819167 $\pm$ 0.00056124 & N/A & 0.423865 $\pm$ 5.01946\text{E-}05 & N/A & 0.424567429 $\pm$ 0.000241638 \\
      & RIS        & N/A & \textbf{5.552539167 $\pm$ 0.186236877} & N/A & 5.458319 $\pm$ 0.029725596 & N/A & 5.557568429 $\pm$ 0.095350875 \\
      & ROS        & N/A & \textbf{16.3661035 $\pm$ 0.00731861} & N/A & 16.372555 $\pm$ 0.001689458 & N/A & 16.357192 $\pm$ 0.003475647 \\
    \bottomrule
  \end{tabular}
    }%
  \vspace{2pt}
\end{table*}


\subsection{Individual-Level Explanations and Patterns}
 \begin{figure*} 
\centering 
\includegraphics[width=0.9\textwidth]{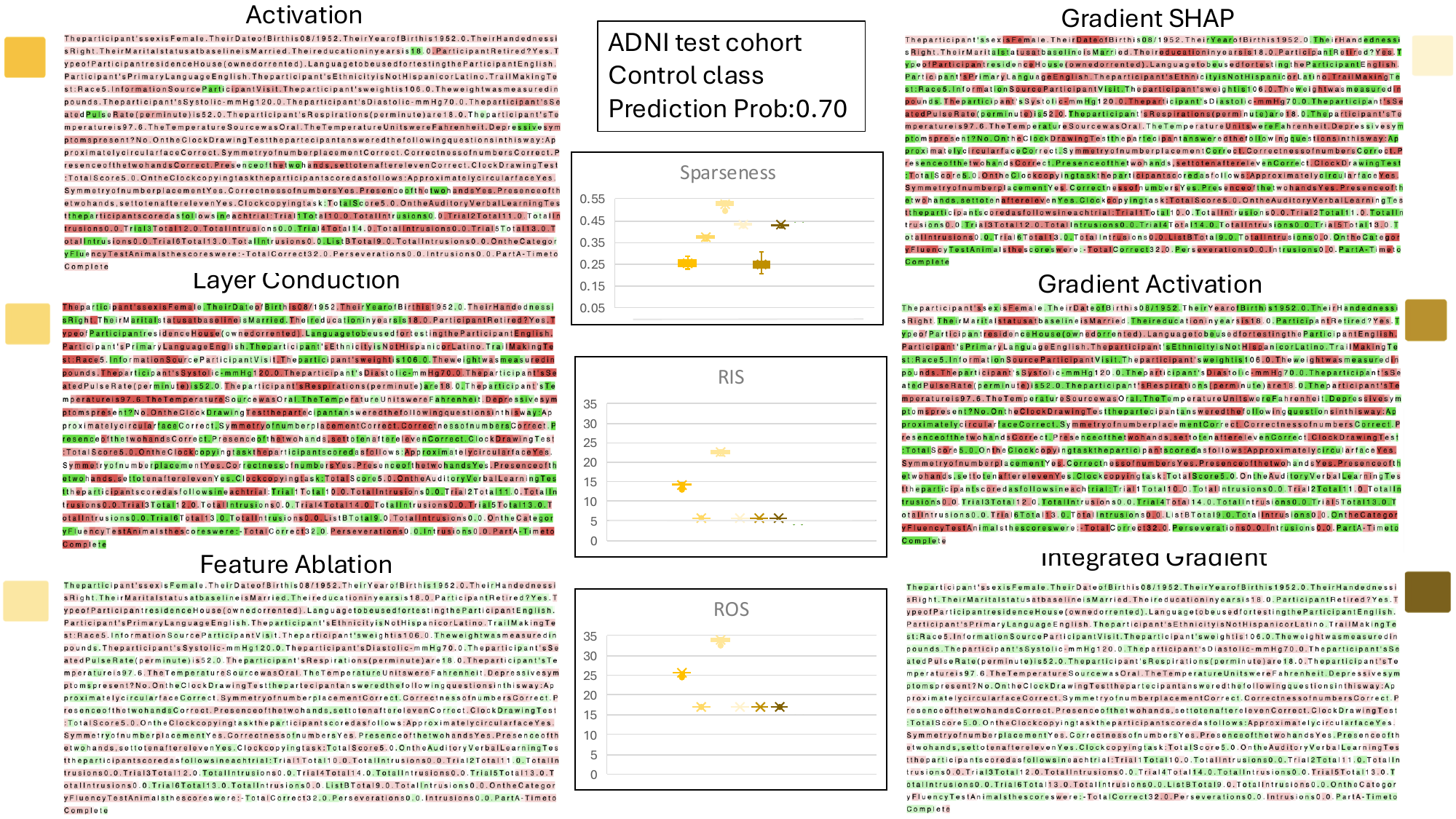}
\caption{Local attribution examples across different explanation methods without the SAE layer. The colour scale ranges from $-1$ (dark red; negative attribution), through 0 (white; neutral), to $+1$ (dark green; positive attribution). For each of the six panels, the small colour swatches at the top-left and top-right indicate the colour keys used for the three summary box-plot metrics—Sparseness, RIS, and ROS—for the corresponding attribution technique. The task is a binary classification (Alzheimer’s disease vs Control) on the ADNI cohort; the examples shown here are from the Control class.}
\label{arcx1}
\end{figure*}
 \begin{figure*} 
\centering 
\includegraphics[width=0.9\textwidth]{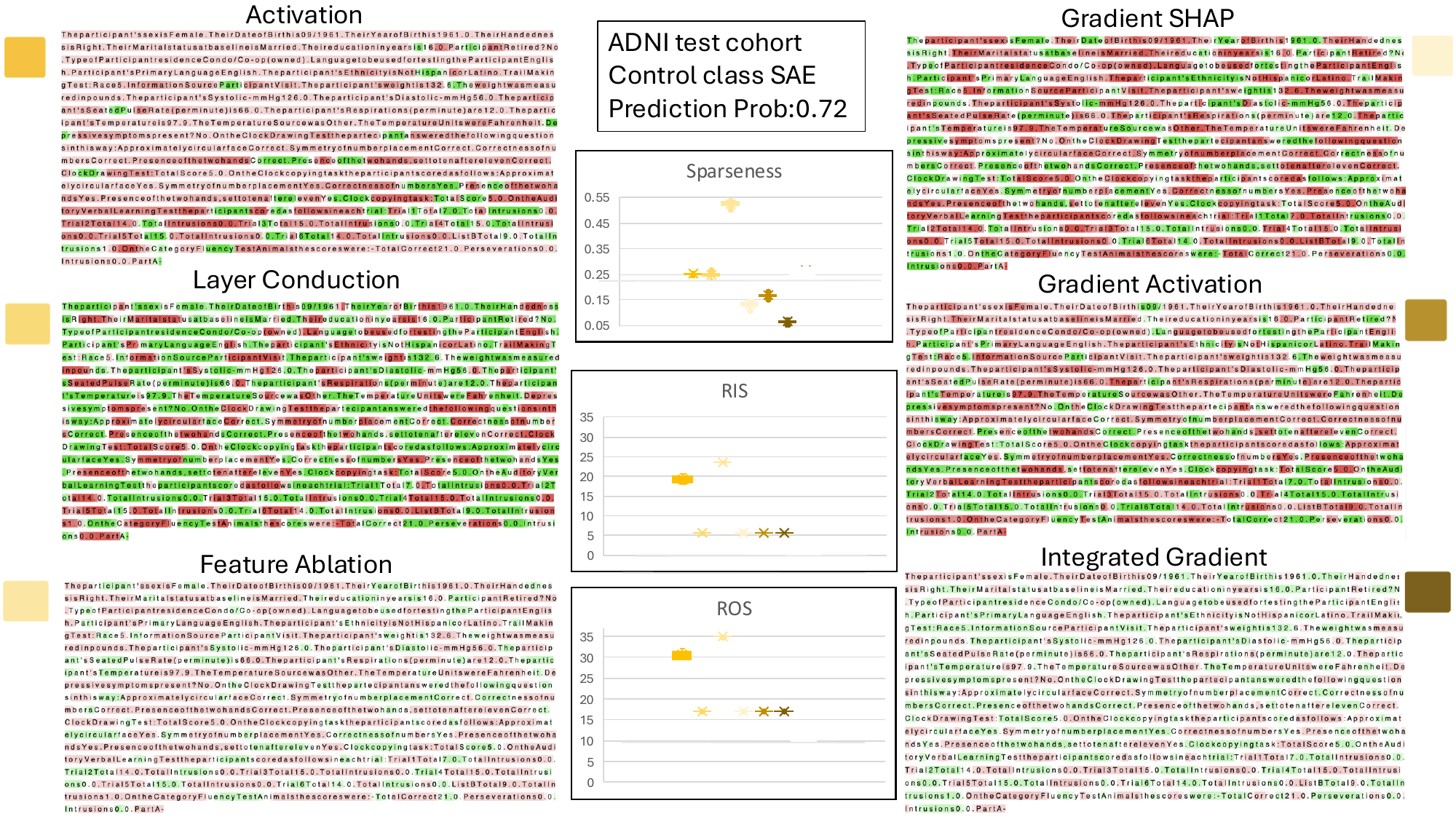}
\caption{Local attribution examples across different explanation methods with the SAE layer. The colour scale ranges from $-1$ (dark red; negative attribution), through 0 (white; neutral), to $+1$ (dark green; positive attribution). For each of the six panels, the small colour swatches at the top-left and top-right indicate the colour keys used for the three summary box-plot metrics—Sparseness, RIS, and ROS—for the corresponding attribution technique. The task is a binary classification (Alzheimer’s disease vs Control) on the ADNI cohort; the examples shown here are from the Control class.}
\label{arcx2}
\end{figure*}
 \begin{figure*}
\centering 
\includegraphics[width=0.9\textwidth]{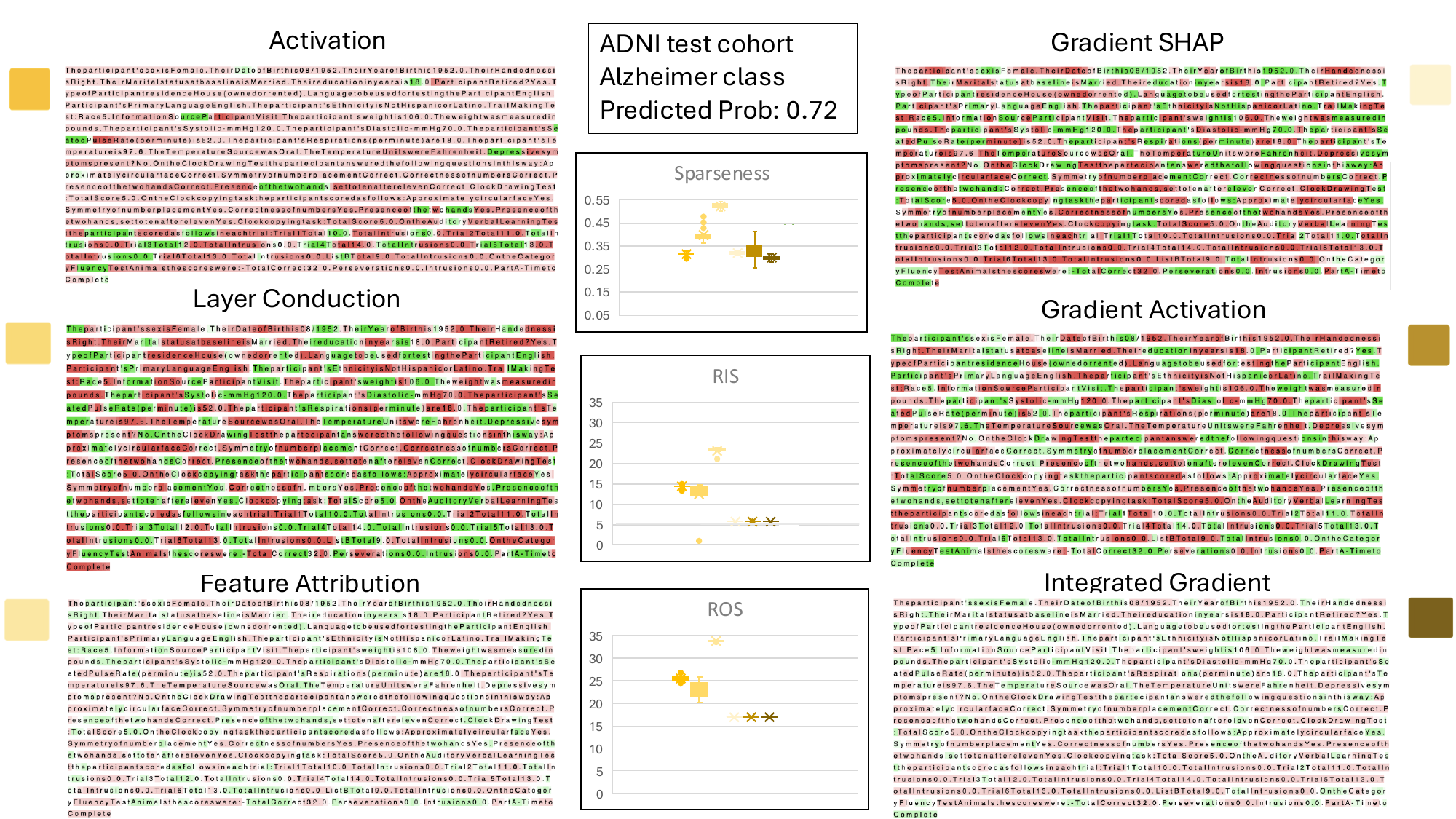}
\caption{Local attribution examples across different explanation methods without the SAE layer. The colour scale ranges from $-1$ (dark red; negative attribution), through 0 (white; neutral), to $+1$ (dark green; positive attribution). For each of the six panels, the small colour swatches at the top-left and top-right indicate the colour keys used for the three summary box-plot metrics—Sparseness, RIS, and ROS—for the corresponding attribution technique. The task is a binary classification (Alzheimer’s disease vs Control) on the ADNI cohort; the examples shown here are from the Alzheimer class.}
\label{arcx3}
\end{figure*}
 
 \begin{figure*}
\centering 
\includegraphics[width=0.9\textwidth]{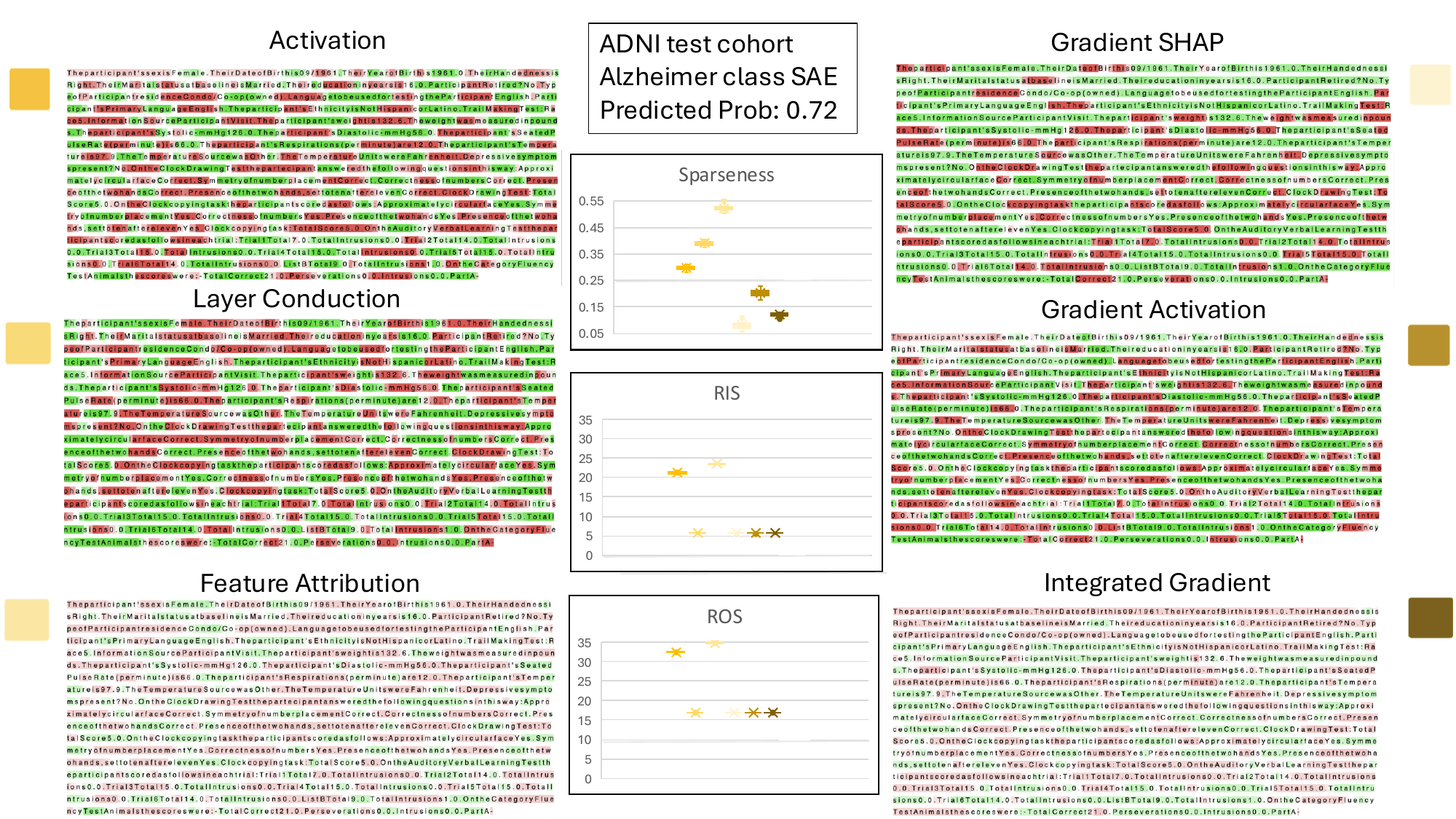}
\caption{Local attribution examples across different explanation methods with the SAE layer. The colour scale ranges from $-1$ (dark red; negative attribution), through 0 (white; neutral), to $+1$ (dark green; positive attribution). For each of the six panels, the small colour swatches at the top-left and top-right indicate the colour keys used for the three summary box-plot metrics—Sparseness, RIS, and ROS—for the corresponding attribution technique. The task is a binary classification (Alzheimer’s disease vs Control) on the ADNI cohort; the examples shown here are from the Alzheimer class.}
\label{arcx4}
\end{figure*}

 \begin{figure*}
\centering 
\includegraphics[width=0.9\textwidth]{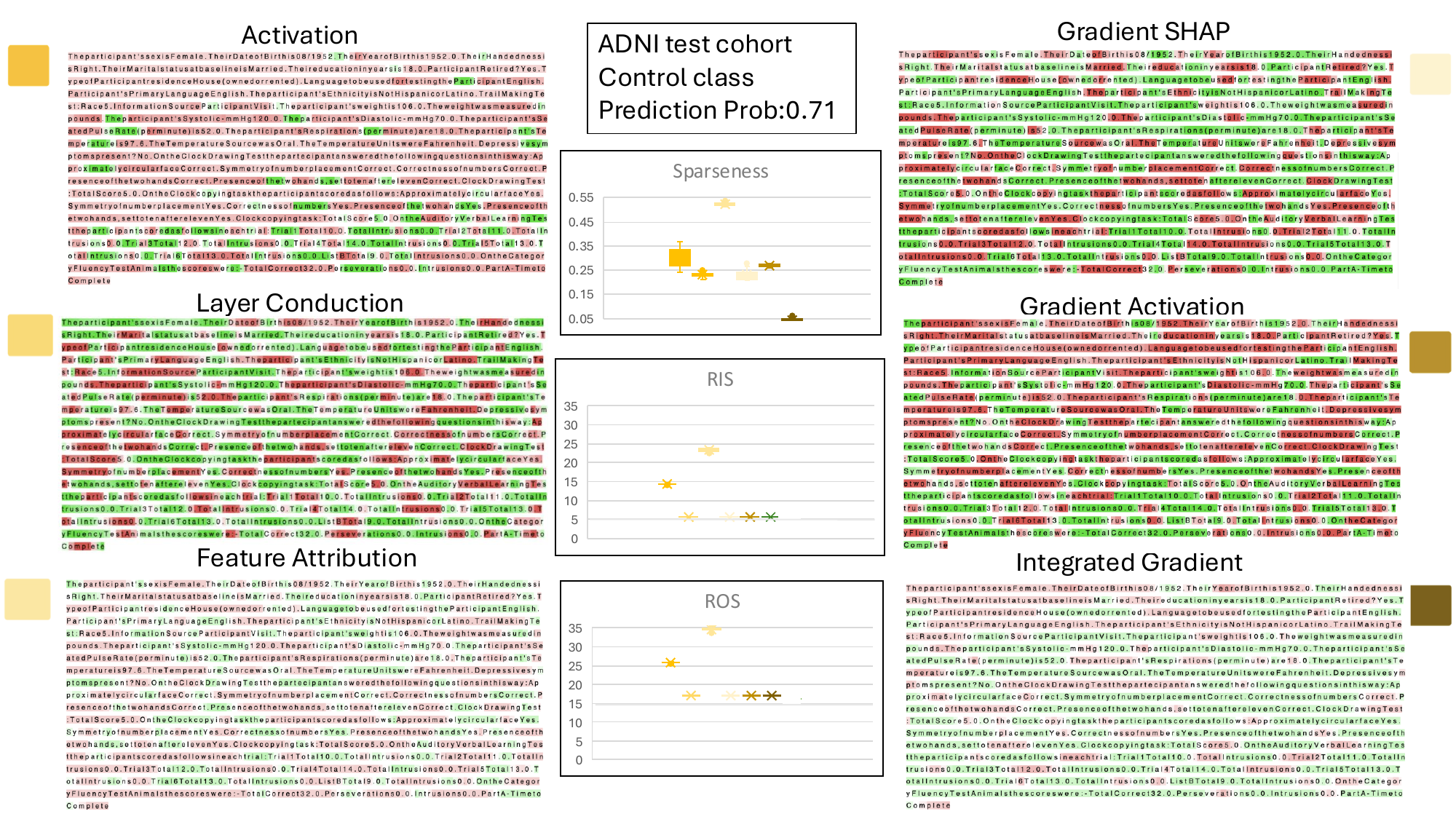}
\caption{Local attribution examples across different explanation methods without the SAE layer. The colour scale ranges from $-1$ (dark red; negative attribution), through 0 (white; neutral), to $+1$ (dark green; positive attribution). For each of the six panels, the small colour swatches at the top-left and top-right indicate the colour keys used for the three summary box-plot metrics—Sparseness, RIS, and ROS—for the corresponding attribution technique. The task is a three-class classification (LMCI, MCI disease vs Control) on the ADNI cohort; the examples shown here are from the Control class.}
\label{arcx5}
\end{figure*}

 \begin{figure*}
\centering 
\includegraphics[width=0.9\textwidth]{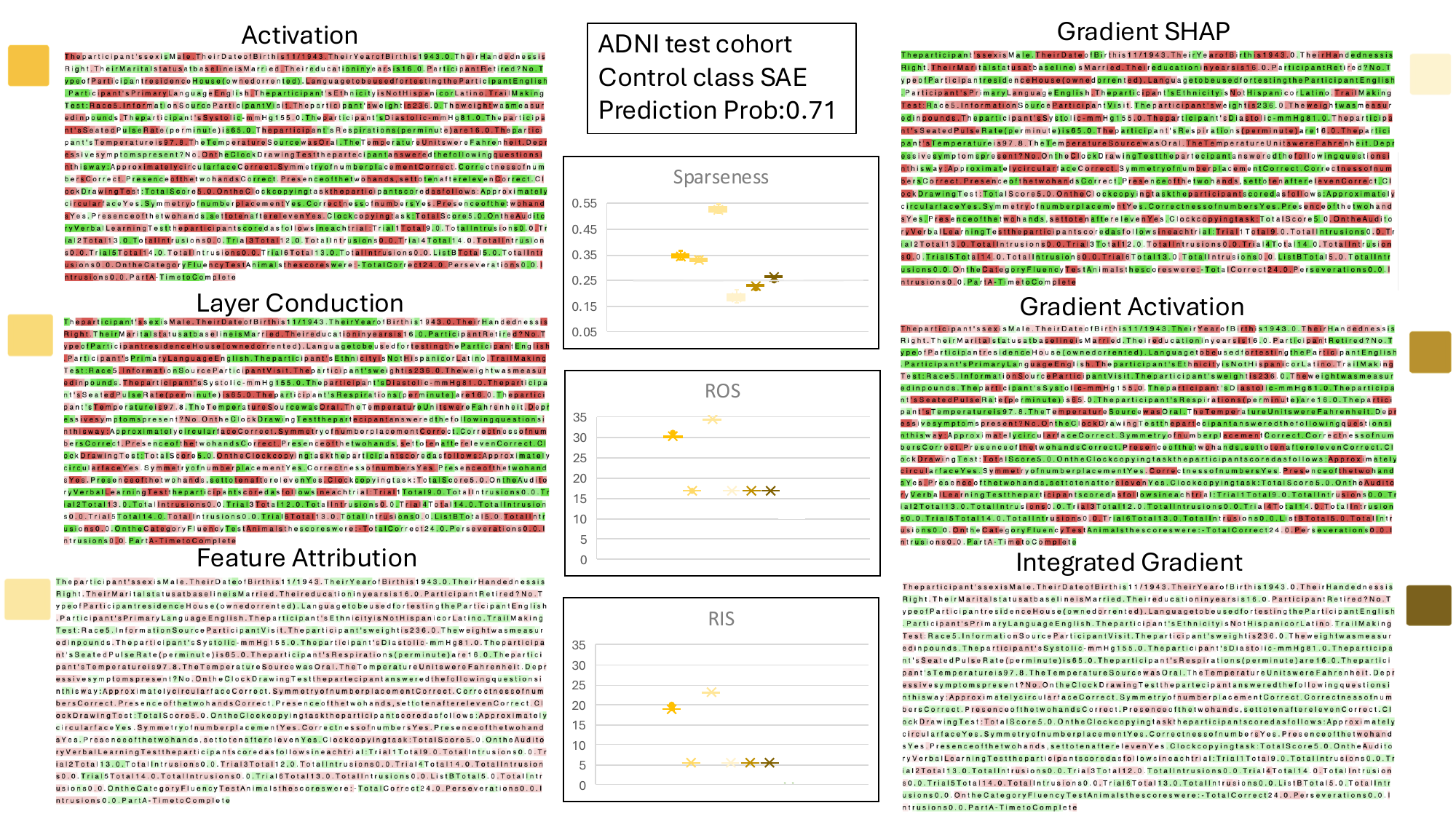}
\caption{Local attribution examples across different explanation methods with the SAE layer. The colour scale ranges from $-1$ (dark red; negative attribution), through 0 (white; neutral), to $+1$ (dark green; positive attribution). For each of the six panels, the small colour swatches at the top-left and top-right indicate the colour keys used for the three summary box-plot metrics—Sparseness, RIS, and ROS—for the corresponding attribution technique. The task is a three-class classification (LMCI, MCI disease vs Control) on the ADNI cohort; the examples shown here are from the Control class.}
\label{arcx6}
\end{figure*}

 \begin{figure*}
\centering 
\includegraphics[width=0.9\textwidth]{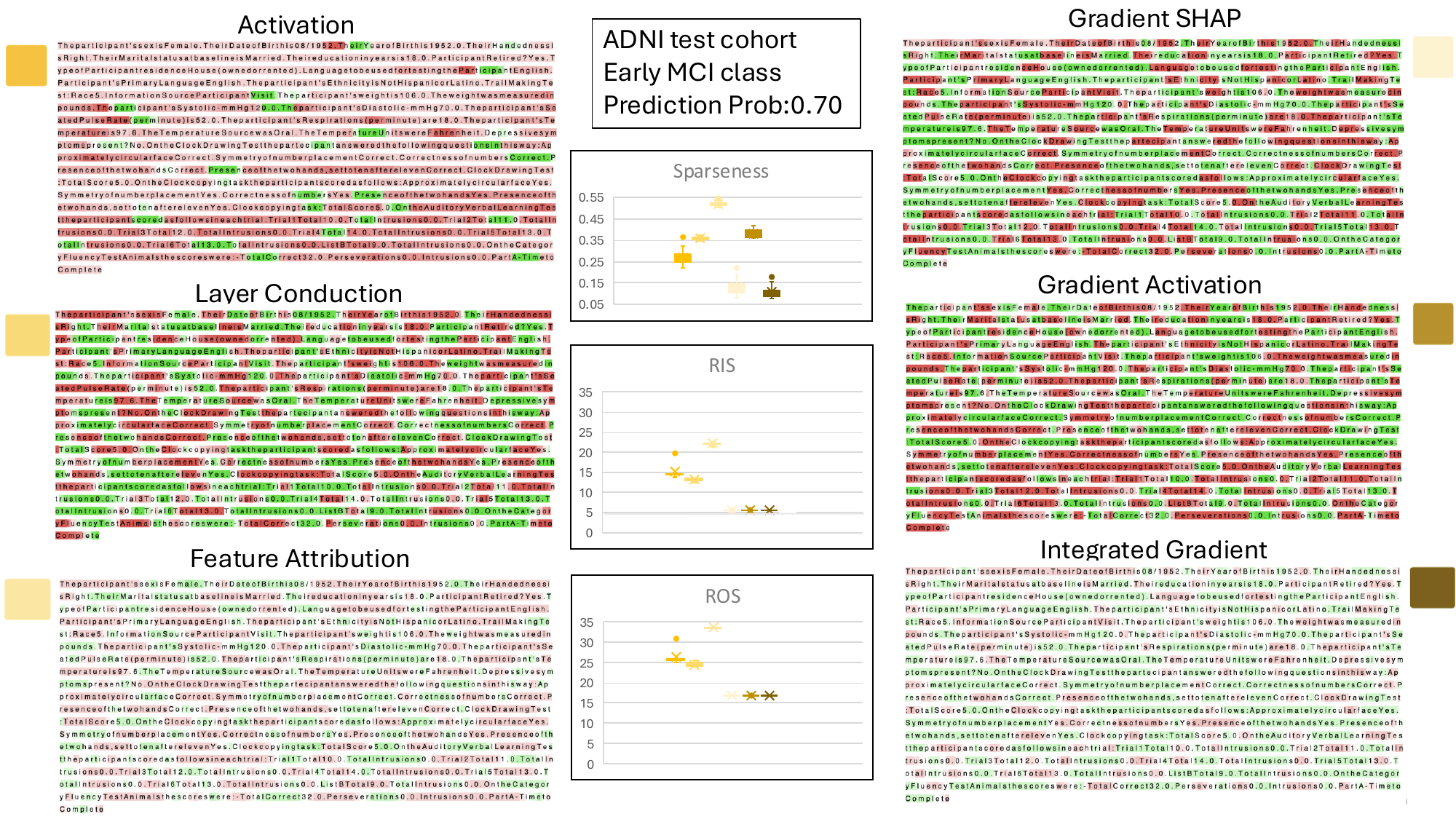}
\caption{Local attribution examples across different explanation methods without the SAE layer. The colour scale ranges from $-1$ (dark red; negative attribution), through 0 (white; neutral), to $+1$ (dark green; positive attribution). For each of the six panels, the small colour swatches at the top-left and top-right indicate the colour keys used for the three summary box-plot metrics—Sparseness, RIS, and ROS—for the corresponding attribution technique. The task is a three-class classification (LMCI, MCI disease vs Control) on the ADNI cohort; the examples shown here are from the LMCI class.}
\label{arcx7}
\end{figure*}

 \begin{figure*}
\centering 
\includegraphics[width=0.9\textwidth]{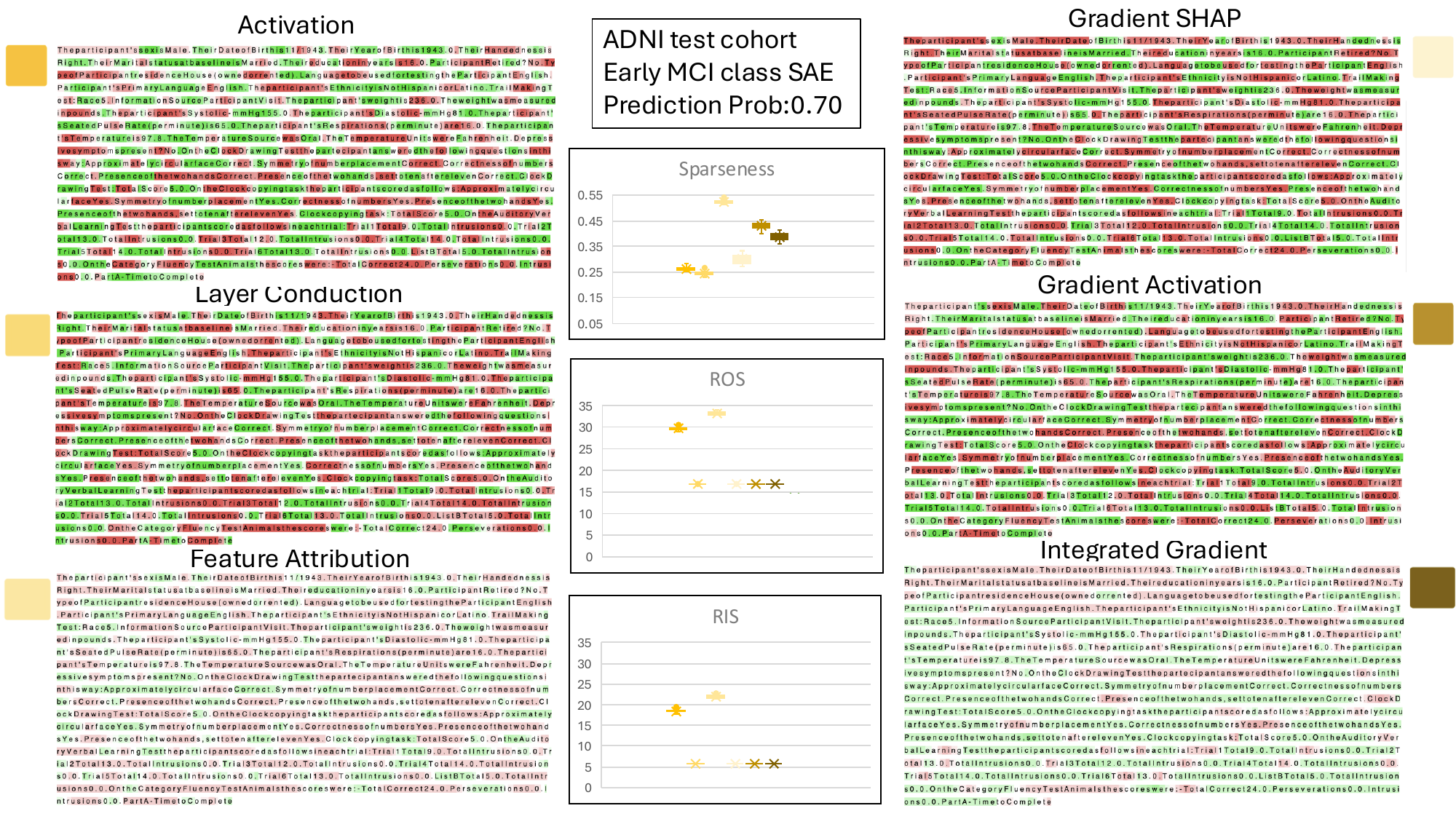}
\caption{Local attribution examples across different explanation methods with the SAE layer. The colour scale ranges from $-1$ (dark red; negative attribution), through 0 (white; neutral), to $+1$ (dark green; positive attribution). For each of the six panels, the small colour swatches at the top-left and top-right indicate the colour keys used for the three summary box-plot metrics—Sparseness, RIS, and ROS—for the corresponding attribution technique. The task is a three-class classification (LMCI, MCI disease vs Control) on the ADNI cohort; the examples shown here are from the LMCI class.}
\label{arcx8}
\end{figure*}

 \begin{figure*}
\centering 
\includegraphics[width=0.9\textwidth]{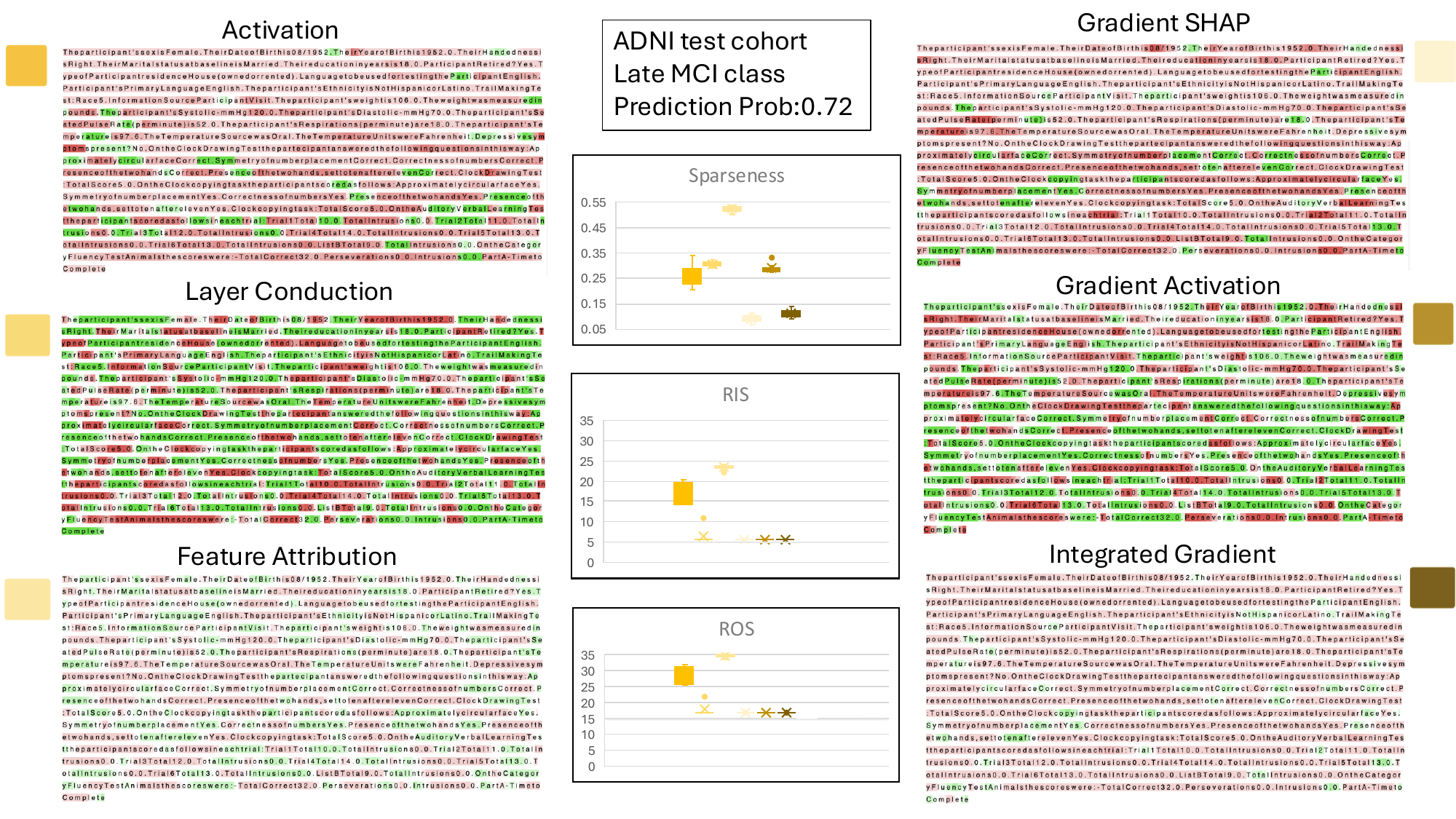}
\caption{Local attribution examples across different explanation methods without the SAE layer. The colour scale ranges from $-1$ (dark red; negative attribution), through 0 (white; neutral), to $+1$ (dark green; positive attribution). For each of the six panels, the small colour swatches at the top-left and top-right indicate the colour keys used for the three summary box-plot metrics—Sparseness, RIS, and ROS—for the corresponding attribution technique. The task is a three-class classification (LMCI, MCI disease vs Control) on the ADNI cohort; the examples shown here are from the MCI class.}
\label{arcx9}
\end{figure*}
 \begin{figure*}
\centering 
\includegraphics[width=0.9\textwidth]{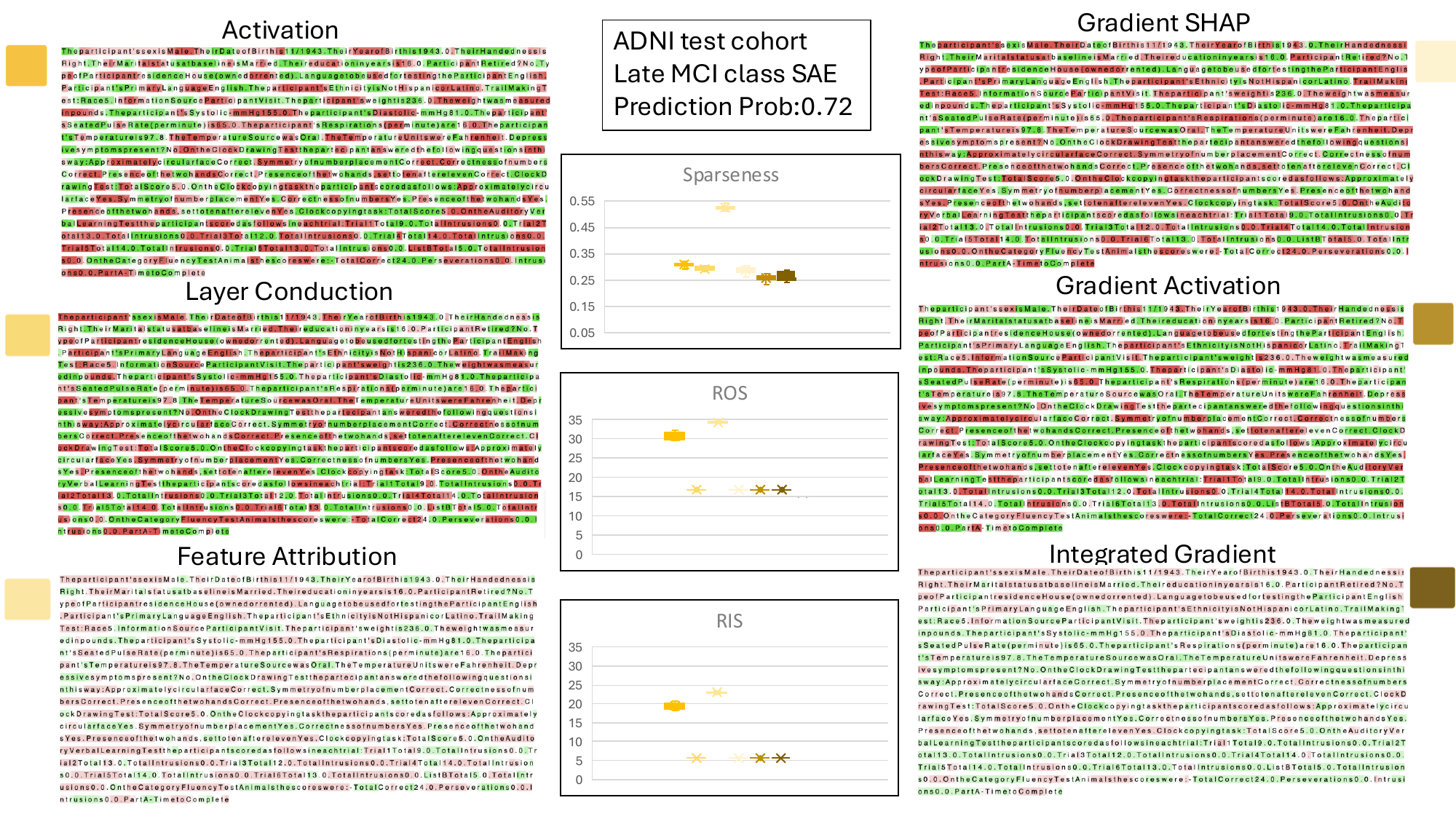}
\caption{Local attribution examples across different explanation methods with the SAE layer. The colour scale ranges from $-1$ (dark red; negative attribution), through 0 (white; neutral), to $+1$ (dark green; positive attribution). For each of the six panels, the small colour swatches at the top-left and top-right indicate the colour keys used for the three summary box-plot metrics—Sparseness, RIS, and ROS—for the corresponding attribution technique. The task is a three-class classification (LMCI, MCI disease vs Control) on the ADNI cohort; the examples shown here are from the MCI class.}
\label{arcx910}
\end{figure*}
\begin{figure*}
\centering 
\includegraphics[width=0.9\textwidth]{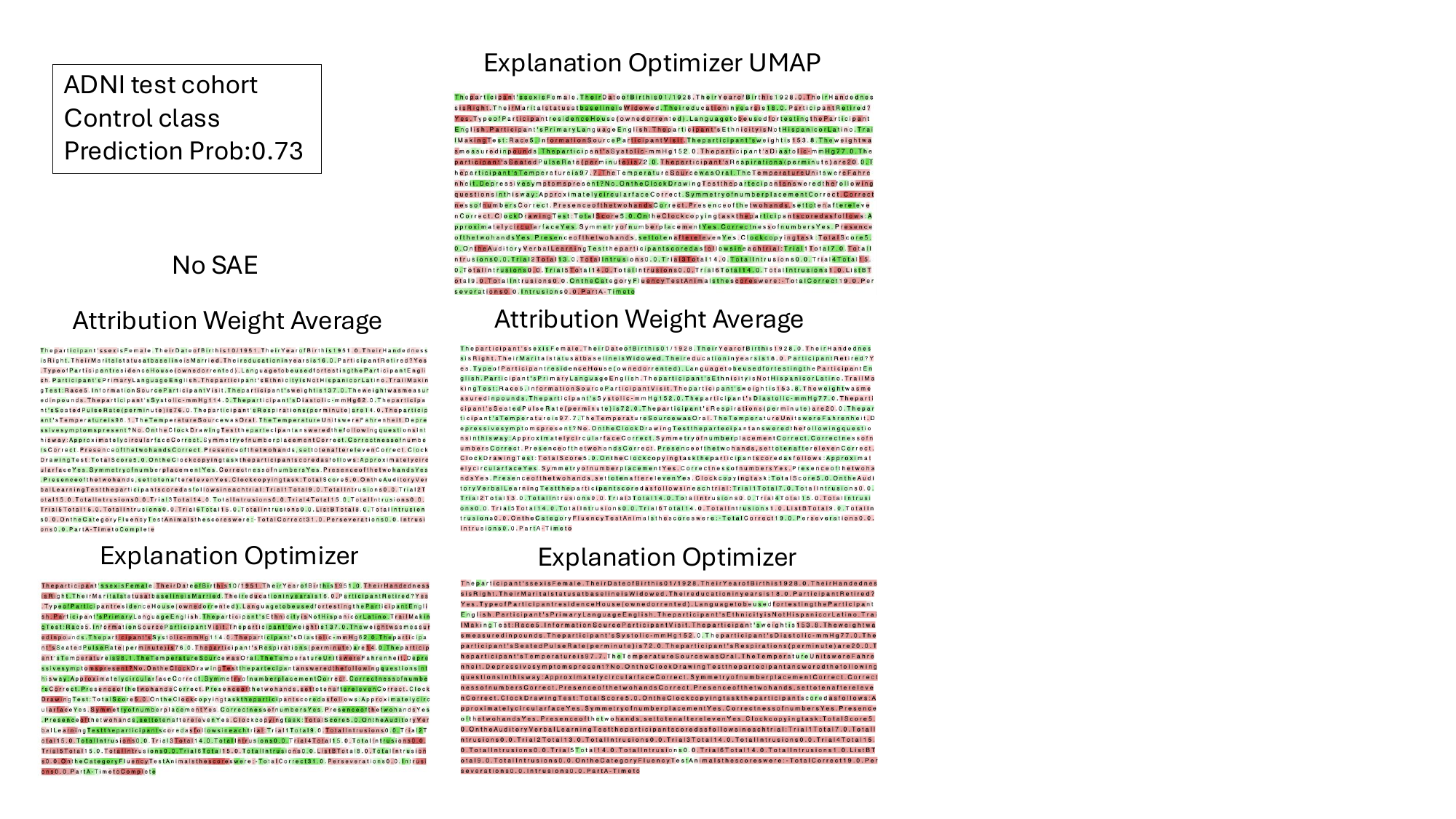}
\caption{Local attribution examples across different explanation methods with the SAE layer. The colour scale ranges from $-1$ (dark red; negative attribution), through 0 (white; neutral), to $+1$ (dark green; positive attribution). For each of the six panels, the small colour swatches at the top-left and top-right indicate the colour keys used for the three summary box-plot metrics—Sparseness, RIS, and ROS—for the corresponding attribution technique. The task is a binary classification (Alzheimer’s disease vs Control) on the ADNI cohort; the examples shown here are from the Control class.}
\label{arcx10}
\end{figure*}

Figures~\ref{arcx1}–\ref{arcx910} present qualitative local attribution examples for the binary (Control and Alzheimer) and three-class classification task (Control, LMCI, MCI) of ADNI cohorts across six explanation methods, each evaluated without (Figures~\ref{arcx1}, \ref{arcx3}, \ref{arcx5}, \ref{arcx7}, \ref{arcx9}) and with (Figures~\ref{arcx2}, \ref{arcx4}, \ref{arcx6}, \ref{arcx8}, \ref{arcx910}) the Sparse Autoencoder (SAE) layer. Each cell shows token-level attributions using colour-coded highlights (green = positive relevance; red = negative relevance). In general, higher Sparseness is associated with a more balanced distribution of positive and negative highlights (i.e., less diffuse maps), particularly for Layer Conduction, Feature Ablation, Gradient SHAP, and Integrated Gradient.
For the Control class (see Figures~\ref{arcx1} and~\ref{arcx2}), the qualitative highlighting patterns are broadly consistent across the six attribution techniques, Activation, Layer Conduction, Feature Ablation, Gradient SHAP, Gradient Activation, and Integrated Gradient—with no marked visual discrepancies. Notably, Feature Ablation, despite exhibiting the strongest Sparseness in the box plots, shows poorer stability (higher variability in inputs/outputs; elevated RIS/ROS), and the addition of the SAE layer tends to worsen this by exposing a larger set of features due to the decoder “decompression” effect; a similar trend is observed for Activation.
For the Alzheimer’s class (Figures~\ref{arcx3} and~\ref{arcx4}), Layer Conduction demonstrates a reduction in Sparseness with the SAE but a gain in stability (decreased RIS/ROS). Comparable improvements in stability with SAE are also observed for Gradient Activation, Integrated Gradient, and Gradient SHAP. In contrast, Activation and Feature Ablation perform worst under SAE, again exposing many more features and yielding less stable explanations.
Across the remaining examples (Figures~\ref{arcx5}–\ref{arcx10}), similar patterns hold: instances with low Sparseness and high RIS/ROS tend to produce saturated red/green explanations (strongly negative or positive attributions), whereas higher Sparseness with lower RIS/ROS yields more compact and stable saliency patterns.

 \begin{figure*}
\centering 
\includegraphics[width=0.9\textwidth]{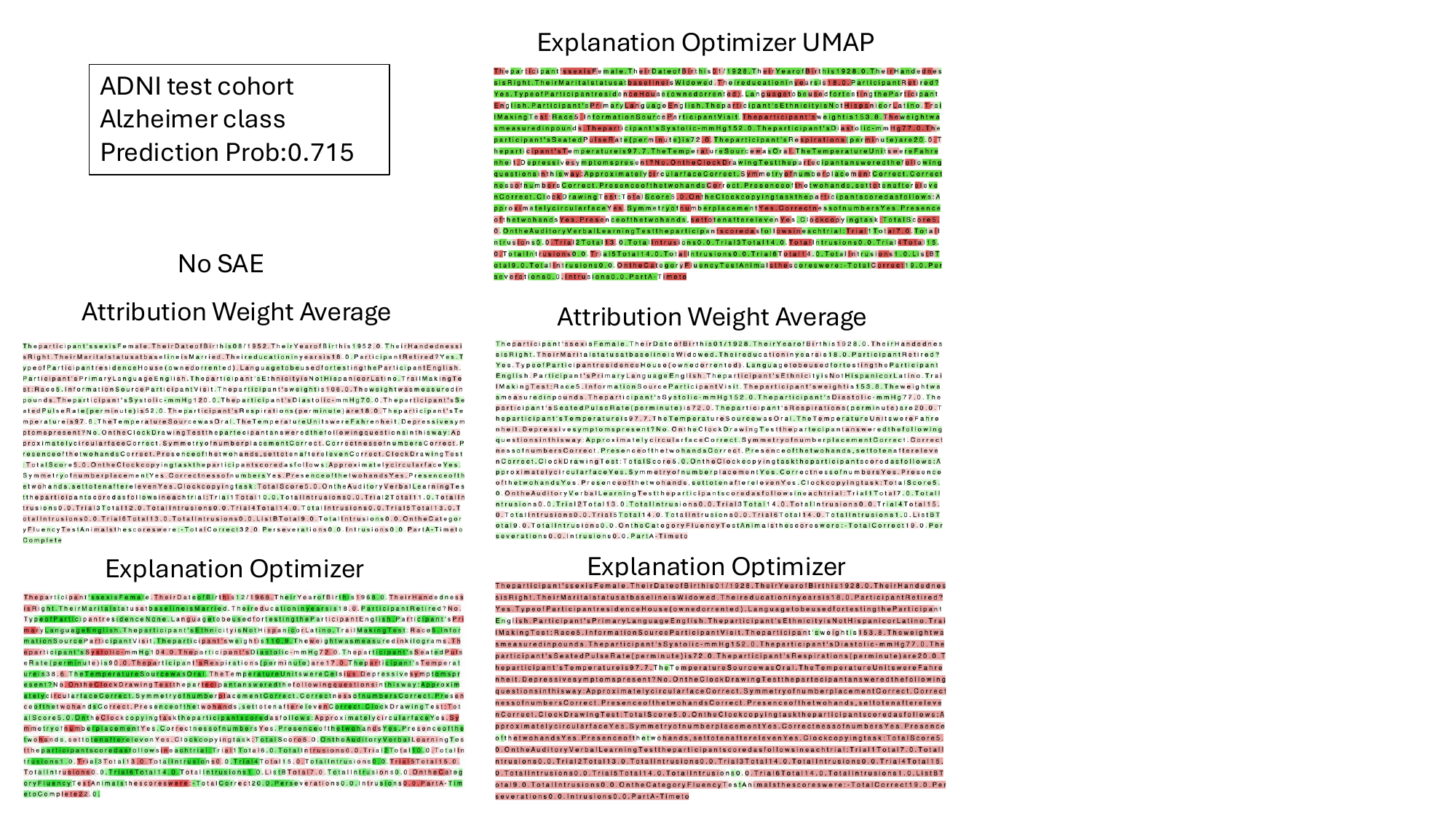}
\caption{Local attribution examples across different explanation methods without the SAE layer. The colour scale ranges from $-1$ (dark red; negative attribution), through 0 (white; neutral), to $+1$ (dark green; positive attribution). For each of the six panels, the small colour swatches at the top-left and top-right indicate the colour keys used for the three summary box-plot metrics—Sparseness, RIS, and ROS—for the corresponding attribution technique. The task is a binary classification (Alzheimer’s disease vs Control) on the ADNI cohort; the examples shown here are from the Alzheimer class.}
\label{arcx11}
\end{figure*}

 \begin{figure*}
\centering 
\includegraphics[width=0.9\textwidth]{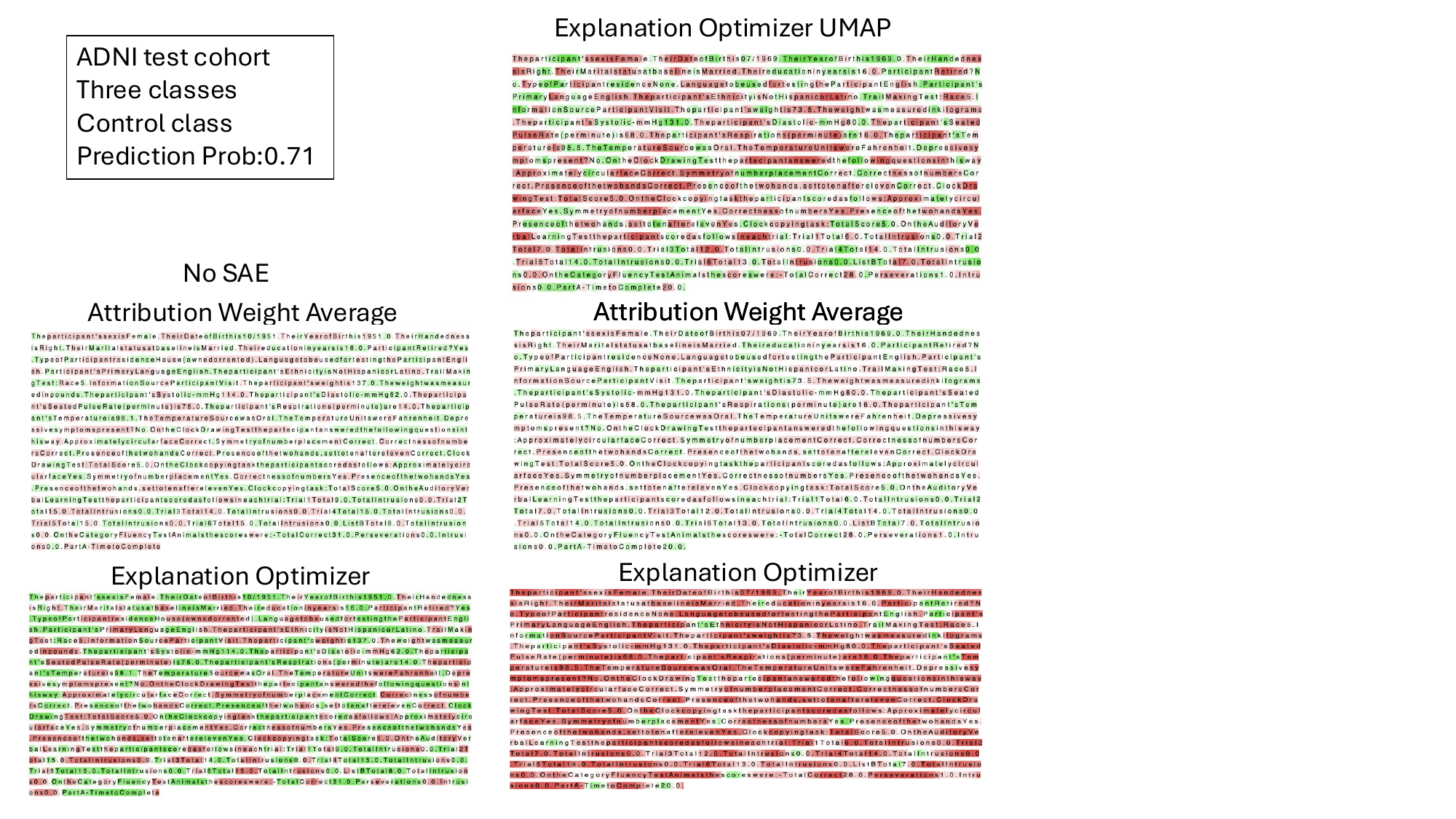}
\caption{Local attribution examples across different explanation methods with the SAE layer. The colour scale ranges from $-1$ (dark red; negative attribution), through 0 (white; neutral), to $+1$ (dark green; positive attribution). For each of the six panels, the small colour swatches at the top-left and top-right indicate the colour keys used for the three summary box-plot metrics—Sparseness, RIS, and ROS—for the corresponding attribution technique. The task is a three-class classification (LMCI, MCI disease vs Control) on the ADNI cohort; the examples shown here are from the Control class.}
\label{arcx12}
\end{figure*}

 \begin{figure*}
\centering 
\includegraphics[width=0.9\textwidth]{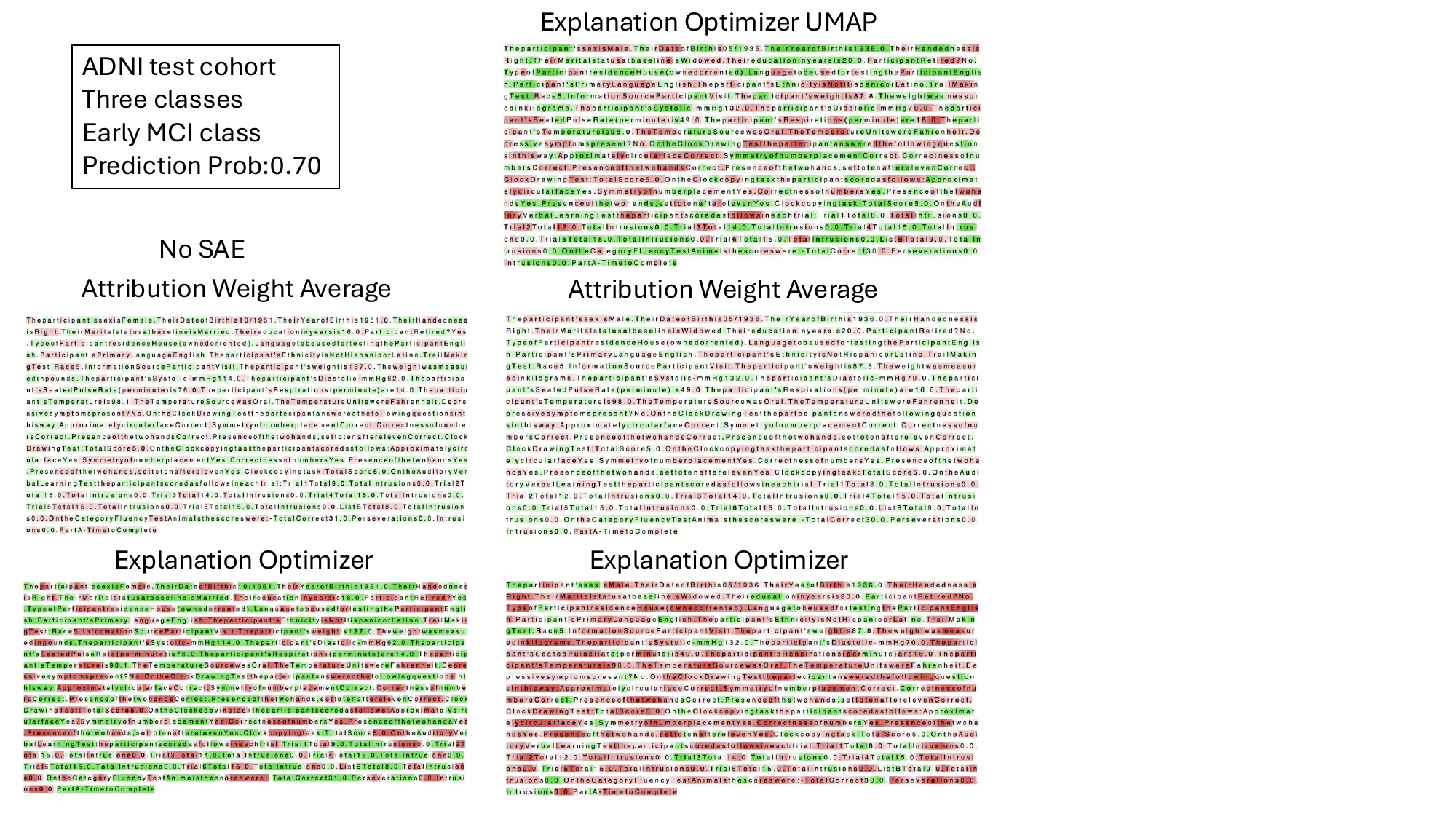}
\caption{Local attribution examples across different explanation methods without the SAE layer. The colour scale ranges from $-1$ (dark red; negative attribution), through 0 (white; neutral), to $+1$ (dark green; positive attribution). For each of the six panels, the small colour swatches at the top-left and top-right indicate the colour keys used for the three summary box-plot metrics—Sparseness, RIS, and ROS—for the corresponding attribution technique. The task is a three-class classification (LMCI, MCI disease vs Control) on the ADNI cohort; the examples shown here are from the LMCI class.}
\label{arcx13}
\end{figure*}

\begin{figure*}
\centering 
\includegraphics[width=0.9\textwidth]{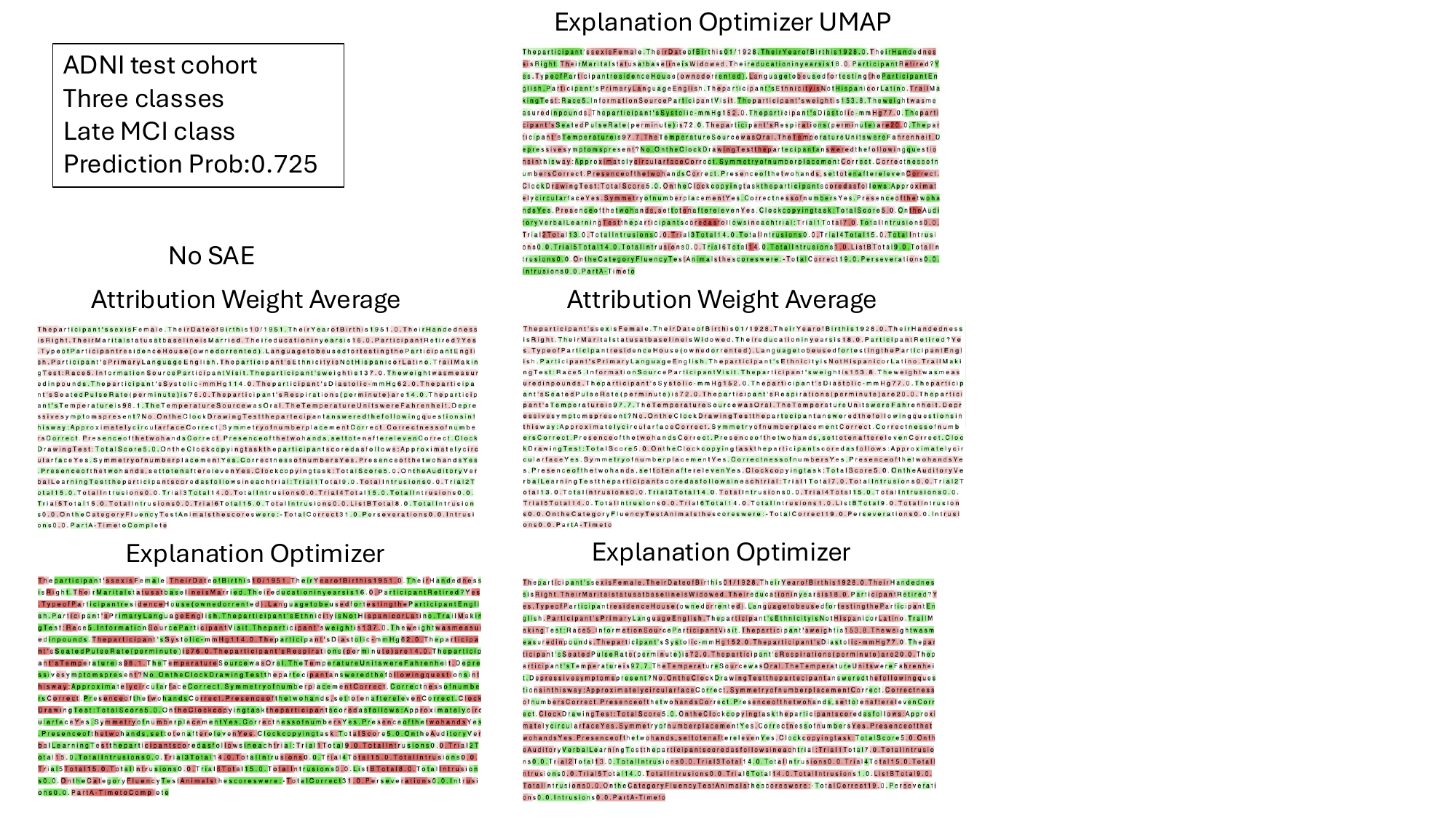}
\caption{Local attribution examples across different explanation methods without the SAE layer. The colour scale ranges from $-1$ (dark red; negative attribution), through 0 (white; neutral), to $+1$ (dark green; positive attribution). For each of the six panels, the small colour swatches at the top-left and top-right indicate the colour keys used for the three summary box-plot metrics—Sparseness, RIS, and ROS—for the corresponding attribution technique. The task is a three-class classification (LMCI, MCI disease vs Control) on the ADNI cohort; the examples shown here are from the MCI class.}
\label{arcx14}
\end{figure*}

Figures~\ref{arcx10}–\ref{arcx14} present qualitative local attribution examples, analogous to Figures~\ref{arcx1}–\ref{arcx10}, for the no-SAE analyses of (i) the attributional weighted average (computed from the six base methods), (ii) the Transformer Explanation Optimizer (TEO), and (iii) TEO with a linear UMAP constraint (TEO-UMAP). As shown in the previous subsection, with the SAE layer TEO achieves the best stability—i.e., the lowest RIS and ROS—but at the cost of a marked reduction in Sparseness; this reduction is clearly visible in the binary task (Figures~\ref{arcx10}–\ref{arcx11}). Introducing the UMAP constraint yields a more balanced trade-off, producing explanations that are more compact and clinically interpretable; the same behaviour is observed across all classes in the three-class setting (Figures~\ref{arcx12}–\ref{arcx14}). By contrast, the weighted-average approach—a linear combination of the six attribution techniques—does not yield superior explanations, consistent with \cite{mamalakis}.

\subsection{UMAP and Cohort-Level Explanation and Patterns.}
\begin{figure*}
\centering 
\includegraphics[width=0.9\textwidth]{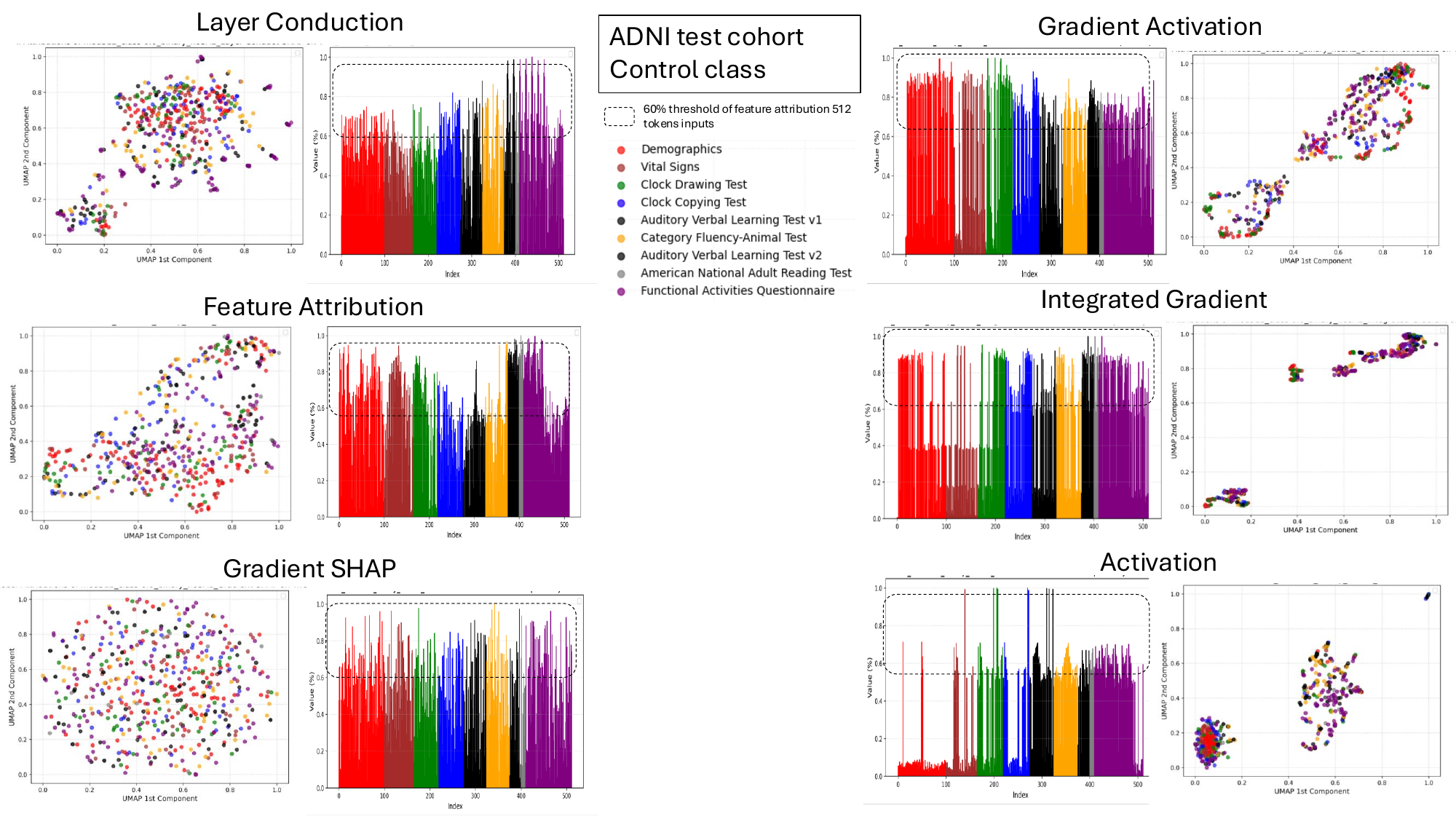}
\caption{Global (cohort-level) feature attribution across explanation methods without the SAE layer. The 2D panel shows a UMAP embedding (UMAP-1 vs UMAP-2) computed on the ADNI test set; the 1D panel shows attribution scores along PCA first component. All plotted values are normalised to [0, 1] and represent positive contributions only. Colours (red→purple) denote the nine ADNI subgroups (see §B3). Square boxes mark the 0.6–1.0 interval, highlighting the most significant tokens in the 1D views. The task is binary classification (Alzheimer’s vs Control) on the ADNI cohort; the examples shown here are from the Control class.The task is a three-class classification (LMCI, MCI disease vs Control) on the ADNI cohort; the examples shown here are from the MCI class.}
\label{ar1}
\end{figure*}
\begin{figure*}
\centering 
\includegraphics[width=0.9\textwidth]{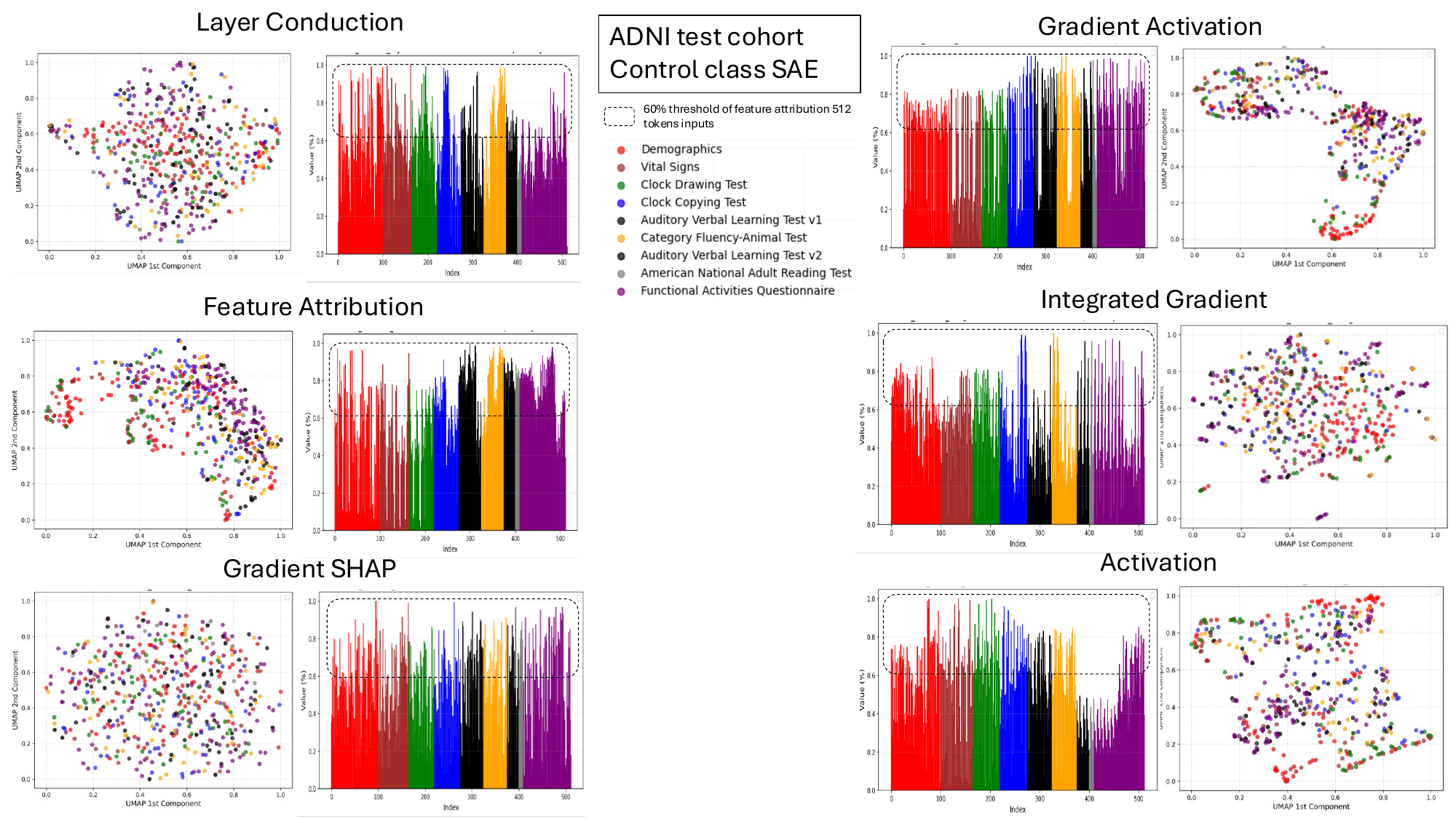}
\caption{Global (cohort-level) feature attribution across explanation methods with the SAE layer. The 2D panel shows a UMAP embedding (UMAP-1 vs UMAP-2) computed on the ADNI test set; the 1D panel shows attribution scores along PCA first component. All plotted values are normalised to [0, 1] and represent positive contributions only. Colours (red→purple) denote the nine ADNI subgroups (see §B3). Square boxes mark the 0.6–1.0 interval, highlighting the most significant tokens in the 1D views. The task is binary classification (Alzheimer’s vs Control) on the ADNI cohort; the examples shown here are from the Control class.}
\label{ar2}
\end{figure*}

\begin{figure*}
\centering 
\includegraphics[width=0.9\textwidth]{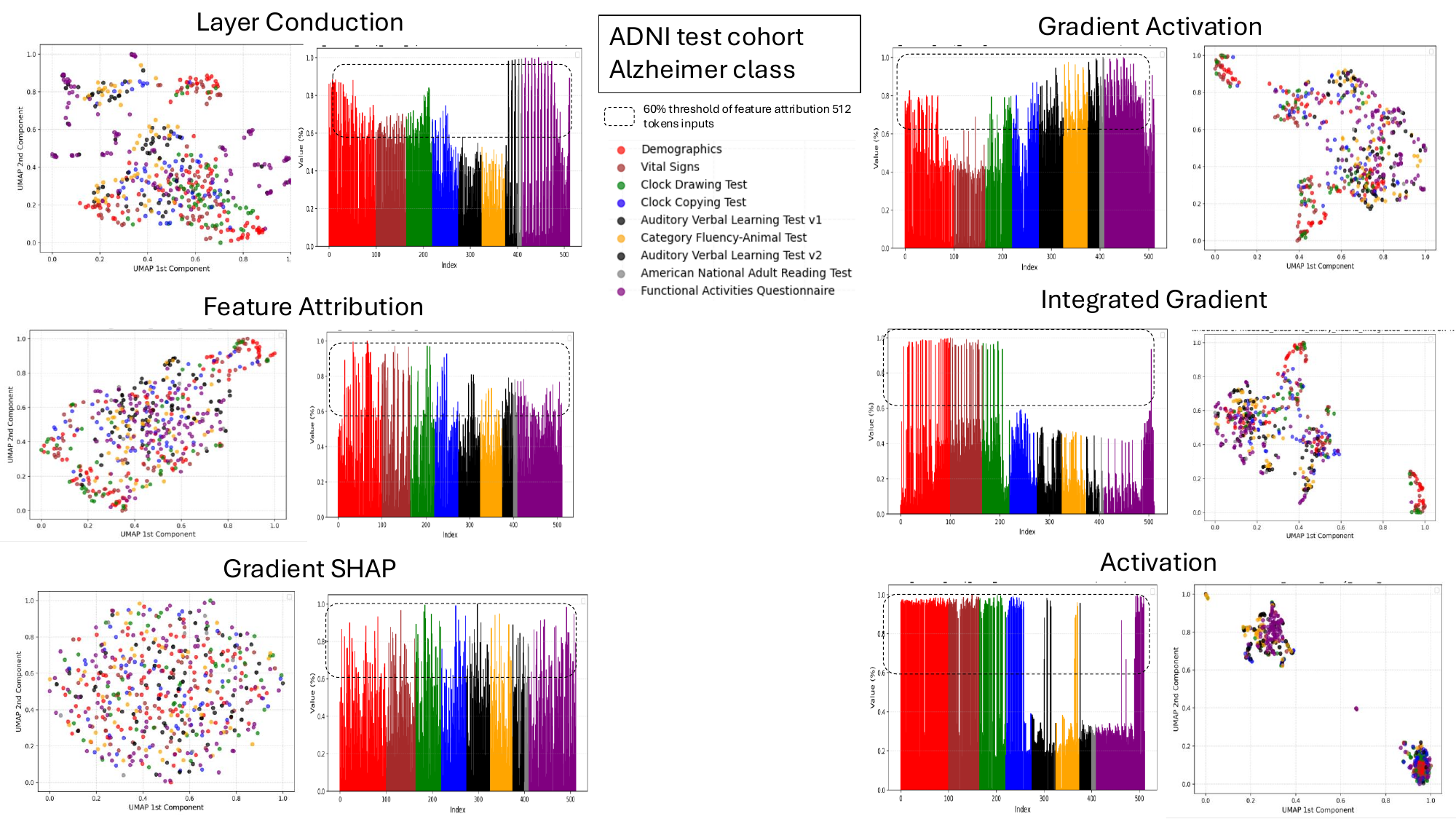}
\caption{Global (cohort-level) feature attribution across explanation methods without the SAE layer. The 2D panel shows a UMAP embedding (UMAP-1 vs UMAP-2) computed on the ADNI test set;the 1D panel shows attribution scores along PCA first component. All plotted values are normalised to [0, 1] and represent positive contributions only. Colours (red→purple) denote the nine ADNI subgroups (see §B3). Square boxes mark the 0.6–1.0 interval, highlighting the most significant tokens in the 1D views. The task is binary classification (Alzheimer’s vs Control) on the ADNI cohort; the examples shown here are from the Alzheimer class.}
\label{ar3}
\end{figure*}
\begin{figure*}
\centering 
\includegraphics[width=0.9\textwidth]{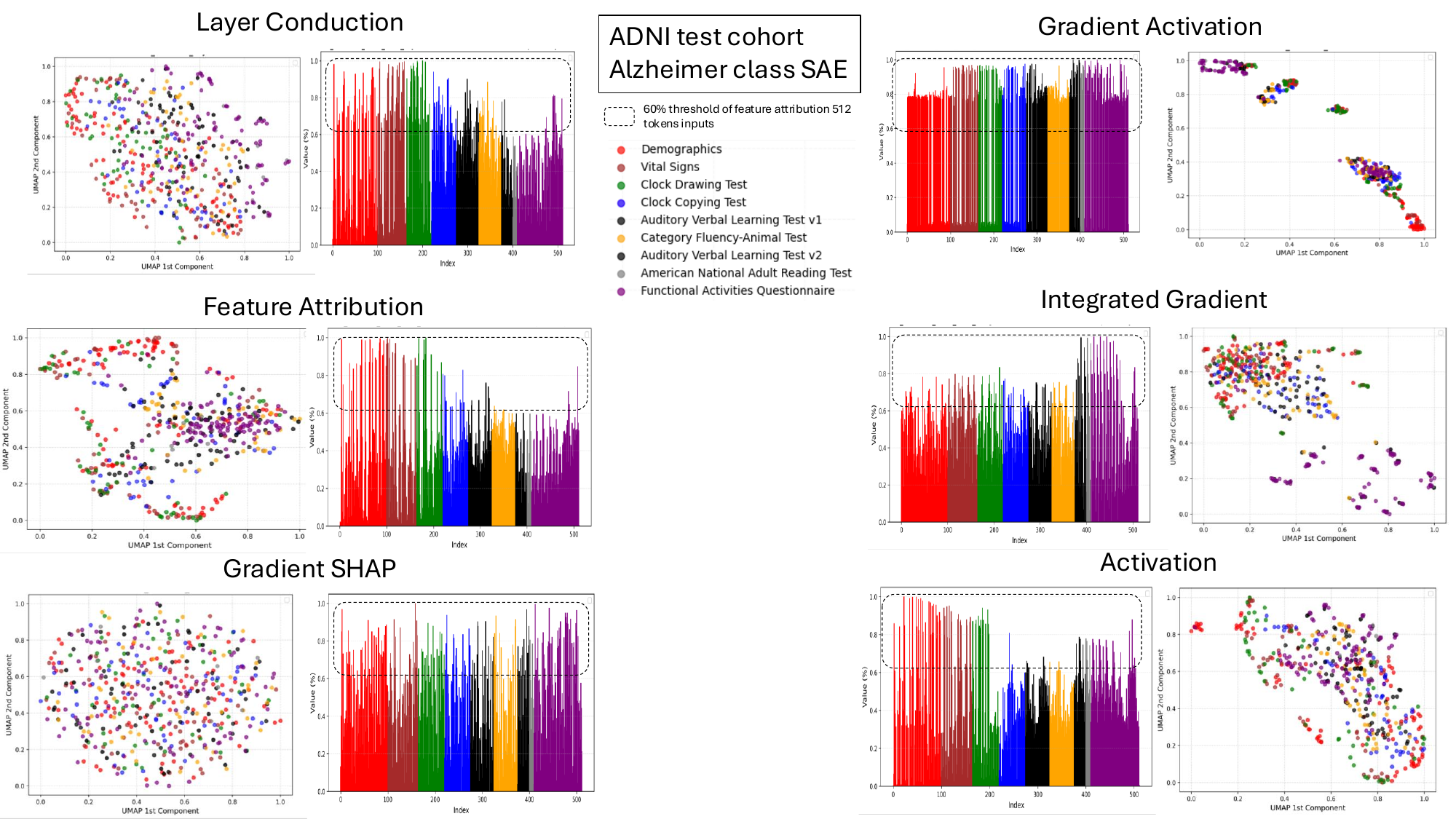}
\caption{Global (cohort-level) feature attribution across explanation methods with the SAE layer. The 2D panel shows a UMAP embedding (UMAP-1 vs UMAP-2) computed on the ADNI test setthe 1D panel shows attribution scores along PCA first component. All plotted values are normalised to [0, 1] and represent positive contributions only. Colours (red→purple) denote the nine ADNI subgroups (see §B3). Square boxes mark the 0.6–1.0 interval, highlighting the most significant tokens in the 1D views. The task is binary classification (Alzheimer’s vs Control) on the ADNI cohort; the examples shown here are from the Alzheimer class.}
\label{ar4}
\end{figure*}

\begin{figure*}
\centering 
\includegraphics[width=0.9\textwidth]{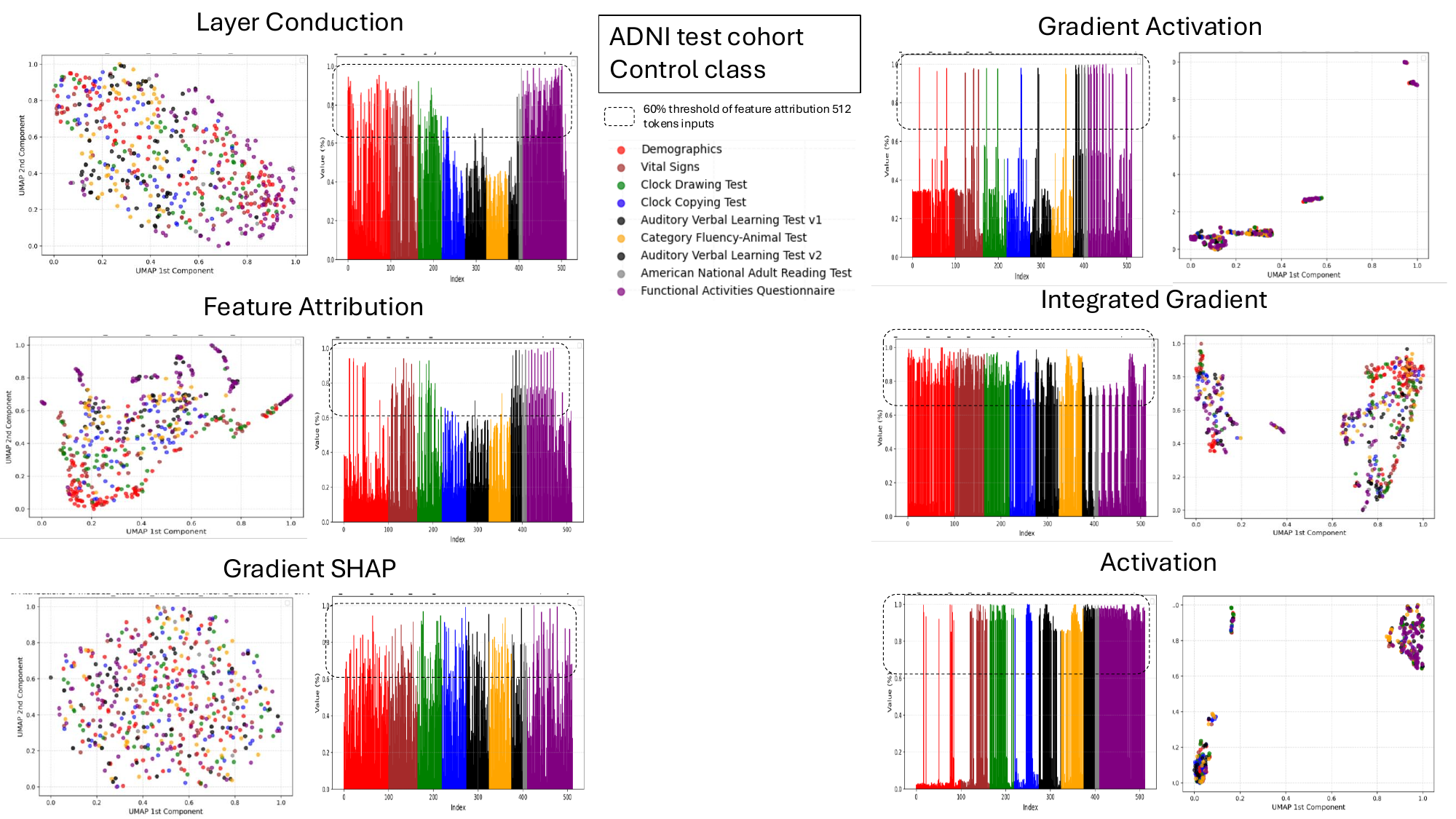}
\caption{Global (cohort-level) feature attribution across explanation methods without the SAE layer. The 2D panel shows a UMAP embedding (UMAP-1 vs UMAP-2) computed on the ADNI test setthe 1D panel shows attribution scores along PCA first component. All plotted values are normalised to [0, 1] and represent positive contributions only. Colours (red→purple) denote the nine ADNI subgroups (see §B3). Square boxes mark the 0.6–1.0 interval, highlighting the most significant tokens in the 1D views. The task is a three-class classification (LMCI, MCI disease vs Control) on the ADNI cohort; the examples shown here are from the Control class.}
\label{ar5}
\end{figure*}
\begin{figure*}
\centering 
\includegraphics[width=0.9\textwidth]{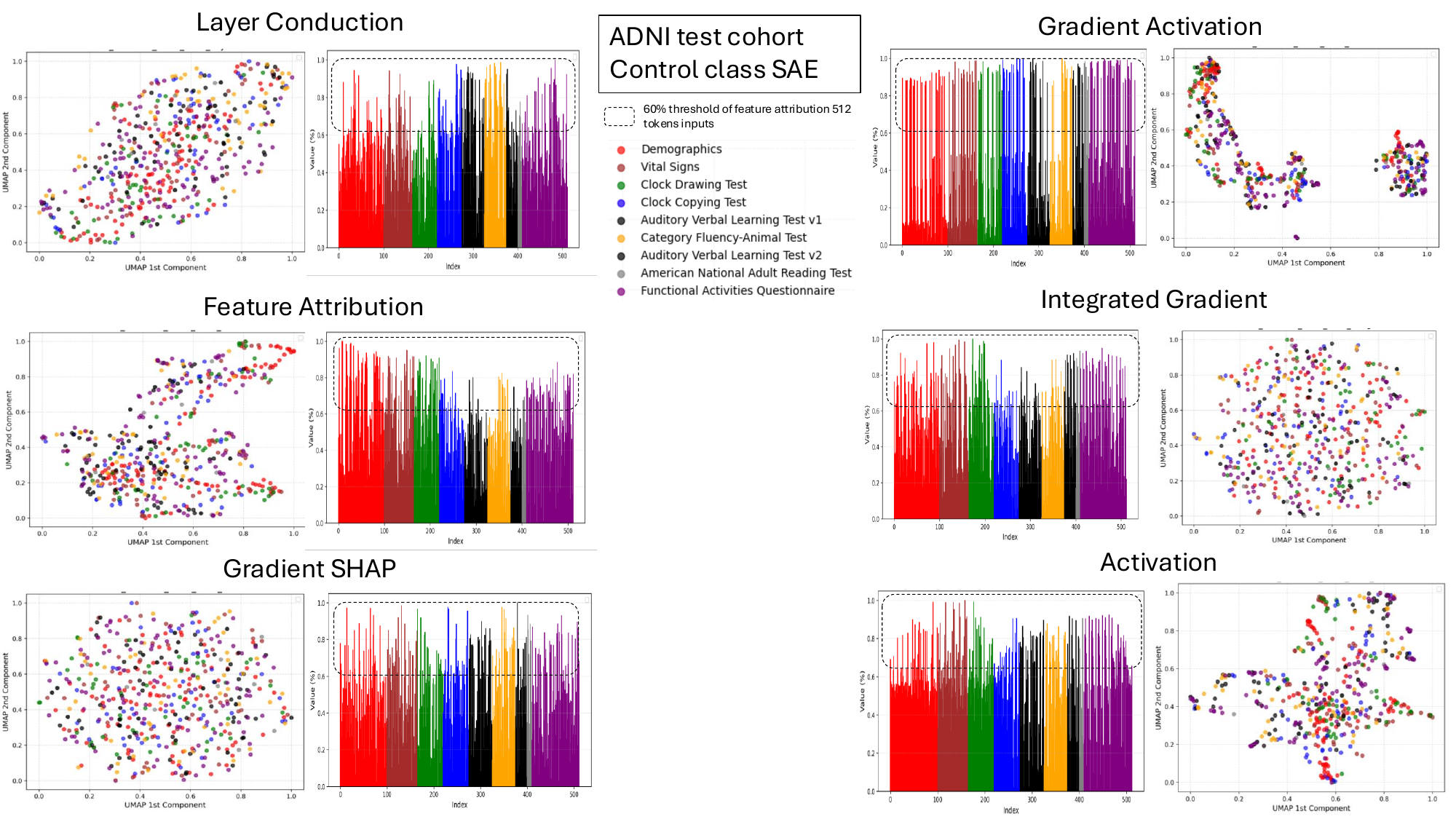}
\caption{Global (cohort-level) feature attribution across explanation methods with the SAE layer. The 2D panel shows a UMAP embedding (UMAP-1 vs UMAP-2) computed on the ADNI test setthe 1D panel shows attribution scores along PCA first component. All plotted values are normalised to [0, 1] and represent positive contributions only. Colours (red→purple) denote the nine ADNI subgroups (see §B3). Square boxes mark the 0.6–1.0 interval, highlighting the most significant tokens in the 1D views. The task is a three-class classification (LMCI, MCI disease vs Control) on the ADNI cohort; the examples shown here are from the Control class.}
\label{ar6}
\end{figure*}

\begin{figure*}
\centering 
\includegraphics[width=0.9\textwidth]{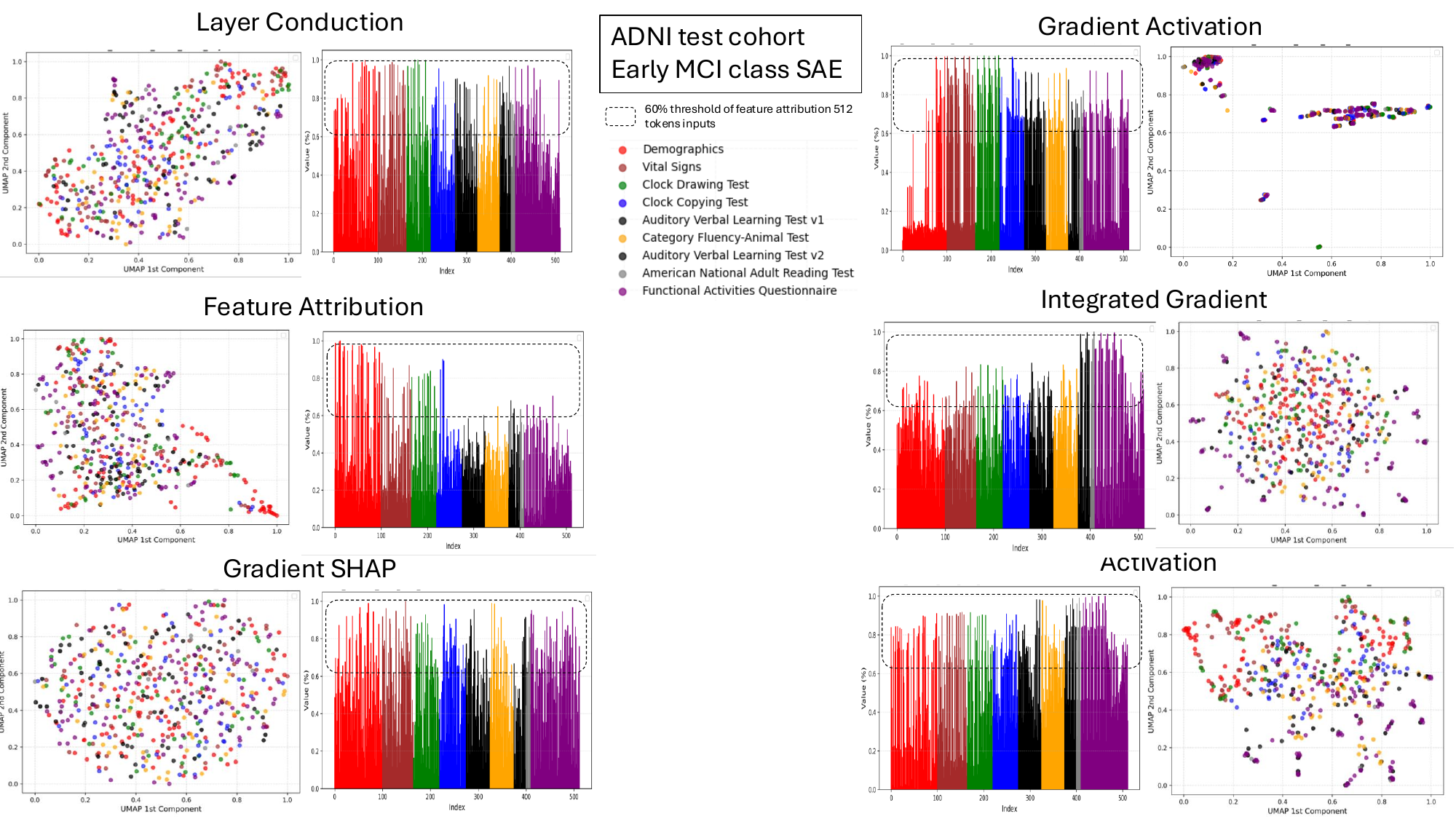}
\caption{Global (cohort-level) feature attribution across explanation methods with the SAE layer. The 2D panel shows a UMAP embedding (UMAP-1 vs UMAP-2) computed on the ADNI test set; the 1D panel shows attribution scores along PCA first component. All plotted values are normalised to [0, 1] and represent positive contributions only. Colours (red→purple) denote the nine ADNI subgroups (see §B3). Square boxes mark the 0.6–1.0 interval, highlighting the most significant tokens in the 1D views. The task is a three-class classification (LMCI, MCI disease vs Control) on the ADNI cohort; the examples shown here are from the LMCI class.}
\label{ar7}
\end{figure*}
\begin{figure*}
\centering 
\includegraphics[width=0.9\textwidth]{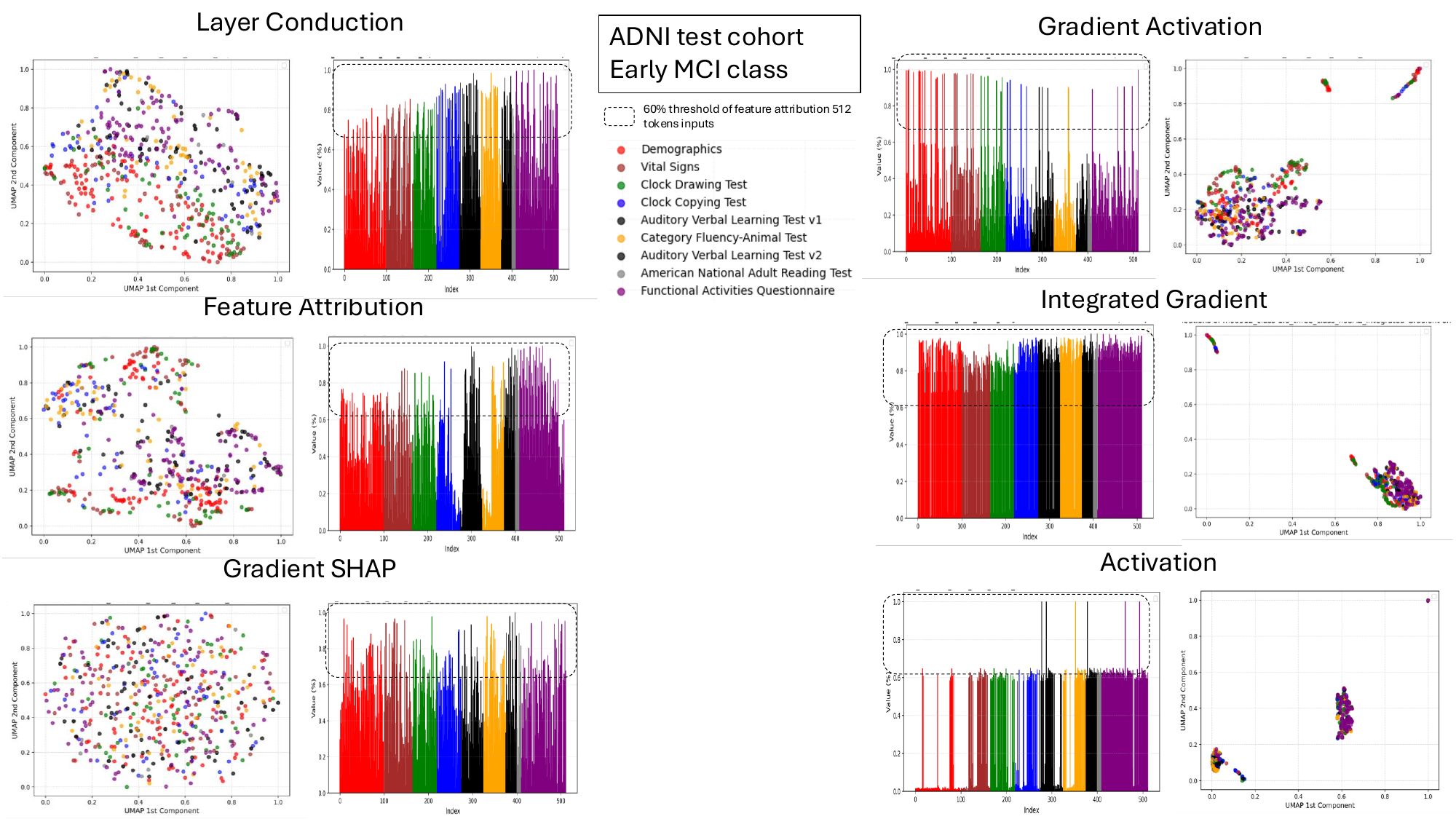}
\caption{Global (cohort-level) feature attribution across explanation methods without the SAE layer. The 2D panel shows a UMAP embedding (UMAP-1 vs UMAP-2) computed on the ADNI test set; the 1D panel shows attribution scores along PCA first component. All plotted values are normalised to [0, 1] and represent positive contributions only. Colours (red→purple) denote the nine ADNI subgroups (see §B3). Square boxes mark the 0.6–1.0 interval, highlighting the most significant tokens in the 1D views. The task is a three-class classification (LMCI, MCI disease vs Control) on the ADNI cohort; the examples shown here are from the LMCI class.}
\label{ar8}
\end{figure*}

\begin{figure*}
\centering 
\includegraphics[width=0.9\textwidth]{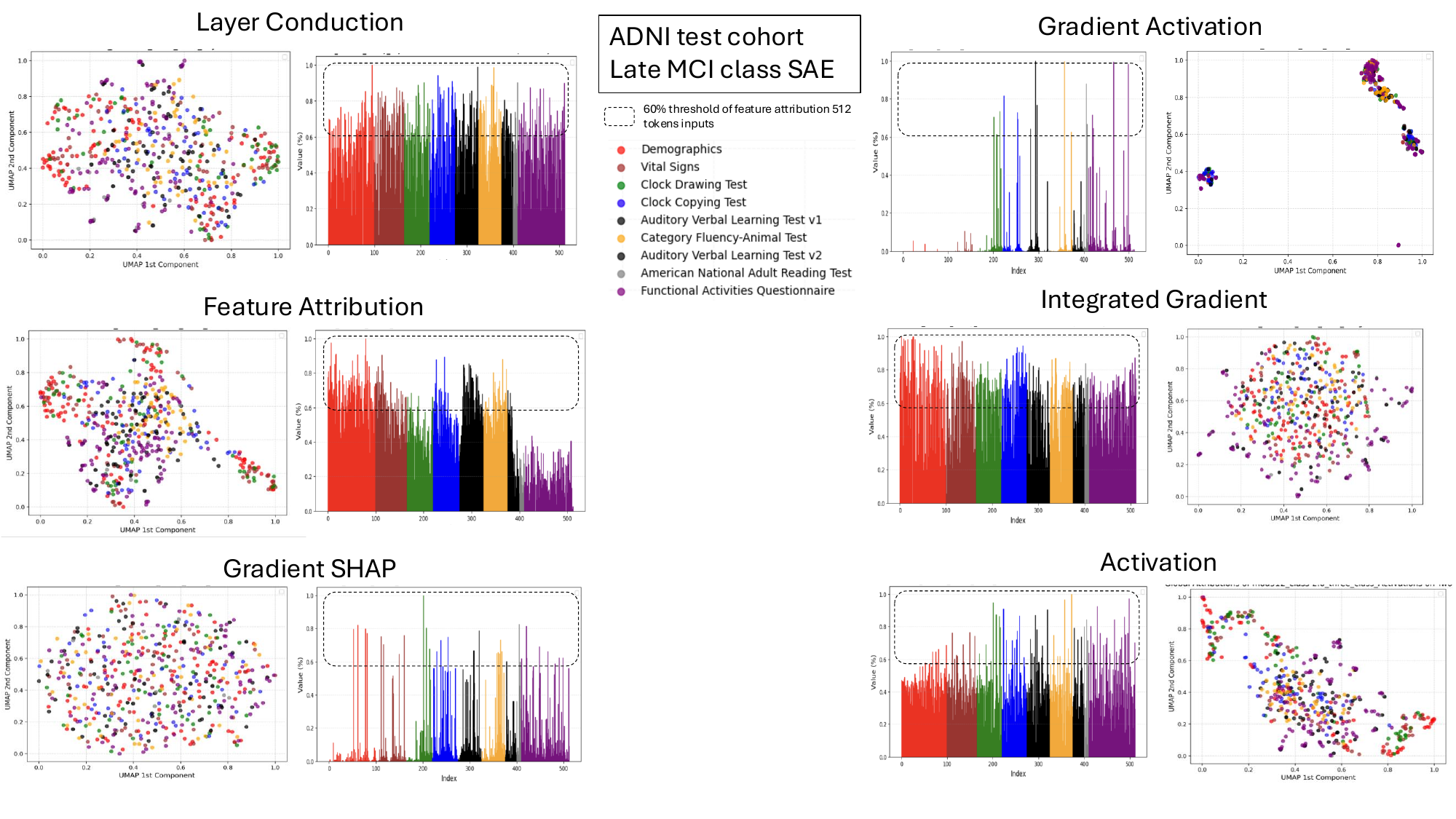}
\caption{Global (cohort-level) feature attribution across explanation methods with the SAE layer. The 2D panel shows a UMAP embedding (UMAP-1 vs UMAP-2) computed on the ADNI test set; the 1D panel shows attribution scores along PCA first component. All plotted values are normalised to [0, 1] and represent positive contributions only. Colours (red→purple) denote the nine ADNI subgroups (see §B3). Square boxes mark the 0.6–1.0 interval, highlighting the most significant tokens in the 1D views. The task is a three-class classification (LMCI, MCI disease vs Control) on the ADNI cohort; the examples shown here are from the MCI class.}
\label{ar9}
\end{figure*}
\begin{figure*}
\centering 
\includegraphics[width=0.9\textwidth]{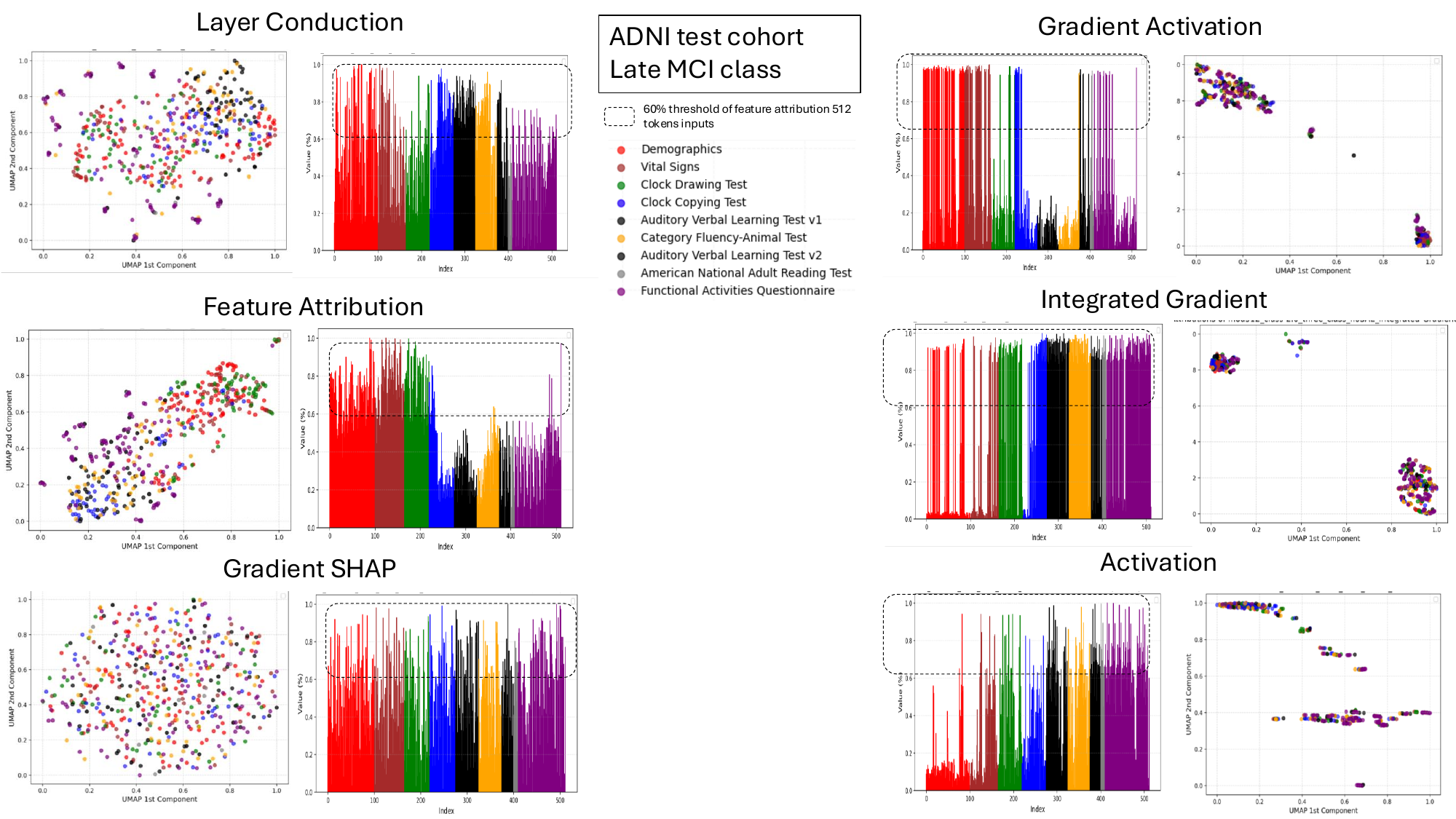}
\caption{Global (cohort-level) feature attribution across explanation methods without the SAE layer. The 2D panel shows a UMAP embedding (UMAP-1 vs UMAP-2) computed on the ADNI test set; the 1D panel shows attribution scores along PCA first component. All plotted values are normalised to [0, 1] and represent positive contributions only. Colours (red→purple) denote the nine ADNI subgroups (see §B3). Square boxes mark the 0.6–1.0 interval, highlighting the most significant tokens in the 1D views. The task is a three-class classification (LMCI, MCI disease vs Control) on the ADNI cohort; the examples shown here are from the MCI class.}
\label{ar10}
\end{figure*}

Figures~\ref{ar1}–\ref{ar10} present cohort-level attribution examples for both the binary (Control vs Alzheimer’s disease) and three-class (Control, LMCI, MCI) classification tasks on the ADNI test cohort across six explanation methods. Each method is shown without (Figures~\ref{ar1}, \ref{ar3}, \ref{ar5}, \ref{ar7}, \ref{ar9}) and with (Figures~\ref{ar2}, \ref{ar4}, \ref{ar6}, \ref{ar8}, \ref{ar10}) the Sparse Autoencoder (SAE) layer. The 2D panel shows a UMAP embedding (UMAP-1 vs UMAP-2) computed on the ADNI test set; the 1D panel shows attribution scores along PCA-first component. All plotted values are normalised to [0,1] and represent positive contributions only. Colours (red→purple) denote the nine ADNI subgroups (see §B3). In general, moving from the no-SAE to the SAE condition broadens the distribution of features in 2D and increases the density of high-significance points (upper-right boxed region), consistent with a decoder-induced decompression effect and a corresponding reduction in sparsity in the attribution maps.

\begin{figure*}
\centering 
\includegraphics[width=0.9\textwidth]{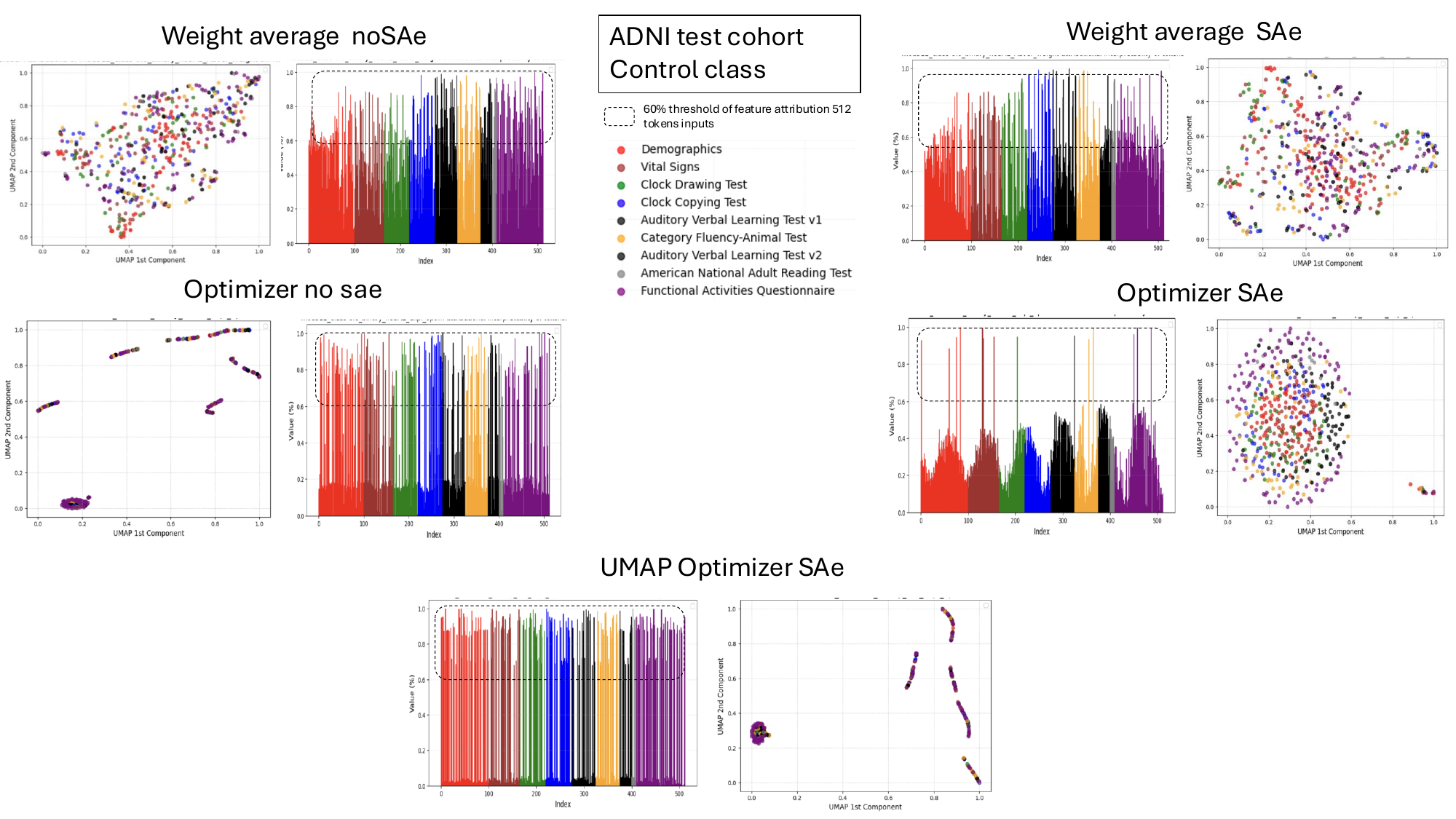}
\caption{Global (cohort-level) feature attribution across explanation methods with the SAE layer. The 2D panel shows a UMAP embedding (UMAP-1 vs UMAP-2) computed on the ADNI test set; the 1D panel shows attribution scores along PCA first component. All plotted values are normalised to [0, 1] and represent positive contributions only. Colours (red→purple) denote the nine ADNI subgroups (see §B3). Square boxes mark the 0.6–1.0 interval, highlighting the most significant tokens in the 1D views. The task is a binary classification (Alzheimer vs Control) on the ADNI cohort; the examples shown here are from the Control class.}
\label{ar11}
\end{figure*}

\begin{figure*}
\centering 
\includegraphics[width=0.9\textwidth]{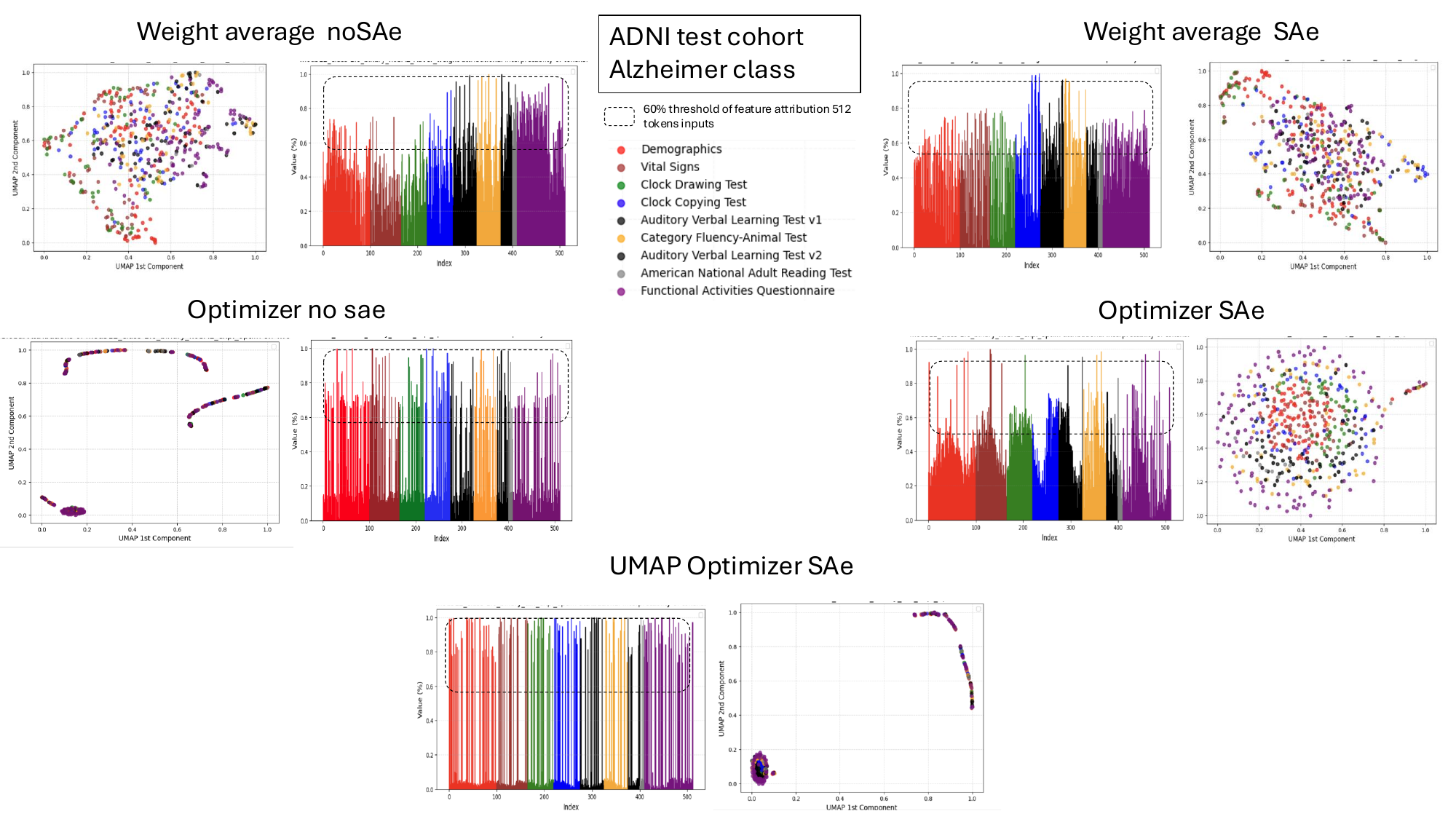}
\caption{Global (cohort-level) feature attribution across explanation methods with the SAE layer. The 2D panel shows a UMAP embedding (UMAP-1 vs UMAP-2) computed on the ADNI test set; the 1D panel shows attribution scores along PCA first component. All plotted values are normalised to [0, 1] and represent positive contributions only. Colours (red→purple) denote the nine ADNI subgroups (see §B3). Square boxes mark the 0.6–1.0 interval, highlighting the most significant tokens in the 1D views. The task is a binary classification (Alzheimer vs Control) on the ADNI cohort; the examples shown here are from the Alzheimer class.}
\label{ar12}
\end{figure*}

\begin{figure*}
\centering 
\includegraphics[width=0.9\textwidth]{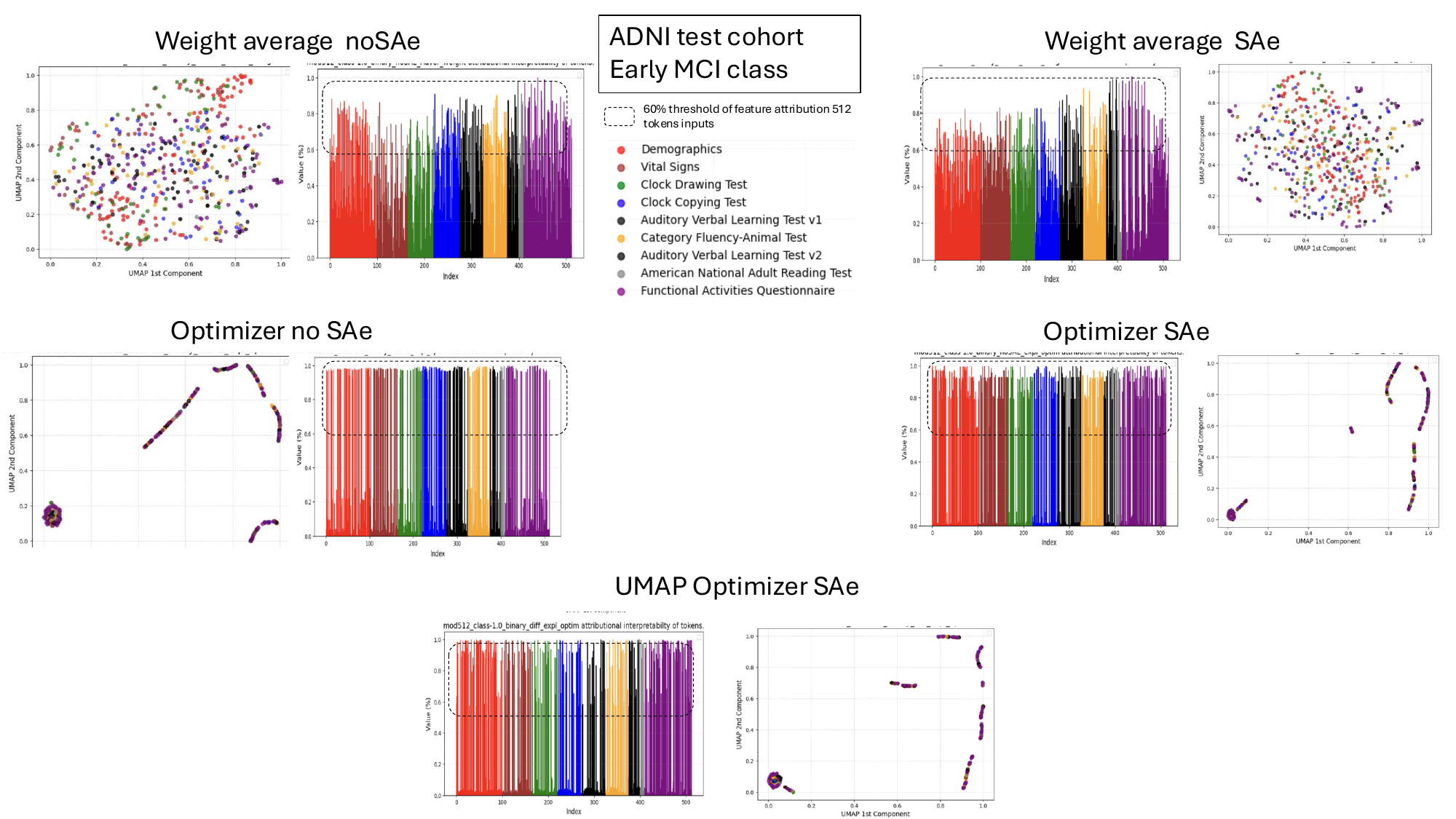}
\caption{Global (cohort-level) feature attribution across explanation methods without the SAE layer. The 2D panel shows a UMAP embedding (UMAP-1 vs UMAP-2) computed on the ADNI test set; the 1D panel shows attribution scores along PCA first component. All plotted values are normalised to [0, 1] and represent positive contributions only. Colours (red→purple) denote the nine ADNI subgroups (see §B3). Square boxes mark the 0.6–1.0 interval, highlighting the most significant tokens in the 1D views. The task is a three-class classification (LMCI, MCI disease vs Control) on the ADNI cohort; the examples shown here are from the LMCI class.}
\label{ar13}
\end{figure*}

\begin{figure*}
\centering 
\includegraphics[width=0.9\textwidth]{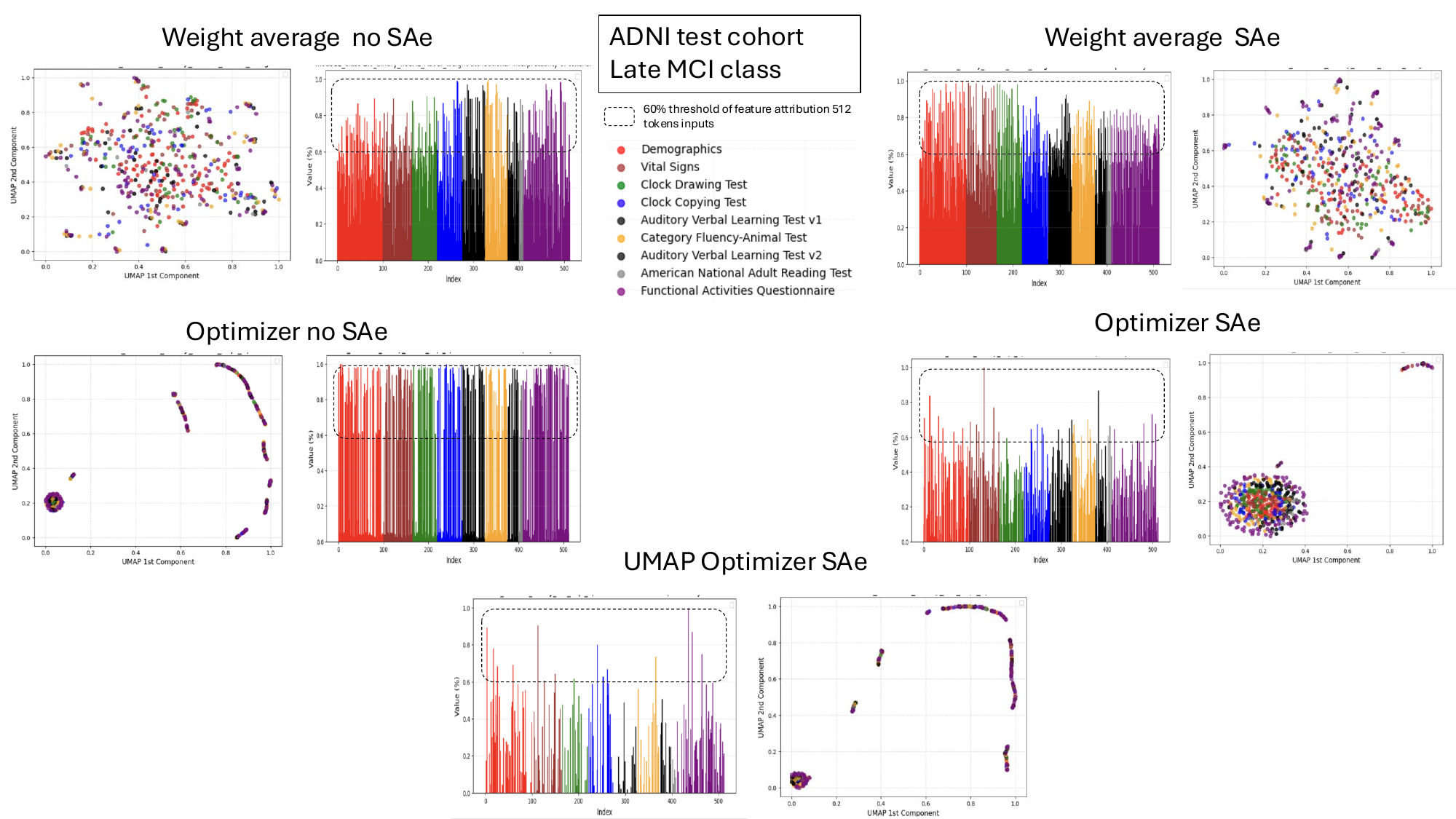}
\caption{Global (cohort-level) feature attribution across explanation methods with the SAE layer. The 2D panel shows a UMAP embedding (UMAP-1 vs UMAP-2) computed on the ADNI test set; the 1D panel shows attribution scores along PCA first component. All plotted values are normalised to [0, 1] and represent positive contributions only. Colours (red→purple) denote the nine ADNI subgroups (see §B3). Square boxes mark the 0.6–1.0 interval, highlighting the most significant tokens in the 1D views. The task is a three-class classification (LMCI, MCI disease vs Control) on the ADNI cohort; the examples shown here are from the MCI class.}
\label{ar14}
\end{figure*}

Figures~\ref{ar11}–\ref{ar14} present cohort-level attribution examples for both the binary (Control vs Alzheimer’s disease) and three-class (Control, LMCI, MCI) classification tasks on the ADNI test cohort, analogous to Figures~\ref{ar1}–\ref{ar10}, for the no-SAE analyses of (i) the attributional weighted average (computed from the six base methods), (ii) the Transformer Explanation Optimizer (TEO), and (iii) TEO with a linear UMAP constraint (UMAP Optimizer). As shown in the previous subsection, with the SAE layer TEO achieves the best stability—i.e., the lowest RIS and ROS—but at the cost of a marked reduction in Sparseness; this reduction is clearly visible in the binary task (Figures~\ref{ar11}–\ref{ar14}), where a spreading of tokens in 2D is observed when moving from no-SAE to SAE, as with the other methods. TEO with SAE reorganises the space, yielding a more homogeneous low-to-high attribution gradient. The drawback is that, without appropriate guidance, there may be too few features in the squares denoting significant contribution, and not all subgroups in the global observations are represented (e.g., Figure~\ref{ar11}). However, this can be mitigated by constraining the 2D manifold in the attribution space. To that end, we proposed a linear constraint to further smooth the regrouping of tokens in the attribution manifold. Introducing the UMAP linear constraint yields an even more balanced trade-off compared with unconstrained TEO with SAE, producing explanations that share similar significant traits across the different subgroups (colours) and are more homogeneous (very clear in Figures~\ref{ar11}, \ref{ar12}, and \ref{ar14}, less so in \ref{ar13}). Consequently, the maps are more compact and clinically interpretable; the same behaviour is observed across all classes in the three-class setting (Figures~\ref{ar12}–\ref{ar14}). By contrast, at both cohort and local levels, the weighted-average approach—a linear combination of the six attribution techniques—does not yield superior explanations, consistent with \cite{mamalakis}.

\subsection{Limitation and Future Work}
Although our framework demonstrates substantial improvements in attribution clarity and robustness, some limitations remain. A generalized outcome about clinical LMs is not feasible at the level of this study, as the analysis was restricted to the neurodegenerative domain, limiting generalisability to other areas such as oncology. Constraining the manifold space of explanations with explicit guidance from clinical experts could further improve explanation quality and enhance pattern discovery within the proposed framework. In addition, while stability–sparsity assessment focused on RIS/ROS and sparsity indices as important first-level evaluation metrics, additional measures such as uncertainty quantification and fairness auditing should be incorporated in future work. Future work will further strengthen these results by 
\textcolor{black}{integrating clinical experts into the loop to refine a more analytical vocabulary and to further characterize the enriched patterns identified by the proposed framework}. Lastly, we aim to prospectively validate the approach, extend it to additional centres, modalities, and clinical domains (e.g., oncology), explore alternative constraints, and incorporate uncertainty and fairness auditing.


\end{document}